\documentclass[twoside]{article}

\usepackage[accepted]{mytemplate}

\usepackage{hyperref}

\usepackage[normalem]{ulem}
\usepackage{bm}                  
\usepackage{amsmath}             
\usepackage{amssymb}			 
\usepackage{amsfonts}            
\usepackage{cancel}              
\usepackage{bbold}               
\usepackage{graphicx}			 
\usepackage{float}			     
\usepackage{array}			     
\usepackage{tabularx}			 
\usepackage{longtable}			 
\usepackage{colortbl}            
\usepackage[thinlines]{easytable}
\usepackage{multirow}            
\usepackage{booktabs}			 
\usepackage{subcaption}          

\usepackage{xr}                  

\usepackage{todonotes}

\usepackage{minitoc}

\DeclareMathOperator{\Tr}{Tr}
\DeclareMathOperator{\trans}{T}
\DeclareMathOperator{\dx}{\mathrm{d} \mathbf{x}}

\begin{document}

\doparttoc 
\faketableofcontents 

\twocolumn[

\mytitle{Density estimation via binless multidimensional integration}

\myauthor{ Matteo Carli$^*$ \And  Alex Rodriguez \And Alessandro Laio$^*$ \And Aldo Glielmo$^*$}

\myaddress{ SISSA, Italy\\Harvard University, USA \And University of Trieste, Italy\\ICTP, Italy \And SISSA, Italy\\ICTP, Italy \And  Banca d'Italia$^\dag$, Italy\\SISSA, Italy }]

{\let\thefootnote\relax
\footnotetext{
\textsuperscript{*}\\
mcarli@sissa.it,
laio@sissa.it, aldo.glielmo@bancaditalia.it. }
\footnotetext{
\textsuperscript{$\dag$} The views and opinions expressed in this paper are those of the authors and do not
necessarily reflect the official policy or position of Banca d'Italia. }
}

\begin{abstract}

We introduce the Binless Multidimensional Thermodynamic Integration (BMTI) method for nonparametric, robust, and data-efficient density estimation.
BMTI estimates the logarithm of the density by initially computing log-density differences between neighbouring data points.
Subsequently, such differences are integrated, weighted by their associated uncertainties, using a maximum-likelihood formulation. 
This procedure can be seen as an extension to a multidimensional setting of the \emph{thermodynamic integration}, a technique developed in statistical physics.
The method leverages the manifold hypothesis, estimating quantities within the intrinsic data manifold without defining an explicit coordinate map. 
It does not rely on any binning or space partitioning, but rather on the construction of a neighbourhood graph based on an adaptive bandwidth selection procedure.
BMTI mitigates the limitations commonly associated with traditional nonparametric density estimators, effectively reconstructing smooth profiles
even in high-dimensional embedding spaces.
The method is tested on a variety of complex synthetic high-dimensional datasets, where it is shown to outperform traditional estimators, and is benchmarked on realistic datasets from the chemical physics literature.

\end{abstract}

\section{Introduction}
\label{sec:introduction}
Estimating a Probability Density Function (PDF) from a finite set of samples is a fundamental challenge in statistics and machine learning, arising in a variety of practical applications \cite{silverman1986density,Scott2015,Glielmo2021unsup}.
In pursuing this task, parametric methods assume a functional form for the PDF and try to optimise a few parameters to best fit the observations \cite{Hastie2009,ML_EM,Bengio2013,Lecun2015,Schmidhuber2015}.
They return smooth and robust PDF estimates even with little data, but badly specified parametric models can introduce systematic errors that are not healed by statistics~\cite{Bishop2006}.
Nonparametric density estimators, instead, seek to estimate a PDF without prior assumptions on the distribution 
\cite{Izenman1991}.
This makes them more data-hungry, but also preferable to parametric ones when the shape and topography of the distribution peaks are not known a priori \cite{izenman2008modern}.
%
A popular example of a non-parametric density estimator is the Kernel Density Estimator (KDE) \cite{Parzen1962OnMode}, which reconstructs the PDF as a mixture of local copies of kernel functions -- often Gaussians -- around each data point.
The \textit{k} Nearest Neighbor (\textit{k}NN) estimator \cite{Fix1951DiscriminatoryProperties} is another classic example: it estimates the density around each point proportionally to the inverse volume occupied in embedding space by the \textit{k} nearest neighbours of that point.
It can be thought of as a special type of KDE with a step-function kernel and an adaptive bandwidth selection.
Gaussian KDEs perform remarkably well in very low dimensions, where they are typically preferred to \textit{k}NN estimators as they provide much smoother density estimates. 
In fact, the kernel of \textit{k}NN is non-differentiable. 
Practically, this can translate into noisier and less accurate estimates \cite{silverman1986density,Izenman1991}.
However, the performance of fixed-bandwidth KDEs drastically deteriorates with the increase of the embedding space dimension.
Already beyond 2 or 3 dimensions, these estimates become biased~\cite{Scott2015} and \textit{k}NN outperforms standard KDEs due to its point-adaptive nature.
In fact, high-dimensional data face the so-called \textit{curse of dimensionality}~\cite{Bellman1961AdaptiveCP,Friedman1997,Bishop2006}.

The selection of the smoothing parameter confronts a \textit{bias-variance tradeoff}: opting for a higher value enhances statistical stability while reducing noise, yet it may introduce bias when the underlying probability density function exhibits substantial variations across the spanned region. Achieving a delicate balance becomes even more crucial when dealing with sparse data, as often encountered in high-dimensional spaces.
To address this problem the bandwidth should be carefully selected globally or, preferably, locally (adaptively), a problem to which a great amount of research effort has been devoted \cite{bandwidth_0,bandwidth_1,bandwidth_2}. 
%
In this work, we address the described shortcomings of traditional
estimators and propose a nonparametric method able to provide accurate and smooth estimates of density in high dimensional spaces.
Our method is based on a reconstruction of the density starting from measurements of the difference of the logarithm of PDF (log-density hereafter) among pairs of neighbouring points.
The smoothness of the resulting density surface in our approach is guaranteed by the fact that the densities are estimated consistently on neighbouring points from the log-density difference between all the pairs of data within a neighbourhood. 
The log-density differences are estimated by a modified mean shift algorithm \cite{Fukunaga1975}, extended through a restriction to the intrinsic data manifold and an adaptive selection of local bandwidth.  
The accuracy of our approach in high dimensions arises both from the error-correcting nature of the pair-measurements reconstruction and by the fact that all relevant quantities can be estimated directly on the intrinsic data-manifold, and not in the embedding space.
The method we propose, named Binless Multidimensional Thermodynamic Integration (BMTI) is also remarkably data-efficient.
%

Our contributions are as follows:
\vspace{-0.2cm}
\begin{itemize}
    \setlength\itemsep{-0.2cm}
    \item We propose an extension of the mean-shift algorithm which allows estimating log-density gradients directly on the intrinsic data manifold using a local bandwidth selection. 
    The resulting estimates are considerably more robust against the curse of dimensionality. 
    
    \item We propose a method to reconstruct -- via a ``binless multidimensional integration'' 
    -- a log-density profile from measurements of log-density differences between neighbouring points.
    We demonstrate that the approach gives rise to a robust, data-efficient, and smooth density estimator and show that it outperforms state-of-the-art methods at least up to 20 dimensions. 
    \item We provide an open source code Python implementation of the BMTI log-density estimator and of the improved mean-shift gradient estimator \footnote{Our implementation is available within DADApy~\cite{Glielmo2022DADApyDA} at \url{https://github.com/sissa-data-science/DADApy}
    }
\end{itemize}

\section{Related work}
\label{sec:related_work}

    \paragraph{The manifold hypothesis and ID estimation}
    \label{ssec:related_work_manifold_id}
    According to the manifold hypothesis \cite{Bengio2013}, the distribution of any real-world high-dimensional dataset has support on a manifold, called \textit{intrinsic} or \textit{latent data manifold}, whose \textit{intrinsic dimension} (ID) \cite{Pope2021TheID} is typically much lower than the dimension of the embedding space \cite{Korn2001,Beygelzimer2006}. 
    In recent times the manifold hypothesis has found strong empirical evidence within several scientific fields, from physics and biology \cite{macocco2023intrinsic,facco2019intrinsic} to medicine \cite{konz2022intrinsic} and even epidemiology \cite{varghese2022intrinsic} or deep learning \cite{ansuini2019intrinsic,pope2021intrinsic}.
    This led to a renovated interest in the development of accurate intrinsic dimension estimators \cite{Camastra2016_Intrinsic_dimension,Facco2017,erba2019intrinsic,denti2022generalized,macocco2023intrinsic,Glielmo2022DADApyDA,bac2021scikit} all of which, importantly, need to be able to cope with manifolds which are, in general, curved, twisted and topologically complex \cite{Bengio2013}.
    The idea of estimating the PDF on the intrinsic manifold to alleviate the curse of dimensionality was first proposed in \cite{ozakin2009} for KDEs.
    In \cite{Rodriguez2018}, the authors use the concept of ID to improve traditional $k$NN-based estimators by restricting the computation of the volumes to the intrinsic manifold. 
    In \cite{liu2021density, horvat2023density}, the authors propose deep learning approaches for density estimation directly on the intrinsic manifold. In the first approach, they employ deep generative neural networks, while in the second approach, they extend the normalising flow scheme introduced in \cite{kobyzev2020normalizing}.
    \vspace{-0.4cm}
    \paragraph{Nonparametric density estimation}
    \label{par:related_work_nonparametric}
    Nonparametric methods try to infer the underlying PDF from a dataset without assuming its specific functional form.
    They are fully data-driven and only look at local properties of the sample \cite{Fukunaga1990}.
    They usually depend on at least one parameter, which typically controls this definition of locality, generally named \textit{smoothing parameter} or \textit{bandwidth}.
    The optimal choice of this parameter is a long-standing open problem in statistics \cite{Park1990ComparisonSelectors}.
    The simplest and oldest nonparametric methods are histograms \cite{pearson1894contributions}, which divide the embedding space into fixed partitions.
    KDEs, instead, are among the most popular and versatile.
    They first appeared in \cite{Fix1951DiscriminatoryProperties}, where the uniform KDE and the \textit{k}NN estimator were introduced \cite{Silverman1989}. 
    They were soon generalised to a wider class of functions, among which smooth kernels, yield a continuous PDF.
    Convergence and asymptotic properties were studied \cite{Rosenblatt1956,Parzen1962OnMode} and the approach was extended to the multidimensional case \cite{Cacoullos1964}.
    \textit{k}NN stands out as a brilliant but very simple way to adaptively select the bandwidth of a KDE.
    Instead of fixing explicitly the bandwidth, it sets the number of neighbours to be considered for each estimate, which makes the local level of smoothing depend on the local density.
    An extension of the \textit{k}NN idea to other kernels is the variable kernel estimator\cite{breiman1977variable}.
    Other proposed methods for bandwidth selection include global methods \cite{silverman1986density,jones1996brief,izenman2008modern,Zhao2022kNN} and adaptive ones \cite{abramson1982bandwidth,abramson1982arbitrariness,fukunaga1982kNNparametrically,Hall1988adaptiveker,myles1990AdaptiveKNN,Hastie2009}.
    Another class worth mentioning is composed of Voronoi density estimators \cite{ord1978VoronoiDE}.
    They do not introduce a geometric bias even in the choice of the kernel. 
    However, the computation of the Voronoi tessellation of the embedding space can be very computationally demanding and the estimates are typically discontinuous at the tile boundaries \cite{polianskii22Voronoi}.
    While these discontinuities have been recently treated \cite{Marchetti2023Voronoi}, the resulting PDF is still very spiky with low statistics.
    Finally, in recent years, neural network approaches have gained traction in nonparametric density estimation due to their great flexibility \cite{kingma2013VAE,goodfellow2014generative,dinh2016NormalizingFlows,papamakarios2017autoregressiveFlow,Liu2021DensityEU}.
    However, these approaches are quite data-hungry and computationally intensive, requiring large datasets and complex training schemes.

    \vspace{-0.3cm}
    \paragraph{Unnormalised log-density, energy, and free energy}
    \label{ssec:related_work_unnorm_logden}
    %
    In the task of inferring a PDF from a finite sample, computing the correct normalization can be very challenging if not prohibitive \cite{hyvarinen2005jmlr}. Fortunately, in many practical applications, one is only interested in the derivatives of the PDF \cite{Fukunaga1975} or in the relative value of the PDF at different points \cite{ranzato2007LeCunEBM-UL,PIETRUCCI201732}. It is the case of several pattern-recognition and image processing algorithms \cite{Fukunaga1990,ester1996density,comaniciu1999mean,yang2003mean,LeCun2006ATO,KulczyckiCharytanowicz2010,Rodriguez2014,dErrico2021}, but also 
    of physics, where the log-density over a projection of configuration space is typically proportional to a thermodynamic potential called \textit{free energy}\cite{tuckerman2010statistical,Laio2002,weinan2004metastability,Chipot2007}. In the ML community, the unnormalised negative log-density quantity is referred to as \textit{energy} \cite{Saremi2018DeepEE}, like in the case of \textit{energy-based models} (EBMs) \cite{teh2003energy,lecun2004learning}. 
    Such quantity is the one we are concerned with in this work.

    \vspace{-0.3cm}
    \paragraph{Thermodynamic integration}
    \label{par:related_work_TI}
    The term \emph{thermodynamic integration} (TI) indicates a technique first developed in the field of chemical physics to reconstruct the free energy of a system through the knowledge of its derivatives.
    Traditionally, TI is carried out by integrating a directional derivative along a carefully 
    chosen, physically meaningful, one-dimensional path~\cite{Kirkwood1935,Chipot2007}.
    The approach has been extended to many dimensions~\cite{Dehez2007}, but integrating the gradient in more than one dimension is considered a difficult task~\cite{Ciccotti2005}. 
    Often the gradient is sampled on a dense grid~\cite{Henin2010}, a data-intensive approach allowing for free energy reconstruction only up to dimension $2$ or $3$~\cite{Dehez2007,Comer2015,marinova2019}.
    In these cases, the most accurate integration method is the solution of the Poisson equation on a shifted mesh \cite{Lelievre2010,Alrachid2015,Henin2021b}.
    The most common integration method in literature \cite{Abrams2010,Chen2012}, introduced in~\cite{Maragliano2008} makes use of radial basis functions to reconstruct the free energy from the gradient, which has however been employed in spaces of no more than $4$ dimensions due to its high computational demand.
    Finally, a recently proposed method \cite{Marinelli2021a} uses gradient information to define a Monte Carlo protocol to smartly sample the embedding space and populate a multidimensional histogram. 
    This approach has been pushed, quite astonishingly, up to dimension $6$ and it represents, to our knowledge, the highest embedding dimension at which TI has been previously employed.

\section{Binless Multidimensional Thermodynamic Integration}
\label{sec:BMTI}
Let $\rho(\mathbf{x})$ be the PDF of the multidimensional random variable $\mathbf{x}$. We assume that all the $\mathbf{x}_i$ are harvested independently from $\rho$.
The goal of our approach is estimating the negative of the logarithm of the density, which from now we refer to as the negative log-density (NLD), and denote $F(\mathbf{x}) = - \ln \rho(\mathbf{x})$. 
The minus sign is introduced for consistency with the physics and physical chemistry literature, where such quantity is interpreted as a free energy. 
Due to this convention the maxima of $\rho(\mathbf{x})$ correspond to the minima of $F(\mathbf{x})$. 
We denote the NLD at a given datapoint $\mathbf{x}_i$ by $F_i = F(\mathbf{x}_i)$.

Suppose that each point $\mathbf{x}_i$ has a number $k_i$ of neighbours that can be assumed to have a similar density. Importantly, the value of $k_i$ is point-dependent. This neighbourhood will be defined more precisely in the following.
This induces a -- typically-sparse -- directed \emph{neighbourhood graph} (NG) (see Fig.~\ref{fig:ABCD}) in which $i$ is connected to $j$ only if $j$ is a neighbour of $i$.
Further suppose that for any connected pair $(i,j)$ in the NG we have an estimate of the NLD difference $\delta F_{ij} = F_j - F_i$.
Then, in the spirit of the TI the NLD difference $\Delta F_{il}$ between \emph{any} two points $i$ and $l$, can be computed by choosing any of the multiple paths that connect $i$ to $l$ on the NG
\begin{center}
    $\Delta F_{il} \simeq \delta F_{i,j_1} + \delta F_{j_1,j_2} + \dots + \delta F_{j_n, l}$
\end{center}
where $j_1, \dots, j_n$ define one specific path from $i$ to $l$. 
For each pair of endpoints $i$ and $l$ there are many possible choices of paths, but of course $\Delta F_{il}$ must be the same in all cases.
This fact calls for a procedure that estimates NLD differences among points considering contributions from different paths connecting them. 
In the rest of this section, we will introduce the \emph{Binless Multidimensional Thermodynamic Integration} (BMTI) estimator, which implements this idea.
BMTI estimates negative log-densities 
at each point of the dataset simultaneously as the solution of a linear system obtained by maximizing a likelihood that incorporates the paths' weights.
BMTI is \emph{Binless} as it performs \emph{TI} on a graph rather than using some kind of configuration space partitioning, and is \emph{Multidimensional} since it performs remarkably well for high-dimensional data, for which, we will show, it typically outperforms other non-parametric methods.

    \subsection{Derivation of the BMTI estimator}
    \label{ssec:BMTI_derivation}
\begin{figure*}[!ht]
    \advance\leftskip-0.5cm
     \centering
    \includegraphics[width=.9\textwidth]{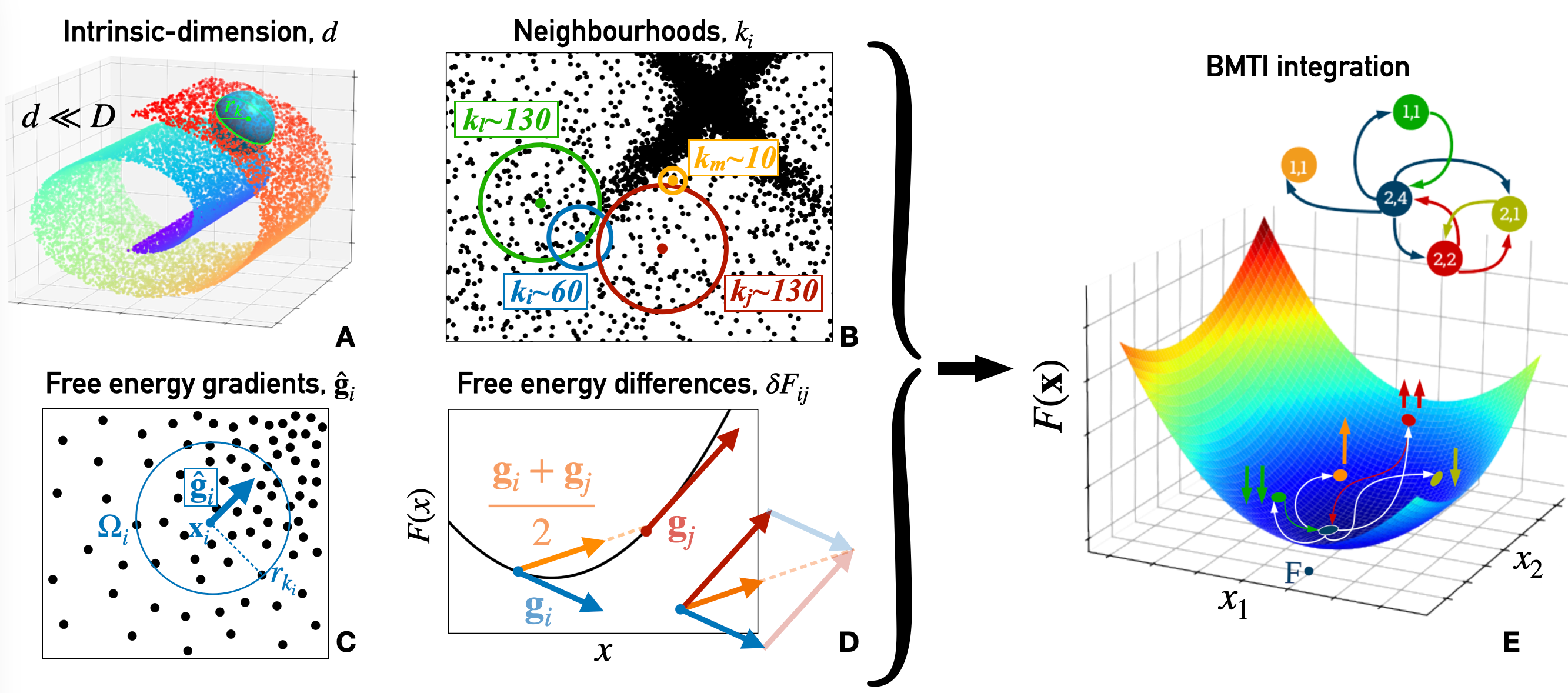}
    \vspace{-0.25cm}
    \caption{
    \textbf{The BMTI method}
    Panels \textbf{A} to \textbf{D} illustrate the 4 steps, described in Sec.~\ref{ssec:BMTI_deltaF}, needed to construct the BMTI log-likelihood: estimating the intrinsic dimension $d$, adaptive neighbourhoods selection and the neighbourhood graph, NLD gradients $\hat{g}_i$, and finally NLD differences $\hat{\delta F}$ estimation.
    Panel \textbf{E} illustrates the reconstruction of the NLD starting from measurements of NLD differences as described in Sec.~\ref{ssec:BMTI_deltaF}.
    In this illustration the NLD $\hat{F}_i$ at point $i$ (blue dot) is computed by taking into consideration $\hat{\delta F}$ contributions from 4 neighbours (green, orange, red, and yellow dots).
    The contributions push for increasing (upward arrows) or decreasing (downward arrows) the $\hat{F}_i$ value.
    }
    \label{fig:ABCD}
    \end{figure*}

    For a lighter notation, let us label a pair of points  linked by an edge on the NG by a single index $a = (i,j)$, so that 
    $\{a,b,\dots \, \}$ is the set of $N_e$ directed edges of the NG $\{(i,j),(l,m),\dots\} $, with $N_e = \sum_{i=1}^N (k_i-1) = N (\langle k\rangle - 1)$.
    %
    %
    We assume that for each pair of neighbouring points $a$ we have an unbiased estimate $\hat{\delta F}_a$ of the true NLD difference $\delta F_a$.
    %
    We further assume that our vector of estimates $(\hat{\bm{\delta F}})_a:= \hat{\delta F}_a$ is normally distributed around the true values $(\bm{\delta F})_a := \delta F_a$. This implies that our estimator $\hat{\bm{\delta F}}$ follows a multivariate normal distribution centred on the true vector $\bm{\delta F}$ with covariance matrix $\mathbf{C} := \mathrm{\mathbf{cov}}[\hat{\bm{\delta F}},\hat{\bm{\delta F}}]$.
    In short, we assume
    \begin{equation} 
        \hat{\bm{\delta F}} \sim \mathcal{N}( \bm{\delta F} , \mathbf{C}) \ .
    \label{eq:BMTI_deltaF_normally_distributed}
    \end{equation}
    In Sec.~\ref{ssec:BMTI_deltaF} we introduce a procedure to effectively estimate NLD differences on the NG and their covariance.
    Now, by focusing on the $\hat{\delta F}$'s distribution 
    \small
    \[
    \mathcal{N}( \bm{\delta F} , \mathbf{C}) \propto \exp
    \left[
    -\frac{1}{2} \sum_{a,b} (\delta F_a - \hat{\delta F}_a)^{\trans} C^{-1}_{a,b} (\delta F_b - \hat{\delta F}_b)
    \right],
    \]
    \normalsize
    %
    we note that the argument of the exponential, in square brackets, can be recast into a quadratic form for the $F$'s.
    By calling $(\mathbf{F})_i:= F_i$ the vector of all the negative log-densities at sample points, this quadratic form reads $\mathbf{F}^{\trans} \mathbf{A} \mathbf{F} + \mathbf{b}^{\trans} \mathbf{F} + c$, where the $N \times N$ matrix $\mathbf{A}$, the $N$-vector $\mathbf{b}$ and the scalar $c$ are explicit functions of the estimated NLD differences $\hat{\bm{\delta F}}$ and on their covariance matrix $\mathbf{C}$. 
    Therefore, we can interpret the logarithm of this $N_{e}$-variate Gaussian as a log-likelihood for the parameters $\mathbf{F}$ given
    the error-affected observations $\bm{\hat{\delta F}}$
    \begin{equation}
        \mathcal{L} ( \mathbf{F} \mid \hat{\bm{\delta F}}\,,\, \mathbf{C} ) \; \propto \; \ln \,\mathcal{N}( \bm{\delta F} \,,\,\mathbf{C})\ .
        \label{eq:BMTI_loglikelihood_full}
    \end{equation}
    By maximizing this log-likelihood over the parameters $\mathbf{F}$, one obtains a linear system
    \begin{equation}
        \hat{\mathbf{F}} := \underset{\mathbf{F}}{\operatorname{argmax}} ~ \mathcal{L} ( \mathbf{F} \mid \hat{\bm{\delta F}}\,,\, \mathbf{C} )
        ~~~ \Rightarrow ~~~
        \mathbf{A} \; \hat{\mathbf{F}} = \mathbf{b}
    \label{eq:BMTI_likelihood_maximisation}
    \end{equation}
    whose solution defines the BMTI NLD estimator
    \begin{equation}
        \hat{\mathbf{F}} = \mathbf{A}^{-1} \, \mathbf{b} \; .
    \label{eq:BMTI_F_estimator}
    \end{equation}
    The maximum-likelihood estimates $\hat{\mathbf{F}}$ are thus found simultaneously and coherently for all points in the dataset through a mechanism that is illustrated in Fig.~\ref{fig:ABCD}E.
    Locally, the estimated $\hat{F}_i$ receives additive contributions from its neighbours. These contributions are positive or negative depending on the sign of $\hat{\delta F}_{ij}$ and inversely proportional to the uncertainties on the $\hat{\delta F}$'s involving $i$ or $j$.
    This framework also allows to estimate Cramér–Rao uncertainties from the Hessian of the log-likelihood (see App. \ref{ssec:BMTI_support_error})
    \begin{equation}
        \sigma^2[\hat{F}_i] := A^{-1}_{ii}\; .
    \label{eq:BMTI_F_error}
    \end{equation}
    %
    Since the NLD enter the log-likelihood only in terms of differences, the values of the NLDs are determined up to an arbitrary additive constant.
    This makes the linear system in Eq.~(\ref{eq:BMTI_likelihood_maximisation})  underdetermined and the symmetric matrix $\mathbf{A}$ singular.
    In practice, the linear system can be regularised without loss of accuracy and solved using any standard linear algebra library implementing,  e.g., the conjugate gradient method\cite{Hestenes&Stiefel:1952}, while the matrix $\mathbf{A}^{\!-1}$ appearing in Eq.s (\ref{eq:BMTI_F_estimator}) and (\ref{eq:BMTI_F_error}) can be interpreted as a Moore-Penrose pseudoinverse of $\mathbf{A}$\cite{penrose_1956}.

    \subsection{Estimation of the BMTI log-likelihood}
    \label{ssec:BMTI_deltaF}
    %
    In order to compute the log-likelihood in Eq.~(\ref{eq:BMTI_loglikelihood_full}) we need to evaluate the $\hat{\delta F}$'s and their covariances.
    This is done in four steps: (i) the estimation of the intrinsic dimension, (ii) an adaptive neighbourhood selection and the construction of an NG, (iii) the estimation of local NLD gradients, and finally (iv) the estimation of NLD differences and their correlations.
    These steps are summarised in Fig.~\ref{fig:ABCD}A-D. 
    The first two are based on previous work and are described in Sec.~\ref{sssec:BMTI_deltaF_manifold+adaptive}.
    Step (iii) is described in Sec.~\ref{sssec:BMTI_deltaF_grad_F}, while step (iv) is covered in Sec.~\ref{sssec:BMTI_deltaF_delta_F} and in the SM.

        \subsubsection{Restriction to the intrinsic manifold and adaptive neighbourhood selection}
        \label{sssec:BMTI_deltaF_manifold+adaptive}
         In order to restrict density estimates to the intrinsic data manifold, the first step is to reliably estimate its ID, $d$.
        In this work we adopt the \textit{TwoNN} estimator~\cite{Facco2017} as implemented in~\cite{Glielmo2022DADApyDA}. 
        The next assumption is that the data manifold $\mathcal{M}$ is Riemannian \cite{ozakin2009}.
        In this setting, on a small enough scale, geodesic distances on $\mathcal{M}$ are equal to Euclidean distances in the $d$-dimensional tangent hyperplane to $\mathcal{M}$;
        these are in turn equivalent to Euclidean distances in the embedding space of dimension $D$, since the PDF vanishes outside $\mathcal{M}$ \cite{Rodriguez2018,Carli2021} (see Fig.~\ref{fig:ABCD}A).
        Therefore, small local hyperspherical volumes in $\mathcal{M}$ can be computed as the volume of an Euclidean $d$-ball. We remark that an accurate estimate of $d$ is crucial to estimate these volumes.
        BMTI also relies on the definition of the number $k_i$ of nearest neighbours for each point $i$ of the dataset.
        This is needed for two reasons:
        it selects a local region of space $\Omega_i$ for the estimation of NLD gradients, making the gradient estimates, and thus BMTI, point-adaptive and more robust in high dimensions;
        it allows for the construction of a directed NG in which $i$ is connected to $j$ when $j$ is in $\Omega_i$, which is essential for BMTI integration, as depicted at the top-right of Fig.~\ref{fig:ABCD}E.
        In our specific case, for each point $i$, we select the largest number of neighbours $k_i$ over which the density does not vary significantly. For this purpose, we use the likelihood ratio test proposed in \cite{Rodriguez2018}, where $k_i$ gradually increases until the density of point $i$ and its $k_i$-th nearest neighbour are statistically different according to a certain tolerance.
        We refer the reader to Sec. \ref{append:k_adaptive-kstar_dc} of the SM for an intuitive explanation and visualisation, and to the original paper for an extended explanation of this technique.
        Fig.~\ref{fig:ABCD}B illustrates the adaptive selection of $k_i$ and the construction of the neighbourhood graph.
        Note that alternative NG selection methods can also be adopted. 
        While BMTI performance will depend on the adaptivity and ability of the NG to define appropriately-sized neighbourhoods, the BMTI algorithm is defined for any choice of $\{k_i\}_i$. 
        In Sec \ref{append:k_adaptive-effect_on_BMTI} of the SM, we test the performance of BMTI with different choices of adaptive and fixed neighbourhood sizes. Overall, BMTI displays robustness with respect to the NG selection and consistently outperforms $k$NN estimators.
        %
        %
        %
        \subsubsection{Estimation of log-density gradients}
        \label{sssec:BMTI_deltaF_grad_F}
        
        For notation convenience, let us refer to the gradient of the NLD at point $i$ as $\mathbf{g}_i := \nabla_{\mathbf{x}}F(\mathbf{x}_i) = \nabla_{\mathbf{x}} \,\rho (\mathbf{x}_i) / \rho (\mathbf{x}_i)$.
        %
        %
        The \textit{mean shift} {$\mathbf{m}_i$} around point $i$ within the region $\Omega_i$ of radius $r_i$ is defined as the centered expectation $\mathbf{m}_i := \langle \mathbf{x} - \mathbf{x}_i \rangle_{\Omega_i}$. More explicitly
        \begin{equation}
        \mathbf{m}_i
            := \frac{\int_{\Omega_i} \rho(\mathbf{x}) (\mathbf{x} - \mathbf{x}_i) \, \mathrm{d}\mathbf{x}}{\int_{\Omega_i} \rho(\mathbf{x})\, \mathrm{d}\mathbf{x}}\ .
        \label{eq:grad_F_mean_shift_definition}
        \end{equation}
        By expanding the density $\rho(\mathbf{x})$ around $\mathbf{x}_i$ as 
        \begin{equation}
        \rho(\mathbf{x}) \approx
        \rho(\mathbf{x}_i) +
        \nabla^{\trans}_{\mathbf{x}}\rho(\mathbf{x}_i) (\mathbf{x}-\mathbf{x}_i),
        \label{eq:taylor-expansion}
        \end{equation}
        and by evaluating the integral in Eq.~(\ref{eq:grad_F_mean_shift_definition}) with the expression in Eq.~(\ref{eq:taylor-expansion}), we obtain the following approximate relationship between NLD gradient and mean shift, for which an intuition is provided by Fig.~\ref{fig:ABCD}C
        \begin{equation}
        \mathbf{g}_i
        \approx - \, \frac{d+2}{r_i^2} \; 
        \mathbf{m}_i \,.
        \label{eq:grad_F_grad_F_analytical_mean_shift}
        \end{equation}
        %
        By substituting the expectation $\mathbf{m}_i$ in the above equation with a sample average $\hat{\mathbf{m}}_i$, and by using the point-adaptively optimised neighbourhood size $k_i$, defined as described in Sec.~\ref{sssec:BMTI_deltaF_manifold+adaptive}, to identify the region $\Omega_i$ we estimate the NLD gradient (see again Fig.~\ref{fig:ABCD}C) as
        \begin{align}
          \hat{\mathbf{g}}_i &= -\, \frac{d+2}{r_{k_i}^2} \,  \hat{\mathbf{m}}_i \\
           \hat{\mathbf{m}}_i &= \frac{1}{\tilde{k}_i} \sum_{j \in \Omega_i} (\mathbf{x}_j - \mathbf{x}_i)
           \label{eq:grad_F_sample_grad_F}
        \end{align}
        where we defined $\tilde{k}_i = k_i -1$ for notation convenience, and $r_{k_i}$ is the distance between point $i$ and its $k_i$-th nearest neighbour.
        In Sec.~\ref{sssec:mean_shift_mean_shift_grad_F_mean_shift_anal} and \ref{sssec:mean_shift_mean_shift_grad_F_mean_shift_sample_estimator} of the SM we report a rigorous derivation of the two equations above, which generalise the procedure in Ref. \cite{Fukunaga1975}.
        It is worth stressing that $\hat{\mathbf{g}}_i$ in our approach is doubly adaptive in the same sense discussed in Sec.~\ref{sssec:BMTI_deltaF_manifold+adaptive}
        , since it restricts to the intrinsic manifold of dimension $d \ll D$ and operates an adaptive bandwidth selection; this is crucial to enhance the estimator's performance
        and robustness against the curse of dimensionality with respect to the original mean-shift gradient estimator \cite{Fukunaga1975}.
        Since the the $\hat{\mathbf{g}}_i$ estimator is proportional to the arithmetic average of $k \!- \!1$ shift random variables (RVs), i.e. to the sample mean-shift estimator $\hat{\mathbf{m}}_i$ defined in Eq.~(\ref{eq:grad_F_sample_grad_F}), its auto-covariance $\mathrm{\mathbf{var}}[\hat{\mathbf{g}}_i]$ is proportional to the auto-covariance of the mean shift.
        The sample gradient autocovariance estimator reads
        %
            \begin{align}
                    &\hat{\mathrm{\mathbf{var}}}[\hat{\mathbf{g}}_i]
                    =
                    \left( \frac{d+2}{r_{k_i}^{\!2}} \right)^2 
                    \!
                    \hat{\mathrm{\mathbf{var}}}[\hat{\mathbf{m}}_i]
                    \, .
                \label{eq:grad_F_grad_F_sample_autocovariance}
            \end{align}
        Eq.~(\ref{eq:grad_F_grad_F_sample_autocovariance}) allows to estimate uncertainties on $\hat{\mathbf{g}}_i$ and its components, as discussed more in depth in Sec.~\ref{sssec:grad_F_grad_F_mean_shift_grad_F_cov_structure}.

        \begin{figure*}[!ht]
        \centering
            \includegraphics[width=\textwidth]{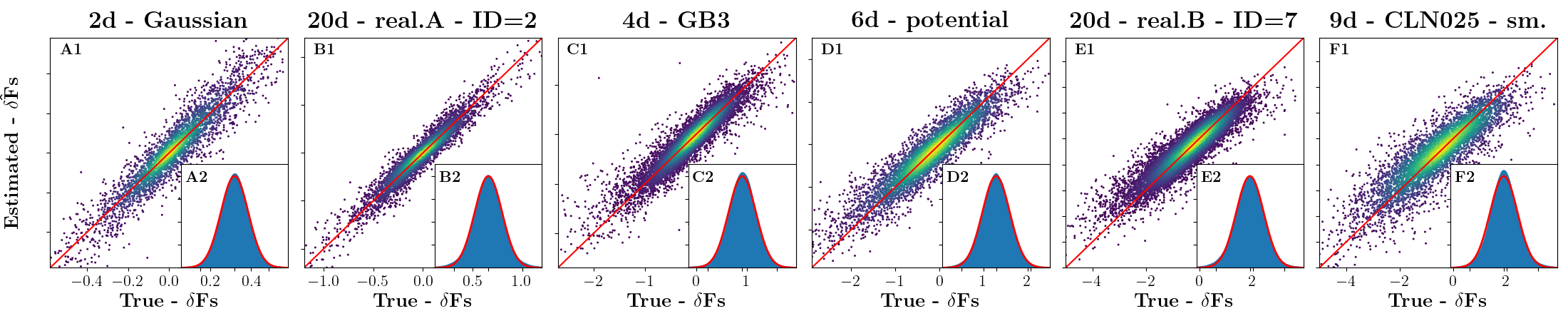}
            \vspace{-0.8cm}
        \caption{
        \textbf{Accuracy in the estimation of
        $\boldsymbol{\hat{\delta F}}$ and its error}.
        Density scatter plots of true vs estimated $\delta F$'s for 6 test datasets. 
        The insets show the distribution of the standardised variables $ (\hat{\delta F_{ij}} - \delta F_{ij})/\varepsilon_{ij}$ in blue, and a standard normal PDF in red; the agreement between the two demonstrate the accuracy of error estimates.
        }
        \label{fig:deltaF_perform}
        \vspace{-0.2cm}
        \end{figure*}
        
        Note that within our framework  other radially-symmetric kernels can be employed ~\cite{Comaniciu2002,yang2003mean,KulczyckiCharytanowicz2010}. We choose the extension of Mean Shift described above because, as we will show, it provides a reliable estimate of the NLD differences and of the error of their error even in high dimensions.
        Importantly, Ref.  \cite{Fukunaga1975} proves asymptotic unbiasedness, consistency, and uniform consistency of the estimator in Eq.~(\ref{eq:grad_F_sample_grad_F}) for all well-behaved kernel shapes \cite{Cacoullos1964}, thus these are guaranteed also in our case.
        The performance of $\hat{\mathbf{g}}$ and its error estimator are assessed in Sec.~\ref{ssec:grad_F_grad_F_performance} of the SM.
        \subsubsection{Estimation of the $\delta F$'s}
        \label{sssec:BMTI_deltaF_delta_F}
        Using the NLD gradient estimator $\hat{\mathbf{g}}$ in Eq.~(\ref{eq:grad_F_sample_grad_F}), we estimate the NLD differences $\delta F_{ij}$ between neighbouring points $\mathbf{x}_i$ and $\mathbf{x}_j$ as     
        \begin{equation}
            \hat{\delta F}_{ij} :=
            \frac{\hat{\mathbf{g}}_i + \hat{\mathbf{g}}_j}{2} \cdot \mathbf{r}_{ij}\ ,
        \label{eq:deltaFij_estim}
        \end{equation}
        where $\mathbf{r}_{ij} = \mathbf{x}_j-\mathbf{x}_i$.
        Contracting $\mathbf{r}_{ij}$ with the average of the two gradient estimates is more accurate than using any of the two singularly.  
        The intuition behind this choice is illustrated in Fig.~\ref{fig:ABCD}D. 
        Formally, it is simple to show that by taking the half-sum of $\hat{\mathbf{g}}_i$ and $\hat{\mathbf{g}}_j$ the quadratic error terms of the two estimators cancel out, leading to errors of order $\mathcal{O}(\|\mathbf{r}_{ij}\|^3)$ in the final estimator for $\delta F_{ij}$. A derivation of Eq.~(\ref{eq:deltaFij_estim}) can be found in Sec.~\ref{ssec:deltaFs_derivation} of the SM.

        The uncertainty on the estimate $\hat{\delta F}_{ij}$ can be quantified by its variance $\varepsilon^2_{ij} := \textrm{var}[\hat{\delta F}_{ij}]$, which can be derived from Eq.~(\ref{eq:deltaFij_estim}) to be
        \begin{align}
                \varepsilon^2_{ij}
                & =
                \mathbf{r}_{ij}^{\rm{T}} \,
                \mathbf{var}\left[\frac{\hat{\mathbf{g}}_i+ 
                \hat{\mathbf{g}}_j}{2} \right]\,
                \mathbf{r}_{ij} 
                \label{eq:deltaFij_variance}
                \\
                & = 
                \frac{1}{4}\,
                \mathbf{r}_{ij}^{\rm{T}}
                \left(
                \textbf{var}[\hat{\mathbf{g}}_i]
                +
                \textbf{var}[\hat{\mathbf{g}}_j]
                +
                2 \, \textbf{cov}[\hat{\mathbf{g}}_i,\hat{\mathbf{g}}_j]
                \right)
                \mathbf{r}_{ij} 
                \, .
                \nonumber
        \end{align}
        Note that the $\varepsilon^2_{ij}$ are indeed the diagonal elements of the covariance matrix $\mathbf{C}$ in Eq.~(\ref{eq:BMTI_deltaF_normally_distributed}) as $C_{a,a} = \mathrm{cov}[\hat{\delta F}_a,\hat{\delta F}_a] = \varepsilon^2_{a}$ .

        By defining the single-gradient estimates of the NLD as
        $\hat{\delta F}_{ij}^{\bm{i}} = \hat{\mathbf{g}}_i \cdot \mathbf{r}_{ij}$, with the upper bold index indicating whether the gradient on $i$ or $j$ is used, we can equivalently write the NLD estimates of Eq.~(\ref{eq:deltaFij_estim}) as
        $\hat{\delta F}_{ij} = \frac{1}{2}(\hat{\delta F}_{ij}^{\bm{i}}+\hat{\delta F}_{ij}^{\bm{j}})$ and their variance as ${\varepsilon_{ij}^{\bm{i}}}^2 :=\mathbf{r}_{ij}^{\rm{T}} \cdot \mathbf{var}[\hat{\mathbf{g}}_i] \cdot \mathbf{r}_{ij}^{\trans}$. 
        Finally, this allows us to rewrite Eq.~(\ref{eq:deltaFij_variance}) only in terms of scalar quantities
        %
        \begin{equation}
            \varepsilon^2_{ij} = \frac{1}{4}({\varepsilon_{ij}^{\bm{i}}}^2 + {\varepsilon_{ij}^{\bm{j}}}^2 + 2 \, p^{\bm{ij}} \, \varepsilon_{ij}^{\bm{i}} \,\varepsilon_{ij}^{\bm{j}})\ ,
        \label{eq:deltaFij_err_deltaF_ij_with_p}
        \end{equation}
        where $p^{\bm{ij}}$ is the Pearson correlation coefficient between $\hat{\delta F}_{ij}^{\bm{i}}$ and $\hat{\delta F}_{ij}^{\bm{j}}$ and is defined by
        \begin{equation}
            p^{\bm{ij}} 
            := 
            \frac{
            \mathrm{cov} \left[ \hat{\delta F}_{ij}^{\bm{i}},\hat{\delta F}_{ij}^{\bm{j}} \right]
            }{\varepsilon_{ij}^{\bm{i}} \,\varepsilon_{ij}^{\bm{j}}
            } 
            \, .
        \label{eq:deltaFij_pearson_def}
        \end{equation}
        The formulation in Eq.s (\ref{eq:deltaFij_err_deltaF_ij_with_p}) and (\ref{eq:deltaFij_pearson_def}) is at the basis of our approach for estimating the full covariance structure of the $\delta F$'s, as will become clear in the next section. 
        
        \subsubsection{Estimation of $\mathbf{C}$: the $\delta F$'s covariance}
        \label{sssec:BMTI_deltaF_C}
        
        In order to accurately reconstruct the NLD according to Eq.~(\ref{eq:BMTI_loglikelihood_full}), it is crucial 
        to estimate the covariance matrix $\mathbf{C}$ of the $\hat{\delta F}$'s, which -- evidently from Eq.~(\ref{eq:deltaFij_estim}) -- depends on the gradient covariance structure.
        The auto-covariance matrices of the gradient estimators appearing in Eq.~(\ref{eq:deltaFij_variance}), $\mathbf{var}[\hat{\mathbf{g}}_i]$ and $\mathbf{var}[\hat{\mathbf{g}}_j]$, can be estimated from the neighbourhoods $\Omega_i$ and $\Omega_j$ by means of Eq.~(\ref{eq:grad_F_grad_F_sample_autocovariance}).
        The cross-covariance between $\hat{\mathbf{g}}_i$ and $\hat{\mathbf{g}}_j$, instead, depends on a statistic computed on overlapping region between the two neighbourhoods, i.e. $\Omega_{i,j} := \Omega_i \,\cap\, \Omega_j$.
        In fact, as we show in Sec.~\ref{sssec:grad_F_grad_F_mean_shift_grad_F_cov_structure} of the SM, Eq.s (\ref{eq:mean_shift_sample_means_shift_cross_covariance}) and (\ref{eq:mean_shift_sample_grad_F_cross_covariance}), its expression reads
        \small
        \begin{align} 
            \mathbf{cov}[\hat{\mathbf{g}}_i,\hat{\mathbf{g}}_j] 
            \!
            & =  
            \frac{(d+2)^2}{r^2_{k_i} r^2_{k_j}}
            \mathrm{\mathbf{cov}}[\hat{\mathbf{m}}_i,\hat{\mathbf{m}}_j]
            \label{eq:deltaF_C_sample_grad_F_cross_covariance}  
            \\
            \mathbf{cov}[\hat{\mathbf{m}}_i,\hat{\mathbf{m}}_j]
            \!
            \nonumber       
            & = 
            \!
            \frac{k_{i,j} }{\tilde{k}_{i} \tilde{k}_{j}} 
            \!
            \left[
                \langle (\mathbf{x}-\mathbf{x}_{j}) (\mathbf{x}-\mathbf{x}_{j})^{\rm{T}} \rangle_{\Omega_{ij}}
                -
                \mathbf{m}_i\mathbf{m}_j^{\trans}
            \right] ,
        \end{align}
        \normalsize
        where the expectation in intersection region $\Omega_{ij}$ could theoretically be estimated using the $k_{i,j}$ points in the region.
        Unfortunately, it is common for $i$-$j$ pairs to have only a small number of common neighbours $k_{i,j}$, and this would make the estimator very noisy.
        We here exploit our geometrical intuition on the NG (see Fig.~\ref{fig:ABCD}, panels B and E) to give an empirical estimate of the Pearson correlation coefficient $p^{\bm{ij}}$ in Eq.s (\ref{eq:deltaFij_err_deltaF_ij_with_p}) and (\ref{eq:deltaFij_pearson_def}). 
        By calling $\chi^{\bm{ij}}=|p^{\bm{ij}}|$ and $s_{ij} = \mathrm{sgn}(p^{\bm{ij}})$, so that $p^{\bm{ij}} = s^{\bm{ij}}\, \chi^{\bm{ij}}$, we define the following empirical estimators
        \begin{align}
            \hat{p}^{\bm{ij}}
            &=
            \hat{s}^{\bm{ij}}
            \;
            \hat{\chi}^{\bm{ij}}
            \label{eq:deltaFij_pearson_est_def}
            \\
            \hat{s}^{\bm{ij}} 
            &=
            \mathrm{sgn}\left(\hat{\delta F}_{ij}^{\bm{i}} \; \hat{\delta F}_{ij}^{\bm{j}} \right)
            \label{eq:deltaFij_s_ij_est_def}            
            \\
            \hat{\chi}^{\bm{ij}}
            &=
            \frac{k_{i,j}}{k_{i}+k_{j}-k_{i,j}}
            \label{eq:deltaFij_chi_ij_def}
        \end{align}
        Thus, the absolute value of $p^{\bm{ij}}$, namely $\chi^{\bm{ij}}$, is estimated by the \textit{Jaccard index} \cite{jaccard1912distribution} of the two neighbourhoods $\Omega_i$ and $\Omega_j$; it is bound to be between $0$, when $\hat{g}_i$ and $\hat{g}_j$ are not correlated, i.e. $\Omega_i \cap \Omega_j = \varnothing$, and $1$, when $i=j$.
        Its sign, instead, is approximated by Eq.~(\ref{eq:deltaFij_s_ij_est_def}): this approximation, as we will see, works well in practice.
        In Sec.~\ref{sssec:deltaFs_errors_variance} and \ref{sssec:deltaFs_errors_pearson} of the SM we motivate the expressions in Eq.s (\ref{eq:deltaFij_pearson_est_def}), (\ref{eq:deltaFij_s_ij_est_def}) and (\ref{eq:deltaFij_chi_ij_def}).
        By using them in Eq. (\ref{eq:deltaFij_err_deltaF_ij_with_p}) to estimate $p^{\bm{ij}}$, we obtain an estimator for the variances of the $\delta F$'s in Eq.~(\ref{eq:deltaFij_variance}), i.e. for the diagonal elements of the covariance matrix $\mathbf{C}$.
        The great accuracy of these estimators is empirically assessed in Fig.~\ref{fig:deltaF_perform}, together with the Gaussianity and unbiasedness of the $\hat{\delta F}$ estimators in Eq.~(\ref{eq:deltaFij_estim}), required for BMTI in Eq.~(\ref{eq:BMTI_deltaF_normally_distributed}).

        One last step is represented by the estimation of generic elements of the $\delta F$'s covariance matrix, including off-diagonal terms, namely $C_{ij,lm}:=\textrm{cov}[\hat{\delta F_{ij}}\, , \,\hat{\delta F_{lm}}]$, which read
        \begin{multline}
            C_{ij,lm} = \frac{1}{4} \,
                        \mathbf{r}^{\rm{T}}_{ij}
                        \,(\,
                        \textbf{cov} [ \hat{\mathbf{g}}_i , \hat{\mathbf{g}}_l]+
                        \textbf{cov} [ \hat{\mathbf{g}}_i , \hat{\mathbf{g}}_m] \\
                        + \textbf{cov} [ \hat{\mathbf{g}}_j , \hat{\mathbf{g}}_l] 
                        + \textbf{cov} [ \hat{\mathbf{g}}_j , \hat{\mathbf{g}}_m]
                        \,
                        )
                        \,
                        \mathbf{r}_{lm} \; .
            \label{eq:deltaFij_C_ijlm_def}
        \end{multline}
        Eq.~(\ref{eq:deltaFij_C_ijlm_def}) is a general version of Eq.~(\ref{eq:deltaFij_variance}).
        Again, to obtain the general version of Eq.~(\ref{eq:deltaFij_err_deltaF_ij_with_p}), one can contract the vector differences $\mathbf{r}$ with the gradient cross-covariance matrices and define Pearson correlation coefficients.
        We report here the final expression that we propose to estimate $C_{ij,lm}$ in Eq.~(\ref{eq:deltaFij_C_ijlm_def})
        \begin{multline}
             \hat{C}_{ij,lm} =
                    \frac{1}{4}
                    \,(\,
                    \hat{p}_{ij,lm}^{\bm{il}}
                    \,\hat{\varepsilon}_{ij}^{\bm{i}} \,\hat{\varepsilon}_{lm}^{\bm{l}}
                    +
                    \hat{p}_{ij,lm}^{\bm{im}}
                    \,\hat{\varepsilon}_{ij}^{\bm{i}} \,\hat{\varepsilon}_{lm}^{\bm{m}}\\
                    +
                    \hat{p}_{ij,lm}^{\bm{jl}}
                    \,\hat{\varepsilon}_{ij}^{\bm{j}} \,\hat{\varepsilon}_{lm}^{\bm{l}}
                    +
                    \hat{p}_{ij,lm}^{\bm{jm}}
                    \,\hat{\varepsilon}_{ij}^{\bm{j}} \,\hat{\varepsilon}_{lm}^{\bm{m}}
                    \,)
                    \; ,
        \label{eq:detaFij_C_ijlm_estim}
        \end{multline} 
        where the  correlation coefficients are estimated as
        \begin{equation}
            \hat{p}_{ij,lm}^{\bm{il}}
            \,=\, 
            \mathrm{sgn}\left(\hat{\delta F}_{ij}^{\bm{i}} \; \hat{\delta F}_{lm}^{\bm{l}} \right)
            \,
            \hat{\chi}^{\bm{il}}
            \, .
        \label{eq:deltaFij_pil_ijlm_est}
        \end{equation}
        %
        A more thorough derivation and justification of Eq.s (\ref{eq:deltaFij_C_ijlm_def}), (\ref{eq:detaFij_C_ijlm_estim}) and (\ref{eq:deltaFij_pil_ijlm_est}) are contained in Sec.~\ref{sssec:deltaFs_errors_cross-covariance},\ref{sssec:deltaFs_errors_pearson} and \ref{sssec:deltaFs_errors_cross-covariance-estimator} of the SM.
        Eq (\ref{eq:detaFij_C_ijlm_estim}) and (\ref{eq:deltaFij_pil_ijlm_est}) provide a formula to estimate any element of the covariance matrix $\mathbf{C}$ defining the BMTI log-likelihood in Eq.~(\ref{eq:BMTI_loglikelihood_full}).
        Note that the expression in Eq.~(\ref{eq:deltaFij_pil_ijlm_est}) is coherent with Eq.~(\ref{eq:deltaFij_pearson_est_def}) by identifying $p^{\bm{ij}} \equiv p^{\bm{ij}}_{ij,ij}$.
        Therefore, Eq.~(\ref{eq:deltaFij_err_deltaF_ij_with_p}) is recovered from Eq.~(\ref{eq:detaFij_C_ijlm_estim}) by setting $ij=lm$, .
                \begin{figure*}[!ht]
        \centering
            \includegraphics[width=\textwidth]{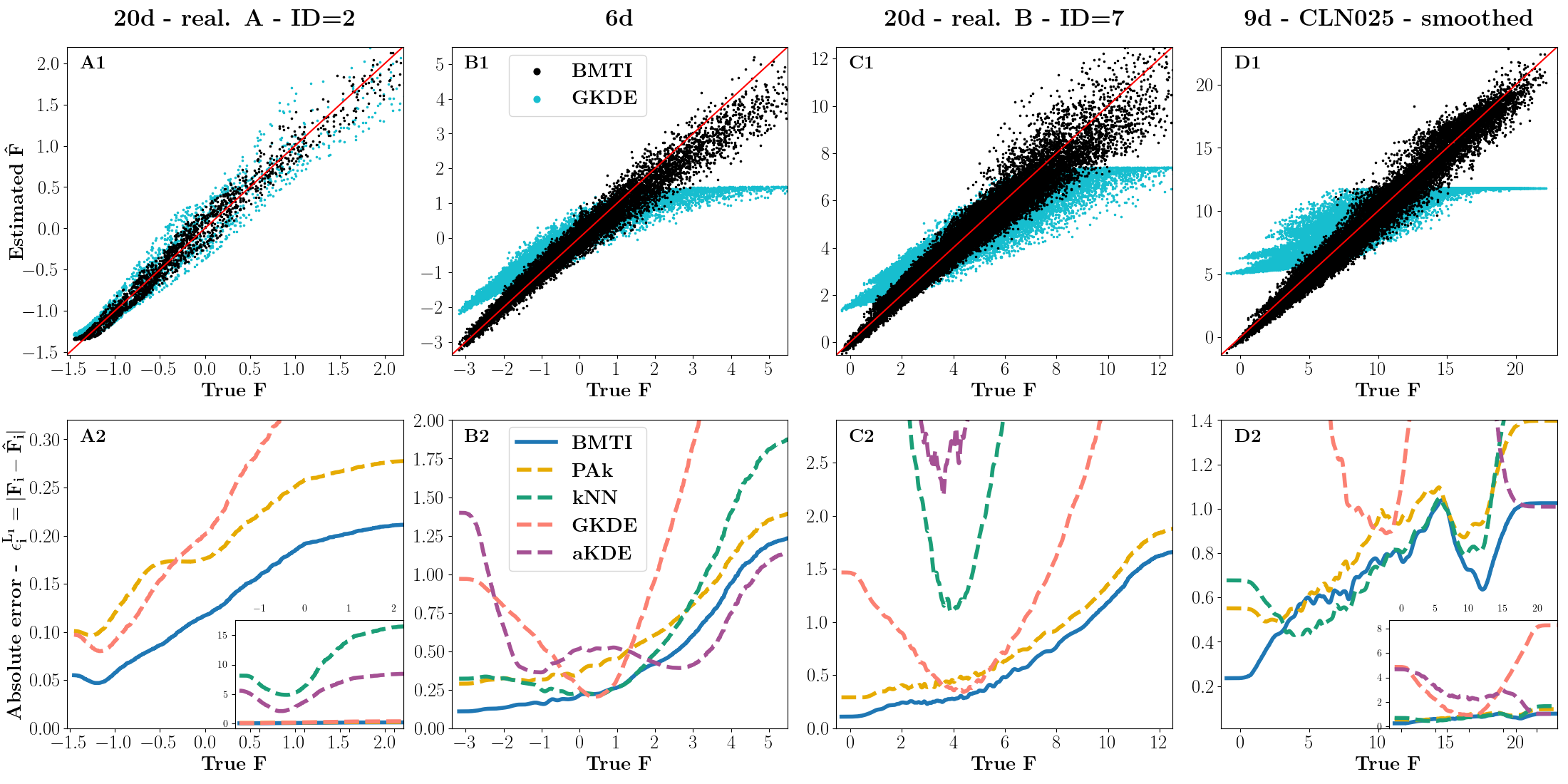}
        \vspace{-0.6cm}
        \caption
        {
        \textbf{BMTI performance on various datasets.}
        \textbf{Top}:  Scatter plots of estimated vs GT negative log-densities for BMTI and GKDE on 4 datasets of increasing intrinsic dimensionality.
        \textbf{Bottom}: Running averages of the absolute error of $\hat{F}$ as a function of the GT value of $F$ for BMTI and other baseline methods; the insets show zoomed-out versions when the error is too large to be visualised in a single graph.
        }
        \label{fig:BMTI_corrplot_MAE}
        \vspace{-0.2cm}
        \end{figure*}
    \subsection{Approximate BMTI likelihood}
    \label{ssec:BMTI_practical_sol}
    
    \vspace{-0.1cm}
    It is not the covariance matrix $\mathbf{C}$ itself that enters the definition of BMTI likelihood in Eq.~\ref{eq:BMTI_loglikelihood_full}, but rather its inverse, the precision matrix $\mathbf{C}^{-1}$.
    Since the $N_e$ elements of $\hat{\bm{\delta F}}$ are estimated based only on $N$ independent RVs (the sample points), the matrix $\mathbf{C}$ can in general be singular.
    Therefore, $\mathbf{C}^{-1}$ could be taken as a Moore-Penrose pseudoinverse. 
    Since estimating the exact pseudoinverse becomes rapidly numerically costly as the sample size increases, in practice we approximate $\mathbf{C}^{-1}$ by a diagonal matrix $\mathbf{D}$ whose non-zero elements are $D_{ij}$.
    With it the BMTI likelihood reads
    \vspace{-0.15cm}
    \begin{equation}
        \mathcal{L} ( \mathbf{F} \mid \hat{\bm{\delta F}},\mathbf{D} ) := - \sum_{i=1}^N \sum_{j\in \Omega_i}  \frac{D_{ij}}{2} (\delta F_{ij} - \hat{\delta F}_{ij})^2 .
    \label{eq:BMTI_loglikelihood_approx}
    \vspace{-0.09cm}
    \end{equation}
    %
    In Sec.~\ref{ssec:BMTI_support_approx_inverse_C} of the SM we discuss various possible approaches to define the approximate precision $\mathbf{D}$ --all of which yield accurate predictions $\hat{\mathbf{F}}$ for all tested datasets and in all sampling regimes-- and the cases in which they also allow recovering accurate uncertainty estimations. 
    Here, in the main text, we recall only the simplest of these possible choices for $\mathbf{D}$, namely $D_{ij} := 1/C_{ij,ij} = 1/{\varepsilon_{ij}}^2 $,
    with which (\ref{eq:BMTI_loglikelihood_approx}) becomes
    %
    \vspace{-0.15cm}
        \begin{equation}
            \mathcal{L} ( \mathbf{F} \mid \hat{\bm{\delta F}}\,,\, \mathbf{D} )
            :=
            - \sum_{i=1}^N \sum_{j\in \Omega_i} \frac{(F_j - F_i - \hat{\delta F}_{ij})^2}{2 \varepsilon_{ij}^2} \; .
        \label{eq:BMTI_loglikelihood_approx_gCorr}
        \end{equation}
    All the numerical experiments presented in Sec.~\ref{sec:numerical} are conducted using this setting.

    The complexity of BMTI is dominated by the solution of the linear system in Eq. \eqref{eq:BMTI_F_estimator}.
    The direct solution of a dense linear system generally requires $\mathcal{O}(N^2)$ in memory and $\mathcal{O}(N^3)$ in time  \cite{2020SciPy-NMeth}. However, since the matrix $\mathbf{A}$ is typically very sparse, we can efficiently use sparse solvers. In all cases considered, we found that an iterative sparse solver, namely \textit{conjugate gradient} \cite{Hestenes&Stiefel:1952}, drastically reduces the memory and time complexity to linear and almost-linear respectively. A time-scaling assessment of this method for increasing dataset sizes is presented in Fig.~\ref{fig:times_pak_bmti}A. In Sec.~\ref{ssec:BMTI_support_complexity_linear_solvers} of the SM we discuss the scaling and implementation details.
    %
    %
    %
    \vspace{-0.3cm}
    \subsection{Regularisation of the BMTI likelihood for disconnected neighbourhood graphs}
    \label{ssec:BMTI_likelihood_regularisation}
    BMTI integrates NLD differences on the neighbourhood graph and estimates the NLD up to a constant offset due to the PDF normalisation.
    An important shortcoming appears in cases in which, due to low sampling or to the PDF morphology, the NG on which the NLD differences are estimated becomes disconnected.
    In such a scenario, the BMTI formulation provided in Sec.~\ref{ssec:BMTI_derivation}, fails to
    correctly reconstruct the relative density of the connected subgraphs.
    Fortunately, this problem can be healed by combining the log-likelihood of a strictly-local NLD estimator with the BMTI log-likelihood via a mixing hyperparameter $\alpha$
    as a regularisation term, so that the total log-likelihood becomes    
    \begin{equation}
        \mathcal{L}^{\mathrm{tot}} (\mathbf{F} |  \hat{ \boldsymbol{\delta} \mathbf{ F}}, \mathbf{C} , \bm{\theta})
        =
        \alpha \mathcal{L}^{\mathrm{BMTI}} (\mathbf{F} |  \hat{ \boldsymbol{\delta} \mathbf{ F}}, \mathbf{C})     
        +
        (1-\alpha)
        \mathcal{L}^{\mathrm{reg}} ( \mathbf{F}| \bm{\theta} )
    \label{eq:BMTI_regularisation_BMTI+reg_likelihood}
    \end{equation}
    where $\bm{\theta}$ indicates the parameters that $\mathcal{L}^{\mathrm{reg}}$ depends on.
    As long as the regularising $\mathcal{L}^{\mathrm{reg}}$ produces normalised NLD estimates, the NLD coming from $\mathcal{L}^{\mathrm{tot}}$ will also be correctly normalised, also in the case of a disconnected NG.
    Our choice for $\mathcal{L}^{\mathrm{reg}}$ falls on a series of \textit{k}NN-based estimators which can be formulated in terms of maximum-likelihood estimators \cite{Rodriguez2018,carli2022Tesi}.
    Implementation details are discussed in the SM, Sec.~\ref{ssec:BMTI_support_BMTI_regularisation}.

    \begin{figure*}[ht]             
        \advance\leftskip-0.1cm
        \includegraphics[width=\textwidth]{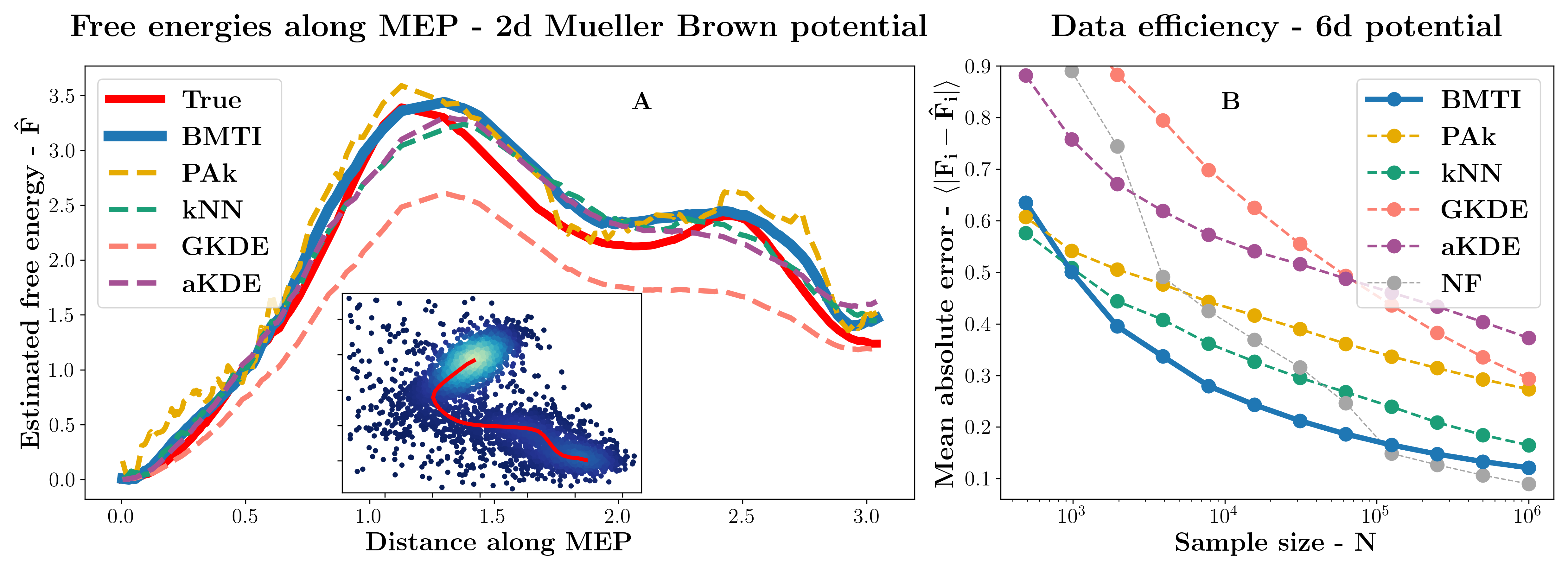}
        \vspace{-0.2cm} 
        \caption{
        \textbf{A: BMTI smoothness and accuracy}
        $\hat{F}$ along the minimum energy path connecting the two main minima of a 2d Mueller-Brown potential, with inverse temperature $\beta = 0.035$, for various methods.
        The inset depicts the dataset used in the analysis and, as a red curve, the minimum energy path.
        \textbf{B: BMTI data-efficiency}
        Mean absolute error of various nonparametric methods as a function of the number of training points for the 6-dimensional dataset. Points in the plot are computed as mean MAE over 5 different runs. The standard deviations are very small even with a few hundred points, so they are not plotted.
        }
        \label{fig:killer_graph}
        \vspace{-0.2cm}
    \end{figure*}   
\section{Numerical experiments}
\vspace{-0.2cm}
\label{sec:numerical}
    \textbf{Test datasets}~ We test the approach on various synthetic and realistic datasets, all of which are described in detail in Sec \ref{append:test_sys} of the SM. 
    The synthetic datasets are sampled from known PDFs. 
    We consider four of them, having embedding dimensions $D$ from $2$ to $9$ coincident with their ID.
    Two of them are known analytically but display to some extent a realistic behaviour.
    One is the Mueller-Brown potential \cite{Mueller1979LocationOS}, an analytic $2$-d potential designed to display the typical behaviour of metastable systems (the transitions occur through a non-trivial curved path); one is a $9$-d dataset sampled from an analytic PDF obtained via Gaussian-KDE smoothing of a realistic physical chemistry dataset, of which it retains all the complexity (in fact, we regard it as ``the hardest'' benchmark).
    The realistic datasets are all cases in which the true NLD is known only on the sample points, but the underlying PDF is not known analytically. 
    They are sampled from rugged and complex landscapes taken from the molecular simulations literature \cite{20dB,20dA,20dC}. 
    One is a $4$ dimensional dataset with ID $4$. 
    The other two have IDs $2$ and $7$ but they are both embedded in $20$ dimensions. 
    
    \textbf{Competing estimators}~
    In the case of the NLD estimators, we compare the performance of BMTI against that of other well-established nonparametric methods. 
    For the Gaussian KDE (GKDE) class, we take as baseline the standard GKDE with Silverman's smoothing parameter \cite{silverman1986density} and as state-of-the-art the \textit{adaptive kernel estimator} (aKDE) \cite{abramson1982bandwidth,abramson1982arbitrariness}, both as implemented in \cite{awkdeGitHub}.
    For the \textit{k}NN-based methods we take standard \textit{k}NN with optimal global $k$ selected via Abramson's rule of thumb ($N^{D/{D+4}}$) \cite{Abramson1984} as baseline and the \textit{point-adaptive kNN} (PA\textit{k}) as state-of-the-art, both as implemented in \cite{Glielmo2022DADApyDA}.

    \textbf{Evaluation metrics}~
    We assess the accuracy of the estimators $\hat{\mathbf{g}}$, $\hat{\delta F}$ and $\hat{F}$ by looking at various metrics here described.
    In cases where we do not know the normalized ground truth (GT) PDF, but only the unnormalised NLD $\mathbf{F}$, we align the mean of the predictions $\mathbf{\hat{F}}$ to the mean of the GT.
    The first metric is the \textit{absolute error}, or $L_1$ error, which quantifies the discrepancy of the estimator from the GT value as the $L_1$ norm.
    In the case of the NLD estimator at point $i$ it reads $\epsilon_i = | F_i - \hat{F}_i |$.
    It is studied either as a function of the GT NLD or averaged 
    on the whole dataset, the \textit{mean absolute error} (MAE) $\langle \epsilon_i \rangle$. 
    Another more qualitative but very insightful way to inspect an estimator performance is to plot the estimated estimator vs the GT value for all estimates, the \textit{parity plot}.
    Additionally, we check the joint performance of estimators $\hat{y}$ of mean value $y$ and of their uncertainty estimators $\hat{\sigma}_{y}$ by looking at the distribution of the standardised scores $(\hat{y}-y)/\hat{\sigma}_{y}$, also called the \textit{pull distribution} \cite{demortier2002everything}, which is expected to be a standard Gaussian $\mathcal{N}(0,1)$.       
    Finally, in Sec.~\ref{append:kld} of the SM we compute and compare the Kullback-Leibler (KL) divergence from the GT to the estimated densities for the various methods.

    \subsection{Performance assessment and discussion}
    \label{ssec:numerical_performance}

    \textbf{The performance of the} $\hat{\delta F}$ estimator for neighbouring points and its uncertainty is assessed in Fig.~\ref{fig:deltaF_perform} on a variety of benchmark distributions. 
    In all cases, the parity plot and the pull distribution are in very good agreement with the predictions.
    This suggests that, with the pipeline described in Sec.~\ref{ssec:BMTI_deltaF}, and the estimators in Eq.~(\ref{eq:deltaFij_estim}) provide an accurate estimate of the NLD  difference and, remarkably, of the associated error even if cases in which the intrinsic dimension is high. 
    The exceptional quality of these estimates is a necessary condition to infer the NLD globally.
    Fig.~\ref{fig:deltaF_perform} proves empirically that, with the choice for the $\hat{\bm{\delta F}}$ and $\mathbf{C}$ estimators presented in Sec.~\ref{sssec:BMTI_deltaF_delta_F} and \ref{sssec:BMTI_deltaF_C}, the Gaussianity condition in Eq.~(\ref{eq:BMTI_deltaF_normally_distributed}), required for the BMTI algorithm, is satisfied.

\begin{figure*}[ht!]
    \centering
    \includegraphics[width=0.495\textwidth]{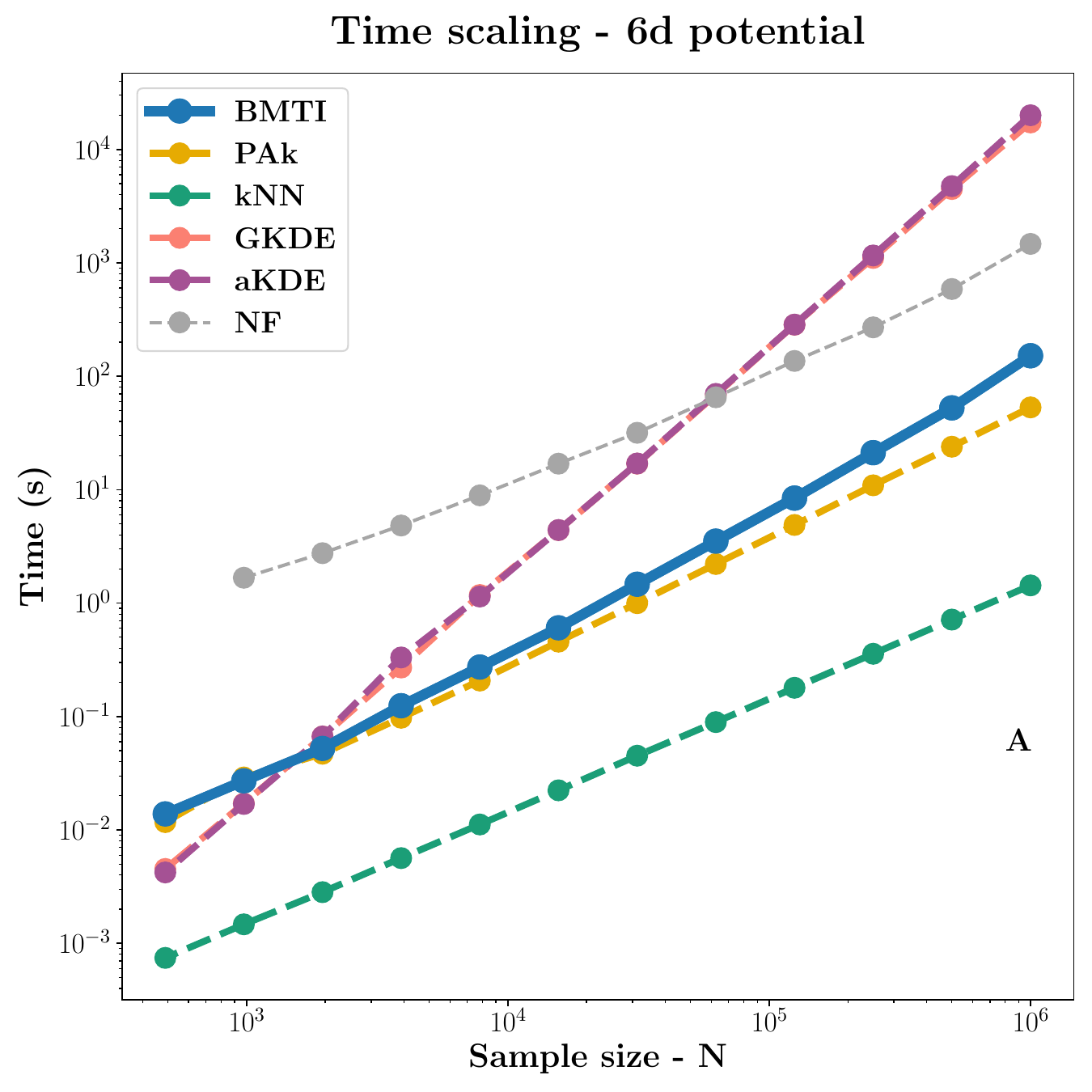}
    \hfill
    \includegraphics[width=0.495\textwidth]{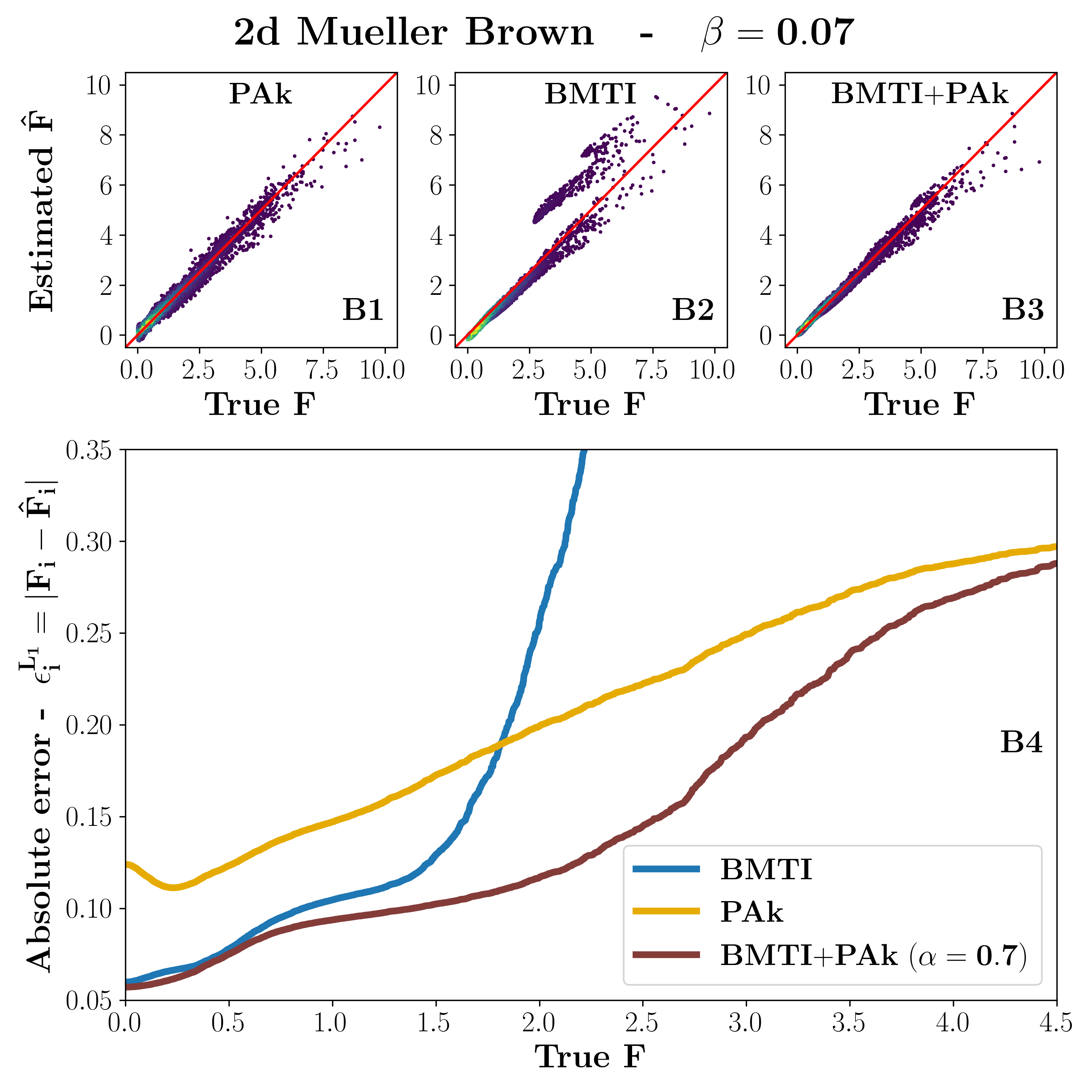}
    \vspace{-0.5cm}
    \caption{\label{fig:times_pak_bmti}
    \textbf{A: Time scaling}
    Single CPU (Intel Core i9-13900H) training times measured in seconds as a function of sample size for the 6-dimensional dataset in the case of uncorrelated $\delta F$'s corresponding to Eq.~(\ref{eq:BMTI_loglikelihood_approx_gCorr}).
    Implementation details can be found in Sec.~\ref{sssec:BMTI_support_complexity_linear_solvers_CG} of the SM.
    \textbf{B: BMTI regularisation}
    Performance of various NLD estimators on a dataset with disconnected NG.
    The dataset consists of $5,000$ points sampled from the Mueller-Brown potential presented in Sec.~\ref{sssec:test_sys_2d_MB}, but with a scaling factor, (namely, an inverse temperature) $\beta = 0.07$, which is double as the one used in Fig.~\ref{fig:killer_graph}A ($\beta = 0.035$).
    %
    %
    \textbf{Top}:  scatter plots of estimated vs GT NLDs for the PA\textit{k} (B1), BMTI (B2) and PA\textit{k}-regularised BMTI (B3) with $\alpha=0.7$ density estimators.
    \textbf{Bottom} (B4): MAE of $\hat{F}$ as a function of the GT value of $F$ for the three NLD estimators considered.}
    \vspace{-0.4cm}    
\end{figure*}
          
    \textbf{The performance of the BMTI estimator} is benchmarked in Fig.s \ref{fig:BMTI_corrplot_MAE},\ref{fig:killer_graph} and in Table \ref{tab:MAE}.
    Table \ref{tab:MAE} summarises the global performance of the estimators on various datasets. 
    BMTI is consistently the best-performing nonparametric estimator on all datasets except in the $9$-dimensional case, in which it is comparable to \textit{k}NN.

    In the top row of Fig.~\ref{fig:BMTI_corrplot_MAE}, we compare the parity plot of BMTI and GKDE, which is probably the most popular density estimation method.
    BMTI's parity plots are slimmer and more symmetric than GKDE's, a sign of both lower variance and absence of bias.
    Instead, GKDE is observed to develop systematic errors as the ID of the dataset increases, underestimating NLD differences.
    The bottom row shows that, while other methods can be particularly accurate at specific values of $F$ for specific datasets, BMTI consistently outperforms all of them across the whole $F$ spectrum except occasionally at very high values. 

    Fig.~\ref{fig:killer_graph}A illustrates that, even in undersampled regimes, the BMTI estimates are not only accurate but also smooth, two characteristics which for instance make it particularly well-suited for free energy reconstructions in physical chemistry applications. In this context, it is necessary to estimate the derivatives of thermodynamic potentials such as the free energy, so a spiky estimated free energy profile would yield unphysically high forces.
    In fact, it is among the most accurate in capturing the NLD differences between the main minimum, where the path starts, and the highest saddle point or the second-lowest minimum.
    While comparably accurate, \textit{k}NN-based methods are visibly rougher and noisier.
    On the other hand, GKDE is smooth but again, it underestimates NLD differences.
    While aKDE looks preferable to GKDE in this case, by looking at Table \ref{tab:MAE} and Fig.~\ref{fig:BMTI_corrplot_MAE} we observe that standard GKDE is more robust than aKDE when $D$ increases.
    
    %
    \newcolumntype{G}{>{\centering\columncolor{gray!20!white}}p{0.7cm}}
    \begin{table}[!b]
    \vspace{-0.5cm}
        \setlength\minrowclearance{4pt}
        \footnotesize
        \begin{tabular}{>{\centering}p{1.9cm} >{\centering\columncolor{gray!20!white}}p{0.8cm} >{\centering}p{0.7cm} >{\centering}p{0.7cm} >{\centering}p{0.8cm} >{\centering\arraybackslash}p{0.7cm}}                
            Dataset & BMTI & PA$k$ & $k$NN & GKDE & aKDE\\ 
            \hline
            2d Gaussian & \textbf{0.11} & 0.16 & 0.24 & 0.22 & \textbf{0.14}\\
            20d-A($d=2$) & \textbf{0.10} & \textbf{0.16} & 7.75 & 0.17 & 4.32\\
            2d-MBx0.035 & \textbf{0.12} & \textbf{0.16} & 0.30 & 0.36 & 0.25\\
            6d potential  & \textbf{0.26} & 0.44 & \textbf{0.37} & 0.72 & 0.53\\
            20d-B($d=7$) & \textbf{0.36} & \textbf{0.52} & 2.87 & 0.80 & 4.34\\
            9d smooth. & \textbf{0.67} & 0.79 & \textbf{0.68} & 2.43 & 2.82\\
        \end{tabular}
        \caption{
        \textbf{Performance of $\mathbf{\hat{F}}$}: MAE of various methods on 6 datasets. 
        The two best performances for each dataset are highlighted in bold.
        }
        \label{tab:MAE}
    \end{table}
    %
    
    The data-efficiency of the various methods is compared in Fig.~\ref{fig:killer_graph}B by tracking the MAE as the sample size increases on the $6$ dimensional dataset.
    As a reference to another state-of-the art method, despite not being a direct competitor, we also include a modern real NVP normalising flows (NF) model \cite{dinh2016NormalizingFlows} as implemented in \cite{NFgithub}. We train the NF for 20 epochs, so that it approaches convergence; we did not push the number of epochs further because, already with this choice, the computation times for NFs are one-to-two orders of magnitude larger than for BMTI.
    BMTI is the best performing of all the competing methods, maintaining a low MAE for all sample sizes. 
    While BMTI shows a better performance in an undersampling regime, the big NF model outperforms BMTI for very large samples. 
    
    By looking at all the benchmarks, we observe that the methods restricting to the intrinsic manifold (BMTI and PA\textit{k}) are the only ones consistently among top performers.
    Standard \textit{k}NN is competitive only in cases in which the ID is close or equal to $D$, i.e. precisely when it computes volumes on the intrinsic manifold.
    All other methods break down as $D$ increases. The only exception is GKDE in the 20d-A potential, where it curiously seems to implicitly restrict to the intrinsic manifold despite not having any notion of it.
    
    Finally, Fig.~\ref{fig:times_pak_bmti}A compares the time scaling as a function of the number of points $N$ for all the various algorithms.
    BMTI uses the conjugate gradient sparse iterative solver, which shows a slight deviation from linear time scaling (as discussed in Sec.~\ref{ssec:BMTI_support_complexity_linear_solvers} of the SM). The GKDEs scale quadratically, while all the other kernel methods scale linearly --although PA$k$ with a prefactor at least one order of magnitude larger than standard $k$NN.
    Finally, training NFs has a linear cost but is computationally one-to-two orders of magnitude slower than BMTI.

    By analysing this in conjunction with the other performance metrics, we can conclude that BMTI stands out as a method of choice:
    \begin{itemize}
        \item[(i)] for datasets up to moderate sizes of $\mathcal{O}(10^5)$ points, but can be preferred to deep neural network estimators even for larger sizes, when time efficiency is important;
        \item[(ii)] for multimodal densities (since in trivial cases simple parametric models
        might be a better choice);
        \item[(iii)] when a quantitative control over the accuracy and smoothness on the log-density is deemed important (this is the case, for example, in many physical chemistry or ML applications).
    \end{itemize}

    \textbf{The performance of a regularised BMTI estimator} is benchmarked in Fig.~\ref{fig:times_pak_bmti}B for the case in which the regularising log-likelihood in Eq.~(\ref{eq:BMTI_regularisation_BMTI+reg_likelihood}) is that of the $k$NN-based PA\textit{k} estimator from Ref. \cite{Rodriguez2018}.
    The benchmark dataset is obtained from the Mueller-Brown potential with an inverse temperature $\beta = 0.07$, which is double w.r.t. the one used in Fig.~\ref{fig:killer_graph}A, $\beta = 0.035$.
    In this case, the three main wells of the potential are sampled separately, since the saddle points on the transition paths between the basins are not populated due to the small size of the sample and the low temperature.
    A strictly-local method, such as the PA\textit{k} estimator, is not affected by the disconnectedness of the NG, since for each point the NLD estimate $\hat{F}_i$ is independent; therefore the parity plot results in a single connected cloud.
    The parity plot of the BMTI estimator, which is global, presents instead three disconnected clouds (Fig.~\ref{fig:times_pak_bmti}.B2).
    By mixing the likelihoods for the two estimators as in Eq.~(\ref{eq:BMTI_regularisation_BMTI+reg_likelihood}), the parity plot becomes unimodal (Fig.~\ref{fig:times_pak_bmti}.B3), and it is actually even slimmer than the case of PA$k$ alone, especially for high density values.
    This improved performance is even clearer by looking at panel B4 of the same figure, in which the MAE is plotted as a function of the ground truth NLD: the regularised BMTI estimator outperforms PA$k$ basically across the whole range.
    Note that even very small values of the mixing scalar $\alpha$ in Eq.~(\ref{eq:BMTI_regularisation_BMTI+reg_likelihood}), such as $10^{-2}$, or even $10^{-3}$, are enough to regularise the BMTI estimator and heal it from the disconnectedness.
    Moreover, this regularisation corrects the singularity of the matrix $\mathbf{A}$ in Eq.~(\ref{eq:BMTI_F_estimator}), as discussed in Sec.~\ref{sssec:BMTI_support_complexity_linear_solvers_CG} of the SM.
    \vspace{-0.3cm}

\section{Conclusions}
\label{sec:conclusions}
\vspace{-0.3cm}
We have presented BMTI, a nonparametric data-efficient method to estimate smooth log-density landscapes even in high dimensional spaces.
BMTI is based on a first estimation of log-density differences between neighbouring points, and a subsequent (implicit) integration of such differences.
Its key ingredients are the possibility of restricting its operation to the intrinsic data manifold, its point-adaptive nature and a rigorous error control. 
Such features make BMTI accurate and efficient, as we demonstrate through numerical experiments.
These characteristics make it suitable candidates for physical or ML applications \cite{du2019implicit,Henin2022EnhancedSM}.
Moreover, we stress that this work also introduces an adaptive nonparametric log-density gradient estimator and a graph-based integration procedure, which can be seen as interesting stand-alone methods.

Importantly, BMTI can be made robust against cases in which the NG is disconnected, a situation in which thermodynamic integration schemes would be doomed to fail.
To avoid this pitfall, the BMTI log-likelihood can be mixed additively with the log-likelihood of a strictly-local and normalised NLD estimator.

Finally, the BMTI framework also naturally incorporates a way to estimate the log-density uncertainties.
The theoretical derivation is rigorous under the assumption of Gaussian noise (Eq. \ref{eq:BMTI_deltaF_normally_distributed}), but the numerical implementation is approximate (we estimate the inverse of the covariance matrix assuming the inverse is diagonal-dominated).
Much of our current effort is in the direction of finding better approximations and numerical solutions to estimate this inverse.

\subsubsection*{Code and data availability statement}
Our code for the BMTI, $k$NN and PA$k$ density estimators is openly available on DADApy~\cite{Glielmo2022DADApyDA} at \url{https://github.com/sissa-data-science/DADApy}.

All datasets that support the findings of this study, described in Sec. \ref{append:test_sys} of the SM, are openly available on Zenodo at \url{https://doi.org/10.5281/zenodo.15002318} \cite{Carli2025Datasets}.

\subsubsection*{Acknowledgements}

This work is supported in part by funds from the European Union’s Horizon 2020 research and innovation program (grant number 824143, MaX ’Materials design at the eXascale’ Centre of Excellence) and by NextGenerationEU through the Italian National Centre for HPC, Big Data, and Quantum Computing (grant number CN00000013). 
The authors would like to thank S. De Gironcoli (SISSA), J. Henin (CNRS), F. Marinelli (MCW), T. D. Swinburne (CNRS), M. C. Marinica (CEA), F. Pellegrini (SISSA), Claudia Biancotti (BdI), Lorenzo Fant (OGS), Laura Zichi (Harvard), Iuri Macocco (UPF) and Claudio Leone (ICTP) for helpful discussions and feedback.
The authors are grateful to one of the anonymous reviewers for their important suggestion to explore the use of efficient solvers for the BMTI linear system.

The views and opinions expressed in this paper are those of the authors and do not necessarily reflect the official policy or position of Banca d’Italia.

\bibliography{Bibliography}

\onecolumn

\addcontentsline{toc}{section}{Supplementary Material} 
\part{Supplementary Material} 
\parttoc 

\newpage

\appendix

\renewcommand{\theequation}{S.\arabic{equation}}
    
\section{Mean shift gradient estimator and related uncertainty estimators}
\label{append:mean_shift}

    \subsection{Mean shift and log-density gradient}
    \label{ssec:mean_shift_mean_shift_grad_F}

        \subsubsection{Recapitulation of useful Euclidean integrals over the $n$-sphere}
        \label{sssec:mean_shift_mean_shift_grad_F_useful_integrals}
        We present these results for an Euclidean space, but they can be generalised to other metrics \cite{Fukunaga1990}.
        
        We express the volume of a $n$-dimensional sphere of radius $r$, $B^n(r)$, as $V_n = \omega_n r^n $. Therefore, its surface is $S_n = \partial_r V_n = n \omega_n r^{n-1}$. The quantity $\omega_n$ is the volume of the $n$-sphere of unitary radius, whose expression can be derived by computing $\omega_n := \int_{B^n(r)} 1\,\mathrm{d} \mathbf{x}$, which gives $\omega_n = \frac{2}{n} \, \pi^{\frac{n}{2}} / \, \Gamma(\frac{n}{2})$ .

        Now, by taking the mean outer product of $\mathbf{x}$ over the ball $B^n(r)$ and calling $\mathbb{1}_n$ the $n$-dimensional identity matrix:
        \[
            V_n \, \langle \mathbf{x} \mathbf{x}^{\trans} \rangle_{B^n(r)} =  \int_{B^n(r)} \mathbf{x} \mathbf{x}^{\trans} \, \mathrm{d} \mathbf{x} = \mathbb{1}_n \, \frac{r^2}{n+2}  V_n ~~\Rightarrow~~\langle \mathbf{x} \mathbf{x}^{\trans} \rangle_{B^n} = \mathbb{1}_n \, \frac{r^2}{n+2}
        \]
        and thus the mean square displacement over the ball $B^n(r)$ is
        \[
            \langle \mathbf{x}^2 \rangle_{B^n(r)} = \frac{1}{V_n} \int_{B^n(r)} \mathbf{x}^2 \, \mathrm{d} \mathbf{x} = 
            \langle \Tr (\mathbf{x} \mathbf{x}^{\trans}) \rangle_{B^n(r)} = \Tr ( \langle \mathbf{x} \mathbf{x}^{\trans} \rangle_{B^n} ) = \Tr ( \mathbb{1}_n ) \frac{r^2}{n+2} = r^2 \, \frac{n}{n+2}
        \]
        \subsubsection{Relation between the mean shift and the PDF}
        \label{sssec:mean_shift_mean_shift_grad_F_mean_shift_anal}
        Let us first consider a distribution $\tilde{\rho}$ varying linearly along a direction (indicated by its gradient) in a given region of configuration space $\Omega_i$ centred around point $\mathbf{x}_i$. For any point $\mathbf{x}$ in $\Omega_i$:        
        \begin{equation}
        \tilde{\rho}(\mathbf{x}) = \tilde{\rho}(\mathbf{x}_i) + \nabla_{\mathbf{x}}\tilde{\rho}(\mathbf{x})|_{\mathbf{x}_i} (\mathbf{x}-\mathbf{x}_i)\ .
        \label{eq:mean_shift_linear_rho}
        \end{equation}
        In these conditions the gradient of the density is proportional to the mean shift around the central point
        \begin{equation} 
        \nabla_{\mathbf{x}}\tilde{\rho}(\mathbf{x}_i) := 
        \nabla_{\mathbf{x}}\tilde{\rho}(\mathbf{x})|_{\mathbf{x}_i} \propto 
        \langle (\mathbf{x} - \mathbf{x}_i) \rangle_{\tilde{\rho}} = 
        \frac{\int \tilde{\rho}(\mathbf{x}) (\mathbf{x} - \mathbf{x}_i) \, \mathrm{d}\mathbf{x}}{\int \tilde{\rho}(\mathbf{x})\, \mathrm{d}\mathbf{x}} .
        \label{eq:beyond_pak_grad_rho_prop_to_mean_shift}
        \end{equation}
        We now show how accurate is the approximation (\ref{eq:beyond_pak_grad_rho_prop_to_mean_shift}) for a generic PDF, in which also quadratic or terms are present. Let us consider the Taylor expansion of a density $\rho(\mathbf{x})$ around a point $\mathbf{x}_i$: 
        \begin{equation}
        \rho(\mathbf{x}) = \rho(\mathbf{x}_i) \,+\,
        \nabla^{\trans}_{\mathbf{x}}\rho(\mathbf{x}_i) (\mathbf{x}-\mathbf{x}_i) \,+\,
        \frac{1}{2} (\mathbf{x}-\mathbf{x}_i)^{\trans}\, \nabla^2_{\mathbf{x}}\,\rho(\mathbf{x}_i)  (\mathbf{x}-\mathbf{x}_i) + \mathcal{O}\left((\mathbf{x}-\mathbf{x}_i)^3\right)\ .
        \label{eq:mean_shift_Taylor_exp_rho}
        \end{equation}
        For a lighter notation we choose the specific case $\mathbf{x}_i = \bm{0}$, but the derivation remains valid also in the more general case. Inserting the expansion (\ref{eq:mean_shift_Taylor_exp_rho}) into Eq.~(\ref{eq:grad_F_mean_shift_definition}) and taking into account the results in Sec.~\ref{sssec:mean_shift_mean_shift_grad_F_useful_integrals}:
        
        \begin{equation}
        \begin{split}
        \langle (\mathbf{x} - \mathbf{x}_i) \rangle_{\Omega_i,\rho} 
        &= \frac{\int_{\Omega_i} \rho(\mathbf{x})\, \mathbf{x}  \, \mathrm{d}\mathbf{x}}{\int_{\Omega_i} \rho(\mathbf{x})\, \mathrm{d}\mathbf{x}}\\
        &= \frac{
        \rho(\mathbf{x}_i) \cancelto{0}{\int_{\Omega_i} \mathbf{x}  \, \mathrm{d}\mathbf{x}} ~~+\;
        \nabla^{\trans}_{\mathbf{x}}\rho(\mathbf{x}_i) \int_{\Omega_i} \mathbf{x}\,\mathbf{x}^{\trans} \, \mathrm{d}\mathbf{x} \;+\;
        \frac{1}{2} \, \nabla^2_{\mathbf{x}}\,\rho(\mathbf{x}_i) \cancelto{0}{\int_{\Omega_i} \mathbf{x}\,\mathbf{x}^{\trans} \mathbf{x} \, \mathrm{d}\mathbf{x}}
        }{
        \rho(\mathbf{x}_i) \int_{\Omega_i} 1 \,\mathrm{d}\mathbf{x} \;+\;
        \nabla^{\trans}_{\mathbf{x}}\rho(\mathbf{x}_i) \cancelto{0}{\int_{\Omega_i} \mathbf{x}\, \mathrm{d}\mathbf{x}} ~~+\;
        \frac{1}{2} \, \Tr \left[ \nabla^2_{\mathbf{x}}\,\rho(\mathbf{x}_i) \int_{\Omega_i} \mathbf{x}\,\mathbf{x}^{\trans} \mathrm{d}\mathbf{x}\right]
        } \,+\,\mathcal{O}(V_d \,r_i^4)\\
        &= \frac{
        \nabla_{\mathbf{x}}\rho(\mathbf{x}_i) \,\cancel{V_d} \,\frac{r_i^2}{d+2}
        }{
        \rho(\mathbf{x}_i) \, \cancel{V_d} \;+\;
        \frac{1}{2} \, \Tr \nabla^2_{\mathbf{x}}\,\rho(\mathbf{x}_i) \,\cancel{V_d} \,\frac{r_i^2}{d+2}
        } \,+\,\mathcal{O}(\cancel{V_d} \,r_i^4)\\
        &=\frac{
        \nabla_{\mathbf{x}}\rho(\mathbf{x}_i) \,\frac{r_i^2}{d+2}
        }{
        \rho(\mathbf{x}_i) \, \left( 1 \;+\;
        \frac{\Tr \nabla^2_{\mathbf{x}}\,\rho(\mathbf{x}_i)}{2\,\rho(\mathbf{x}_i)} \,\frac{r_i^2}{d+2} \right)
        } \,+\,\mathcal{O}(r_i^4)\\
        &= \frac{r_i^2}{d+2} \frac{\nabla_{\mathbf{x}}\rho(\mathbf{x}_i)}{\rho(\mathbf{x}_i)} \left(1-\frac{\mathrm{Tr}\nabla^2_{\mathbf{x}}\rho(\mathbf{x}_i)}{2 \rho(\mathbf{x}_i)} \frac{r_i^2}{d+2} \right) \,+ \,\mathcal{O}(r^4_i), 
        \end{split}
        \label{eq:mean_shift_mean_shift_calculation}
        \end{equation}
        
        where the neglected integrals vanish for integration of an odd function on a symmetric domain. If curvature effects are negligible, i.e. if the correction term in the last line vanishes, then the approximation in Eq.~(\ref{eq:grad_F_grad_F_analytical_mean_shift}) is well justified. Eq.~(\ref{eq:mean_shift_mean_shift_calculation}) also quantifies the order of the approximation made in \cite{Fukunaga1990} when stating the result in Eq.~(\ref{eq:grad_F_grad_F_analytical_mean_shift}).
        
        \subsubsection{Operational definition of sample mean shift}
        \label{sssec:mean_shift_mean_shift_grad_F_mean_shift_sample_estimator}
        We want to give a sample estimate of the mean shift in Eq.~(\ref{eq:grad_F_mean_shift_definition}), Sec.~\ref{sssec:BMTI_deltaF_grad_F}. We replace $\rho$ by a KDE $\hat{\rho}$. We can obtain a flat kernel \cite{Fix1951DiscriminatoryProperties,Cacoullos1964} by combining the sample density estimator $\hat{\rho}_s$ \cite{Fukunaga1990}:
        \begin{equation}
        \hat{\rho}_s(x) = \frac{1}{N} \, \sum_{j=1}^N \, \delta(\mathbf{x}_j-\mathbf{x})
        \label{eq:mean_shift_sample_density}
        \end{equation}
        with a restriction on the $d$-ball region  $\Omega_i = B^d(h,\mathbf{x}_i)$ of radius $h$ centered on $\mathbf{x}_i$, which contains $k_h$ points. The resulting expression for the sample mean shift over $\Omega_i$ is
        \begin{equation}
        \begin{split}
        \langle (\mathbf{x} - \mathbf{x}_i) \rangle_{B^d(h,\mathbf{x}_i),\hat{\rho}_s} \,&=\, \frac{
        \int_{B^d(h,\mathbf{x}_i)}  \hat{\rho}_s(\mathbf{x}) (\mathbf{x} - \mathbf{x}_i) \, \mathrm{d}\mathbf{x}
        }{
        \int_{B^d(h,\mathbf{x}_i)}  \hat{\rho}_s(\mathbf{x}) \, \mathrm{d}\mathbf{x}
        }\\
        &=\, \frac{
        \frac{1}{N} \sum_{j=1}^{N} \int_{B^d(h,\mathbf{x}_i)}  \delta(\mathbf{x}_j-\mathbf{x})\, (\mathbf{x} - \mathbf{x}_i) \, \mathrm{d}\mathbf{x}
        }{
        k_h/N
        }\\
        &=\, \frac{1}{k_h} \sum_{j=1}^{N} \int  I_{B^d(h,\mathbf{x}_i)}\, \delta(\mathbf{x}_j-\mathbf{x})\, (\mathbf{x} - \mathbf{x}_i) \, \mathrm{d}\mathbf{x}\\
        &=\, \frac{1}{k_h} \, I_{B^d(h,\mathbf{x}_i)}\sum_{j=1}^{N} (\mathbf{x}_j - \mathbf{x}_i)\\
        &=\, \frac{1}{k_h} \sum_{j=1}^{k_h} (\mathbf{x}_j - \mathbf{x}_i)~,
        \end{split}
        \label{eq:mean_shift_sample_mean_shift}
        \end{equation}
        where $I_{B^d(h,\mathbf{x}_i)}$ is the indicator function of the selected neighbourhood $\Omega_i$ of point $\mathbf{x}_i$. If $\mathbf{x}_i$ is a point of the dataset, the unbiased estimator of the mean shift should have a scaling factor $k_h/(k_h-1)$ with respect to the final expression in Eq.~(\ref{eq:mean_shift_sample_mean_shift}). If instead of a fixed bandwidth $h$ for the uniform kernel we choose the $\hat{k}$NN framework, then $\Omega_i$ would have a radius corresponding to the distance $r_{k_i}$ from $\mathbf{x}_i$ to its $k_i$-th neighbour and the sample mean shift estimator on a point $i$ of the dataset becomes:
        \begin{equation}
        \hat{\mathbf{m}}_i \,:=\,
        \langle (\mathbf{x} - \mathbf{x}_i) \rangle_{B^d(r_{k_i},\mathbf{x}_i),\hat{\rho}_s} 
        = \frac{1}{k_i-1} \sum_{j=1}^{k_i-1} (\mathbf{x}_j - \mathbf{x}_i)~,
        \label{eq:mean_shift_sample_mean_shift_kNN}    
        \end{equation}
        which is the definition implicitly appearing in Eq.~(\ref{eq:grad_F_sample_grad_F}).
        Note that the derivation of Eq.~(\ref{eq:mean_shift_sample_mean_shift})
        is valid for a generic uniform kernel~\cite{Cacoullos1964} and not only for the $\hat{k}$NN. 
        Estimating the mean shift on the right-hand side of Eq.~(\ref{eq:grad_F_grad_F_analytical_mean_shift}) by Eq.~(\ref{eq:mean_shift_sample_mean_shift}) one obtains the sample gradient estimator $\hat{\mathbf{g}}_i$ in Eq.~(\ref{eq:grad_F_sample_grad_F}).
        As discussed in the next section, Sec.~\ref{sssec:mean_shift_mean_shift_grad_F_Fukunaga_derivation}, an estimator similar to $\hat{\mathbf{g}}_i$ was first proposed in Ref.~\cite{Fukunaga1975} using both the uniform kernel and its shadow, the Epanechnikov kernel as density estimators. 
        To the best of our knowledge, ours is the first explicit and rigorous derivation of the expression in Eq.~(\ref{eq:grad_F_sample_grad_F}) in the case of a uniform kernel in literature.

        \subsubsection{Discussion on the original derivation of the mean shift gradient estimator}
        \label{sssec:mean_shift_mean_shift_grad_F_Fukunaga_derivation}
        In the original paper in which the mean shift gradient estimator $\hat{\mathbf{g}}_i$ defined in Eq.~(\ref{eq:grad_F_sample_grad_F}) was introduced \cite{Fukunaga1975}, the authors derive an expression for the density gradient by computing the gradient of a multidimensional KDE for a generic kernel shape (F75.1). Then, they substitute into the generic expression (F75.9) the explicit form of the Epanechnikov kernel (F75.33) obtaining (F75.35). In this latter expression, they factor out the sample mean shift estimator (F75.38) and the expression for the uniform kernel density estimate (F75.39). Finally, in Eq.~(F75.41), they define, implicitly, an estimator for $\nabla_{\mathbf{x}}\rho(\mathbf{x}) / \rho(\mathbf{x})$ by estimating the numerator with the Epanechnikov KDE and the denominator with the flat KDE, without justifying this choice. Ref. \cite{Cheng1995MeanSM} was the firs one to introduce the concept of shadow kernels. Ref. \cite{Comaniciu2002} generalises the estimator given in \cite{Fukunaga1975} to any well-behaved kernel and provides a more rigorous expression in terms of a kernel and its shadow.

        \subsubsection{Covariance structure of the gradient estimators}
        \label{sssec:grad_F_grad_F_mean_shift_grad_F_cov_structure}
        
            \paragraph{Variance-covariance matrix of the gradients}
            \label{par:grad_F_grad_F_mean_shift_grad_F_cov_structure_autocovariance}
            The estimator $\hat{\mathbf{m}_i}$ in Eq.~(\ref{eq:grad_F_sample_grad_F}) is the sample average of a set of i.i.d random variables $\left\{- \frac{d+2}{r_{k_i}^2}\,(\mathbf{x}_j-\mathbf{x}_i)\right\}_{j=1}^{k_i-1}$, whose mean value, due to the proven unbiasedness \cite{Fukunaga1975}, is the actual negative score $\nabla_{\mathbf{x}}F(\mathbf{x}_i) =: \mathbf{g}_i = \langle \hat{\mathbf{g}}_i \rangle$.
            From the central limit theorem we know that the distribution of $\hat{\mathbf{g}}_i = - \frac{d+2}{r_{k_i}^2} \, \hat{\mathbf{m}}_i $ is well approximated by a $D$-variate normal whose variance-covariance matrix proportional to the variance-covariance matrix of $(\mathbf{x} - \mathbf{x}_i ) |_{\mathbf{x} \in \Omega_i} $ and can be estimated by Eq.~(\ref{eq:grad_F_grad_F_sample_autocovariance}) using 
            \begin{equation}
            \mathrm{\mathbf{var}}[\hat{\mathbf{m}}_i] 
            \,=\,            
            \mathrm{\mathbf{cov}}[\hat{\mathbf{m}}_i,\hat{\mathbf{m}}_i] 
            \,=\, 
            \frac{1}{k_i-1}\,
            \mathbf{var}[(\mathbf{x}-\mathbf{x}_{i})]_{\Omega_i}
            \, .
            \label{eq:mean_shift_mean_shift_m_autocovariance}
            \end{equation}

            Note that in a single sample we observe only a single realisation of the RV $\hat{\mathbf{g}}_i$ for any point $i$.
            For a generic RV, this would make it impossible to compute any statistic.
            However, the shift random variable $(\mathbf{x} - \mathbf{x}_i )$ is observed $k_i-1$ times, so we can estimate $\mathrm{\mathbf{cov}}[(\mathbf{x}-\mathbf{x}_{i}),(\mathbf{x}-\mathbf{x}_{i})]_{\Omega_i} =: \mathrm{\mathbf{cov}}[\hat{\mathbf{m}}_i,\hat{\mathbf{m}}_i]$ from the sample.
            Thus, $\mathrm{\mathbf{var}}[\hat{\mathbf{g}}_i]$ can also be estimated from the sample, substituting the mean values in the equations with sample averages over the $k_i$ points in $\Omega_i$.
            In particular, $\langle (\mathbf{x}-\mathbf{x}_{1}) \rangle_{\Omega_i}$ is estimated by $\hat{\mathbf{m}}_i$ in Eq.~(\ref{eq:mean_shift_sample_mean_shift_kNN}), while $\langle (\mathbf{x}-\mathbf{x}_{1}) (\mathbf{x}-\mathbf{x}_{1})^{\trans} \rangle_{\Omega_i}$ is estimated from the neighbours $\mathbf{x}_j$ of $\mathbf{x}_i$ as $\frac{1}{k_{i}-1} \sum_{j=1}^{k_i-1} (\mathbf{x}_j-\mathbf{x}_{1}) (\mathbf{x}_j-\mathbf{x}_{1})^{\trans}$.
            Thus, for a given point $\mathbf{x}_i$ the variance-covariance matrix $\mathrm{\mathbf{var}}[\hat{\mathbf{g}}_i]$ of the NLD gradient is estimated taking into account Bessel's correction for the unbiased sample variance estimator \cite{Upton2008ADictionaryofStatistics}, as
            \begin{equation}
                \hat{\mathrm{\mathbf{var}}}[\hat{\mathbf{g}}_i]
                =
                \left( \frac{d+2}{r_{k_i}^{\!2}} \right)^2 
                \hat{\mathrm{\mathbf{var}}}[\hat{\mathbf{m}}_i]
                =
                \frac{1}{k_i-2}
                    \left( \frac{d+2}{r_{k_i}^2} \right)^{\!2}
                    \left(
                         \sum_{j \in \Omega_i} \frac{(\mathbf{x}_j-\mathbf{x}_i) (\mathbf{x}_j-\mathbf{x}_i)^{\trans}}{k_i-1}  
                         - \hat{\mathbf{m}}_i\hat{\mathbf{m}}_i^{\trans}
                    \right)
            \label{eq:mean_shift_mean_shift_grad_F_sample_autocovariance}
            \end{equation}
            %
            The variance on a single gradient estimator component $\hat{g}_{i,\alpha}$ is simply the marginal of $\hat{\mathrm{\mathbf{var}}}[\hat{\mathbf{g}}_i]$ over the component $\alpha$ \cite{eaton1983multivariate}, so it can be estimated as:
            \begin{equation*}
                   \hat{\mathrm{var}}[\hat{g}_{i,\alpha}] = \sqrt{(\hat{\mathrm{var}}[\hat{\mathbf{g}}_i])_{\alpha \alpha}} 
            \end{equation*}
        
            \paragraph{Cross-covariance matrix of the NLD gradients}
            \label{par:grad_F_grad_F_mean_shift_grad_F_cov_structure_cross_cov}
            Let us consider the sample means shift estimator in Eq.~(\ref{eq:mean_shift_sample_mean_shift_kNN}) evaluated at two different points of the dataset $\mathbf{x}_1$ and $\mathbf{x}_2$.
            According to the notation introduced in Sec.~\ref{sssec:mean_shift_mean_shift_grad_F_mean_shift_sample_estimator} of the SM, we shall call $\Omega_i = B^d(r_{k_i},\mathbf{x}_i)$, so $I_{\Omega_1}$ and $I_{\Omega_2}$ are the indicator functions over the selected neighbourhoods of the two points.
            Let us also define the intersection between the two neighbourhoods $\Omega_{1,2} := \Omega_1 \,\cap\, \Omega_2$ and the number of sample points contained in it: $k_{1,2} := N \int_{\Omega_{1,2}} \hat{\rho}_s(x) \dx$, where $\hat{\rho}_s$ is the sample density estimator in Eq.~(\ref{eq:mean_shift_sample_density}).
            Finally, let us simplify the notation by relabelling $\tilde{k}_i := k_i-1$ and let us indicate the expected value of the mean shift estimator, i.e. the analytical mean shift defined in Eq.~(\ref{eq:grad_F_mean_shift_definition}), by $\mathbf{m}_i := \langle\hat{\mathbf{m}}_i\rangle$.
            Thus, referring also to Eq.~(\ref{eq:mean_shift_sample_mean_shift}), we can compute the cross-covariance matrix of the two sample mean shift estimators:
            \begin{equation}
            \begin{split}
            \mathrm{\mathbf{cov}}[\hat{\mathbf{m}}_1,\hat{\mathbf{m}}_2]
            \,&=\, 
            \langle \hat{\mathbf{m}}_1 \hat{\mathbf{m}}_2^{\trans}\rangle
            - \langle \hat{\mathbf{m}}_1\rangle \langle\hat{\mathbf{m}}_2^{\trans}\rangle\\
            &= \, \bigg\langle
            \Big[\frac{1}{\tilde{k}_{1}}\sum_{i=1}^{N}
                I_{\Omega_1}(\mathbf{x}_{i}-\mathbf{x}_{1})\Big]
            \Big[\frac{1}{k_{2}}\sum_{j=1}^{N}
                I_{\Omega_2}(\mathbf{x}_{j}-\mathbf{x}_{2})^{\trans}\Big]
            \bigg\rangle - \mathbf{m}_1\mathbf{m}_2^{\trans}\\
            &= \, \frac{1}{\tilde{k}_{1}\tilde{k}_{2}}\sum_{i,j}^{N,N} \left\langle
                I_{\Omega_1}(\mathbf{x}_{i}-\mathbf{x}_{1})\,
                I_{\Omega_2}(\mathbf{x}_{j}-\mathbf{x}_{2})^{\trans}
            \right\rangle
            - \mathbf{m}_1\mathbf{m}_2^{\trans}\\
            &= \, \frac{1}{\tilde{k}_{1}\tilde{k}_{2}} \left[ \sum_{i}^{N}
            \left\langle
                I_{\Omega_1}(\mathbf{x}_{i}-\mathbf{x}_{1})\,
                I_{\Omega_2}(\mathbf{x}_{i}-\mathbf{x}_{2})^{\trans}
            \right\rangle
            +
            \sum_{i \neq j}^{N(N-1)}
            \left\langle
                I_{\Omega_1}(\mathbf{x}_{i}-\mathbf{x}_{1})
            \right\rangle \, \left\langle
                I_{\Omega_2}(\mathbf{x}_{j}-\mathbf{x}_{2})^{\trans}
            \right\rangle
            \right]
            - \mathbf{m}_1\mathbf{m}_2^{\trans}\\
            &= \frac{1}{\tilde{k}_{1}\tilde{k}_{2}} \left[
            \,k_{1,2} \, 
            \langle(\mathbf{x}-\mathbf{x}_{1})(\mathbf{x}-\mathbf{x}_{2})^{\trans} \rangle_{\Omega_{1,2}} 
            \,+ \,(\tilde{k}_1 \tilde{k}_2 - k_{1,2}) 
            \langle(\mathbf{x}-\mathbf{x}_{1})\rangle_{\Omega_{1}} \langle(\mathbf{x}-\mathbf{x}_{2})^{\trans}\rangle_{\Omega_{2}}
            \, \right]
            - \mathbf{m}_1\mathbf{m}_2^{\trans}\\
            &= \frac{1}{\tilde{k}_{1}\tilde{k}_{2}} \left[
            \,k_{1,2} \, 
            \langle(\mathbf{x}-\mathbf{x}_{1})(\mathbf{x}-\mathbf{x}_{2})^{\trans}\rangle_{\Omega_{1,2}} 
            \,+ \,(\tilde{k}_1 \tilde{k}_2 - k_{1,2}) \,
            \mathbf{m}_1\mathbf{m}_2^{\trans}
            \, \right]
            - \mathbf{m}_1\mathbf{m}_2^{\trans}\\
            &= \frac{k_{1,2}}{\tilde{k}_{1}\tilde{k}_{2}} \left[
            \,
            \langle(\mathbf{x}-\mathbf{x}_{1})(\mathbf{x}-\mathbf{x}_{2})^{\trans}\rangle_{\Omega_{1,2}} 
            \,- \,
            \mathbf{m}_1\mathbf{m}_2^{\trans}
            \, \right]~,
            \end{split}
            \label{eq:mean_shift_sample_means_shift_cross_covariance}
            \end{equation}            
            where from the fourth to the fifth line we used the fact that the $\{\mathbf{x}_{i}\}_i$ are identically distributed in the first sum, while in the second sum the $\mathbf{x}_i$'s are independent from the $\mathbf{x}_{j}$'s for all pairs of indices $(i,j)$. Thanks to the proportionality of the sample gradient estimator to the sample mean shift estimator, as in Eq.~(\ref{eq:grad_F_sample_grad_F}), we can also give an expression for the cross-covariance between sample gradient estimates at two different points $\mathbf{x}_1$ and $\mathbf{x}_2$:
            \begin{equation}
            \mathrm{\mathbf{cov}}[\hat{\mathbf{g}}_1,\hat{\mathbf{g}}_2] 
            =
            \langle \hat{\mathbf{g}}_1 \hat{\mathbf{g}}_2^{\trans}\rangle
            - \langle \hat{\mathbf{g}}_1\rangle \langle\hat{\mathbf{g}}_2^{\trans}\rangle
            =
            \left( \frac{d+2}{r_{k_1}^2} \right) \left( \frac{d+2}{r_{k_2}^2} \right) \mathrm{\mathbf{cov}}[\hat{\mathbf{m}}_1,\hat{\mathbf{m}}_2]\, ,
            \label{eq:mean_shift_sample_grad_F_cross_covariance}
            \end{equation}            
            from which Eq.~(\ref{eq:deltaF_C_sample_grad_F_cross_covariance}) is derived.
            Thus, the cross-covariance between estimates at two different points $\mathbf{x}_1$ and $\mathbf{x}_2$ depends on the mean value of the matrix $(\mathbf{x}-\mathbf{x}_{1})(\mathbf{x}-\mathbf{x}_{2})^{\trans}$ over the region $\Omega_{1,2}$ in which the neighbourhoods of the two points overlap.

    \begin{figure}[!ht]
    \centering
        \includegraphics[width=\textwidth]{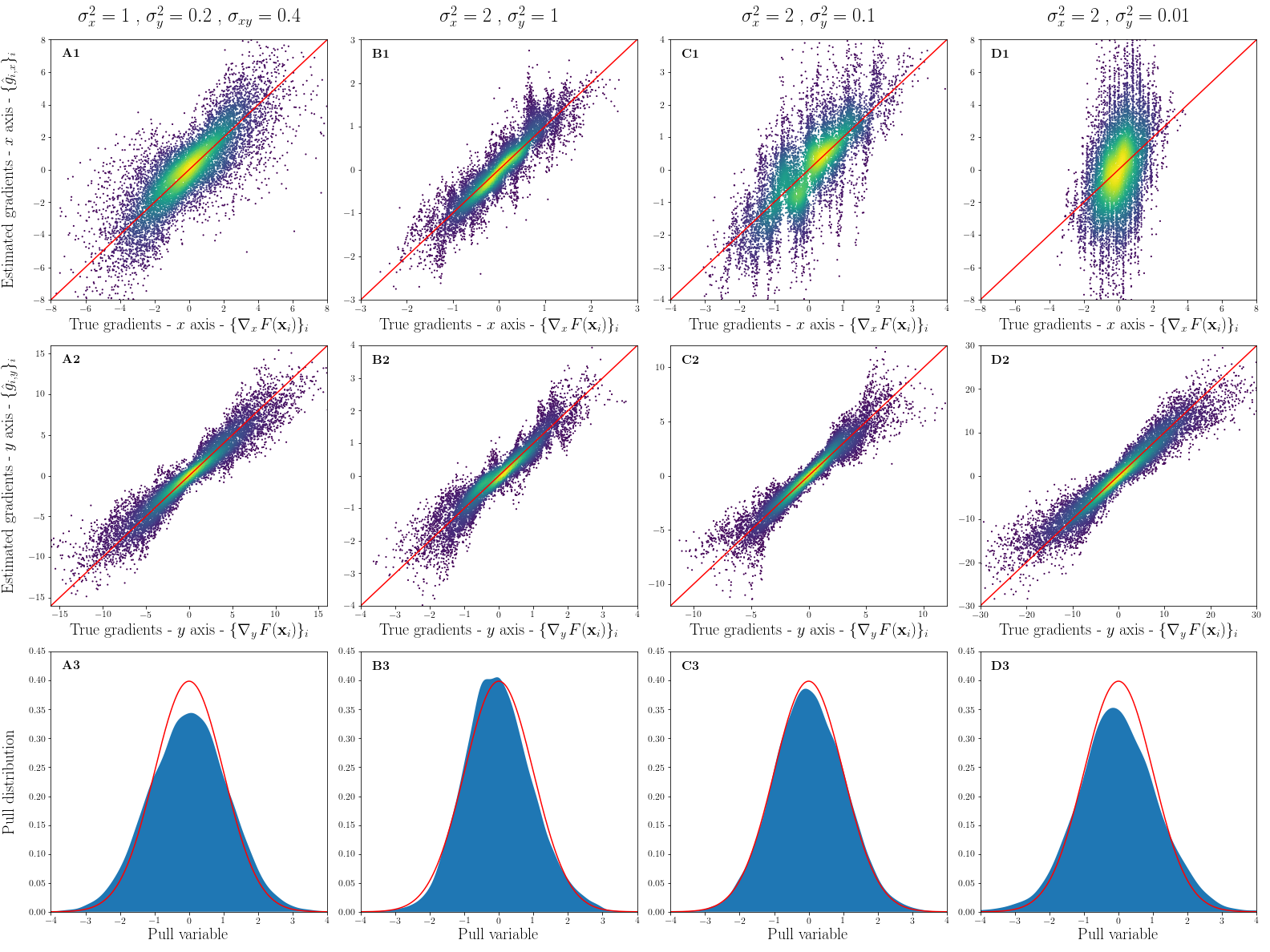}
    \caption[NLD gradient components estimator performance tested on various bivariate Gaussian datasets.]{
    \label{fig:mean_shift_grads_performance_gaussians_12panels}
    \textbf{NLD gradient components estimator performance tested on various bivariate Gaussian datasets}. 
    All four datasets considered, one for each column, have a bivariate normal PDF centred at the origin of the Cartesian plane (see Sec.~\ref{sssec:test_sys_2d_Gauss} of the SM) sampled $10,000$ times. The entries of each dataset's covariance matrix are indicated in the column header.
    \textbf{Top row}: correlation plots of estimated $x$ gradient components against true values. In red the line $\hat{g}_{i,x} = g_{i,x}$.
    \textbf{Middle row}: correlation plots of estimated $y$ gradient components against true values. In red the line $\hat{g}_{i,y} = g_{i,y}$.
    \textbf{Bottom row}: distribution of the pull of individual gradient components. In red the standard normal distribution $\mathcal{N}(0,1)$.
    }
    \end{figure}

    \subsection{Performance assessment for the $\hat{\mathbf{g}}$ estimator}
    \label{ssec:grad_F_grad_F_performance}
    
    In order to test the performance of the $\hat{\mathbf{g}}$ we look at the correlation plots of estimated vs. ground truth gradients and at the distribution of the pull, as explained in Sec.~\ref{sec:numerical}.
    Since we need the GT gradients, which we do not have for the realistic datasets in Sec.~\ref{ssec:test_sys_realistic} of the SM, we can only use synthetic datasets for this assessment. Indeed, we use the bivariate Gaussians in Sec.~\ref{sssec:test_sys_2d_Gauss}, the multimodal bivariate potential on a glassy background (\ref{sssec:test_sys_2d_MB_glassy}), the $6$-dimensional potential in Sec.~\ref{sssec:test_sys_6d} and the $9$-dimensional from Sec.~\ref{sssec:test_sys_CLN025} of the SM.

    \begin{figure}[!b]
    \centering
        \includegraphics[width=\textwidth]{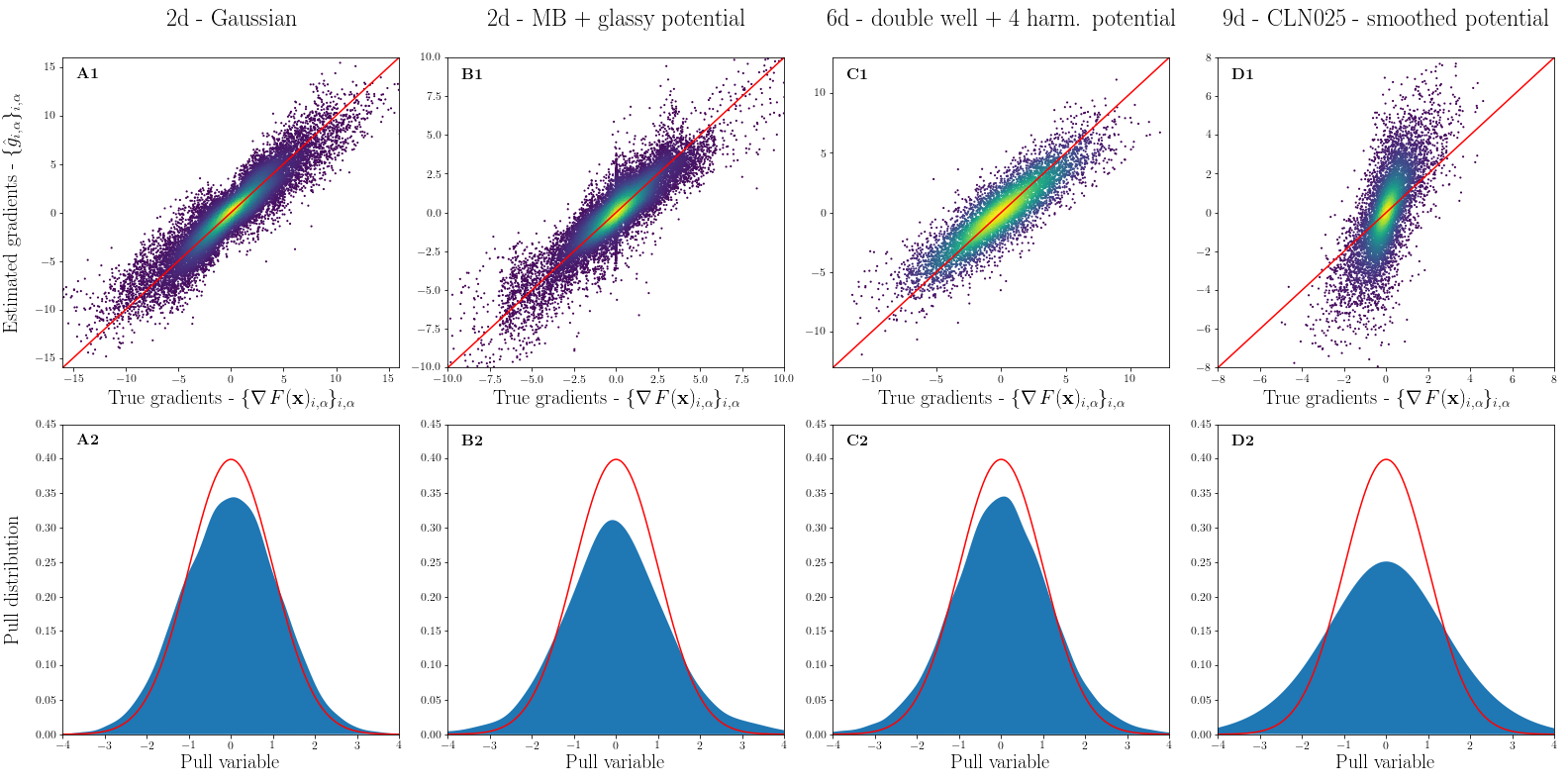}
    \caption[NLD gradient estimator performance tested on various datasets]{
    \label{fig:mean_shift_grads_performance_8panels}
    \textbf{NLD gradient estimator performance tested on various datasets}. 
    The four datasets, one for each column, are indicated in the column header; they are all described in Sec.~\ref{append:test_sys} of the SM; their dimensionality goes from $2$ to $9$. For all of them, the analytic expression of the NLD gradient is known. In the fourth and last column, the nine-dimensional case, $80,000$ sample points are considered; for all other datasets the sample size is $10,000$.
    In the first column, the dataset is the same considered in the first column of Fig.~\ref{fig:mean_shift_grads_performance_gaussians_12panels}.
    \textbf{Top row}: correlation plots of all estimated gradient components against true values. In red the line $\hat{g}_{i,\alpha} = g_{i,\alpha}$.
    \textbf{Bottom row}: distribution of the pull of gradient components. In red the standard normal distribution $\mathcal{N}(0,1)$.
    }
    \end{figure}
    
    \paragraph{Fig.~\ref{fig:mean_shift_grads_performance_gaussians_12panels}} illustrates the performance of the gradient estimator on four two-dimensional Gaussian probability distributions, with variances and covariances defined in the titles. The corresponding NLDs are represented in Fig.~\ref{fig:test_sys_2d_Gaussians}.
    In the top two rows we can see the correlation plots of the two estimated gradient components along the $x$ and $y$ axes against the true values. Looking at the parameters defining the distributions, in the column headers, we see that only the Gaussian in the first column has a non-diagonal covariance matrix. In the remaining columns the width of the Gaussian is kept fixed along the $x$ direction, while it is reduced more and more going from left to right. Along the $y$ axis we see that all estimates correlate well with the true values. Along the $x$ axis, instead, estimates are noisier and noisier going from left to right, namely towards smaller variance along the $y$ axis. 
    In panel B1 the structures we see are due to the finite statistics of the gradient estimates, which emphasises sample fluctuations (these fluctuations are present also in panel B2 but with a smaller amplitude, since $\sigma^2_y$ is smaller than $\sigma^2_x$. These become more and more evident in panels C1 and D1. 
    Indeed, the gradient estimated via the sample mean shift (\ref{eq:grad_F_sample_grad_F}) is good at capturing the gradient direction, but, in these datasets, the gradient is mostly oriented along the $y$ direction, so the relative error on the transverse direction is larger. Another way to understand this effect is that we are considering circular regions $\{\Omega_i\}_i$ in anisotropic landscapes; in these conditions the approximation leading to the mean shift equivalence in equation 
    (\ref{eq:mean_shift_mean_shift_calculation}) is partly violated and higher order corrections play a role, with a higher visible impact on the direction where the free energy varies more slowly. As for the Gaussian in the first column, since is orientation tilted w.r.t. and not aligned with any axis, the noise is present but is less structured in the correlation plot A1 with respect to the other examples. Of course, to obtain more sensible results one could standardise the data before applying the nonparametric gradient estimator, i.e. rescale the data coordinates dividing them by the sample standard deviation along the two directions; however, the point of considering such datasets is to test the estimator in extreme conditions. From this analysis, we see that the method is robust and does not lead to nonsense results even when stress-tested.
    
    As for the pull distribution for the gradient components, in the bottom row, it is in good agreement with the standard normal distribution for all the datasets. This is a sign that our gradient estimates are unbiased and that we correctly estimate their variance. The reason why, in terms of quality of the pull, the dataset in the first column appears to underperform columns $2$ and $3$ is that the ``aspect ratio'' of the first Gaussian is somewhere in between the ones of the third and fourth, as visible in row A of Fig.~\ref{fig:test_sys_2d_Gaussians} in Sec.~\ref{sssec:test_sys_2d_Gauss} of the SM.

    \paragraph{Fig.~\ref{fig:mean_shift_grads_performance_8panels}} shows the performance of the gradient estimator on four different model free energy landscapes (see Sec.~\ref{append:test_sys} of the SM) in terms of the correlation plot of estimated and true gradient components and the distribution of the pull of gradient components. In the correlation plots (top row), differently from Fig.~(\ref{fig:mean_shift_grads_performance_gaussians_12panels}), all gradient components, from $1$ to $D$, are plotted together. We can see that gradient estimates correlate quite well with the true analytical values. Only in the $9$-dimensional case, in panel D1, there is a visible bias: it can happen in fact that the gradient modulus is overestimated for some points, which results in a correlation plot slightly tilted w.r.t. to the identity line. Taking a closer look, it can be seen that this happens for points with few neighbours (see \cite{carli2022Tesi}). Indeed, the gradient of points with smaller neighbourhoods is affected by a large variance. The quality of the pull distributions in the second row testify that even in high dimensionality our error estimates are quite good. The reason why on the $2$-dimensional potential in panel B2 the gradient estimator performs worse than in the $6$-dimensional case, in panel C2, is because the former is designed to put a strain on estimators, being rugged and spiky, so that the selected neighbourhood size $k_i$ is for many points quite small.

\newpage
    
\section{Negative log-density difference estimators between neighbouring points: $\hat{\delta F}$}
\label{appendix:deltaFs}

    \subsection{Derivation of the expression for $\hat{\delta F}$}
    \label{ssec:deltaFs_derivation}
        
        In the spirit of the Taylor expansion, one could be tented to approximate the NLD difference between point $i$ and point $j$ at linear order and express it as the contraction between the estimated gradient at point $i$, $\mathbf{g}_i$, and their vector difference $\mathbf{r}_{ij} := \mathbf{x}_j - \mathbf{x}_i$:
        \begin{equation}
        \delta F_{ij}^{\bm{i}} := \nabla^{\rm{T}}_{\mathbf{x}}F(\mathbf{x}_i) (\mathbf{x}_j-\mathbf{x}_i)
                        = \mathbf{g}_i \cdot \mathbf{r}_{ij} \; ,
        \label{eq:beyond_pak_deltaFij_directional}
        \end{equation}
        where the upper bold index indicates the index of the gradient used. The estimator version of (\ref{eq:beyond_pak_deltaFij_directional}) is
        :
        \begin{equation}
            \hat{\delta F}_{ij}^{\bm{i}} := \hat{\mathbf{g}}_i \cdot \mathbf{r}_{ij}\ .
        \label{eq:deltaFs_deltaFij_directional_estim}
        \end{equation}
        \noindent
        However, the gradients in the two points $\mathbf{g}_i$ and $\mathbf{g}_j$ can be different, so in principle $\delta F_{ij}^{\bm{i}} \neq - \delta F_{ji}^{\bm{j}}$.

        The right quantity to contract with $\mathbf{r}_{ij}$ in order to obtain exactly $\delta F_{ij}$ would be the average NLD gradient along the connecting segment of $\mathbf{x}_i$ and $\mathbf{x}_j$.
        In fact, let us define a parametrisation $\bm{\lambda}: [0,1] \rightarrow \mathbb{R}^D$ of such segment, such that $\bm{\lambda}(t) \, = \, \mathbf{x}_i \, + \, t \, \mathbf{r}_{ij}$ and that the length of the segment is $\int_0^1  \|\bm{\lambda}'(t)\| \, \mathrm{d} t  \, = \, \|\mathbf{r}_{ij}\| = r_{ij}$
        . Then,
        \begin{equation}
            \delta F_{ij} 
            \, = \, 
            F(\mathbf{x}_j) - F (\mathbf{x}_i) 
            \, = \, 
            \int_{\mathbf{x}_i}^{\bm{x}_j}
            \partial_{\bm{\lambda}} F(\bm{\lambda}) \,
            \mathrm{d} \bm{\lambda}   
            \, = \, 
            \int_0^1 
            \nabla^{\rm{T}}_{\mathbf{x}} F(\bm{\lambda}(t))
            \, \cdot \,
            \bm{\lambda}'(t) \; \mathrm{d} t
            \, = \,
            \langle
            \,
            \nabla^{\rm{T}}_{\mathbf{x}} F
            \,
            \rangle_{\bm{\lambda}}
            \cdot \mathbf{r}_{ij}\ .
        \label{eq:deltaFs_deltaFij_average_gradient_lambda}
        \end{equation}
        The average NLD gradient along the segment connecting $\mathbf{x}_i$ and $\mathbf{x}_j$ is well approximated, until third order terms in the Taylor expansion of the NLD become relevant, by the semisum of the gradients in the two neighbouring points.
        In fact, if we write down the Taylor expansions of $F$ around $\mathbf{x}_i$ and $\mathbf{x}_j$ and remember that $\mathbf{r}_{ji} = - \mathbf{r}_{ij}$
        \begin{align*}
            F(\mathbf{x}_j) &= F(\mathbf{x}_i)
            + \nabla^{\rm{T}}_{\mathbf{x}} F(\mathbf{x}_i) \cdot \mathbf{r}_{ij}
            + \mathbf{r}_{ij}^{\rm{T}} \cdot \nabla^2_{\mathbf{x}} F(\mathbf{x}_i) \cdot \mathbf{r}_{ij}
            + \mathcal{O} (\nabla^3_{\mathbf{x}} F(\mathbf{x}_i) \cdot \mathbf{r}_{ij}^3)
            \\
            F(\mathbf{x}_i) &= F(\mathbf{x}_j)
            - \nabla^{\rm{T}}_{\mathbf{x}} F(\mathbf{x}_j) \cdot \mathbf{r}_{ij}
            + \mathbf{r}_{ij}^{\rm{T}} \cdot \nabla^2_{\mathbf{x}} F(\mathbf{x}_j) \cdot \mathbf{r}_{ij}
            - \mathcal{O} (\nabla^3_{\mathbf{x}} F(\mathbf{x}_j) \cdot \mathbf{r}_{ij}^3)
        \end{align*}
        and subtract them -- inserting the notations $\mathbf{g}_i = \nabla_{\mathbf{x}}F(\mathbf{x}_i)$ and $\mathbf{H}_i = \nabla^2_{\mathbf{x}}\,F(\mathbf{x}_i) $ for gradient and Hessian of the NLD respectively -- we obtain
        \begin{equation*}
            2 \,  \left(\,F(\mathbf{x}_j) - F(\mathbf{x}_i)\, \right) \,=\, 
            + \left( 
            \mathbf{g}_i + \mathbf{g}_j
            \right) \cdot \mathbf{r}_{ij}
            + \mathbf{r}_{ij}^{\rm{T}} \cdot \left(
            \mathbf{H}_i - \mathbf{H}_j
            \right) \cdot \mathbf{r}_{ij}
            + \mathcal{O} \left(\,
                \left(
                    \nabla^3_{\mathbf{x}} F(\mathbf{x}_i) + \nabla^3_{\mathbf{x}} F(\mathbf{x}_j) 
                \right) \cdot \mathbf{r}_{ij}^3
            \, \right) \; .
        \end{equation*}
        Since we are interested in neighbouring points and we are assuming, according to Sec.~\ref{sssec:BMTI_deltaF_manifold+adaptive}, this means that the NLD is approximately constant, we can expect the term $\mathbf{r}_{ij}^{\rm{T}} \cdot \left(
            \mathbf{H}_i - \mathbf{H}_j
            \right) \cdot \mathbf{r}_{ij}$ to be of order $\mathcal{O} \left(  \nabla^3_{\mathbf{x}} F \cdot \mathbf{r}_{ij}^3 \right)$, so that
        \begin{equation}
            \delta F_{ij} \,=\, 
            \frac{  \mathbf{g}_i + \mathbf{g}_j }{2} \cdot \mathbf{r}_{ij}
            \,+\, \mathcal{O} \left(  \nabla^3_{\mathbf{x}} F \cdot \mathbf{r}_{ij}^3 \right) \; .
        \end{equation}
        Comparing this to Eq.~(\ref{eq:deltaFs_deltaFij_average_gradient_lambda}), we reckon that
        \begin{equation}
            \langle
            \,
            \nabla^{\rm{T}}_{\mathbf{x}} F
            \,
            \rangle_{\bm{\lambda}} \,=\, 
            \frac{  \mathbf{g}_i + \mathbf{g}_j }{2} 
            \,+\, \mathcal{O} \left(  \nabla^3_{\mathbf{x}} F \cdot \mathbf{r}_{ij}^2 \right)
            \approx
            \frac{  \mathbf{g}_i + \mathbf{g}_j }{2} 
            \; ,
        \end{equation}
        from which we can define the estimator for the NLD difference $\delta F_{ij}$ can as in Eq.~(\ref{eq:deltaFij_estim}), namely
        \begin{equation*}
            \hat{\delta F}_{ij} :=
            \frac{\hat{\mathbf{g}}_i + \hat{\mathbf{g}}_j}{2} \cdot \mathbf{r}_{ij}\ .
        \end{equation*}
    
    \subsection{Covariance structure of the $\hat{\delta F}$ estimators}
    \label{ssec:deltaFs_errors}
    As evident from Eq.~(\ref{eq:deltaFij_C_ijlm_def}), of which Eq.~(\ref{eq:deltaFij_variance}) is a special case, estimating the covariance matrix $\mathbf{C}$ of the $\hat{\delta F}$'s, entering the BMTI log-likelihood in Eq.~(\ref{eq:BMTI_loglikelihood_full}) depends on being able to capture the gradient covariance structure.
    Let us inspect the covariance structure of the $\hat{\delta F}$ estimators starting from the simplest case and going to the most generic one.

        \subsubsection{The variance of the $\hat{\delta F}$'s}
        \label{sssec:deltaFs_errors_variance}
        By recalling Eq.s (\ref{eq:deltaFij_variance}) and (\ref{eq:deltaFij_err_deltaF_ij_with_p}) we can rewrite here, for the reader's convenience
        \begin{equation}
        \begin{split}
            C_{ij,ij}
            =
            \textrm{var}[\hat{\delta F}_{ij}]
            =
            \varepsilon^2_{ij}
            &=
            \mathbf{r}_{ij}^{\rm{T}} \,
            \mathbf{var}\left[\frac{\hat{\mathbf{g}}_i+ 
            \hat{\mathbf{g}}_j}{2} \right]\,
            \mathbf{r}_{ij} 
            \\
            &=
            \frac{1}{4}\,
            \mathbf{r}_{ij}^{\rm{T}}
            \left(
            \textbf{var}[\hat{\mathbf{g}}_i]
            +
            \textbf{var}[\hat{\mathbf{g}}_j]
            +
            2 \, \textbf{cov}[\hat{\mathbf{g}}_i,\hat{\mathbf{g}}_j]
            \right)
            \mathbf{r}_{ij}
            \\
            &=
            \frac{1}{4}
            \left(
            \textrm{var}[\hat{\delta F}^{\bm{i}}_{ij}]
            +
            \textrm{var}[\hat{\delta F}^{\bm{j}}_{ij}]
            +
            2 \, \textrm{cov}[\hat{\delta F}^{\bm{i}}_{ij},\hat{\delta F}^{\bm{j}}_{ij}]
            \right)
            \\
            &=
            \frac{1}{4}
            \left(
            \textrm{var}[\hat{\delta F}^{\bm{i}}_{ij}]
            +
            \textrm{var}[\hat{\delta F}^{\bm{j}}_{ij}]
            +
            2 \, p^{\bm{ij}} \, \textrm{var}[\hat{\delta F}^{\bm{i}}_{ij}] \, \textrm{var}[\hat{\delta F}^{\bm{i}}_{ij}]
            \right)\\
            &=
            \frac{1}{4}
            \left(
            {\varepsilon_{ij}^{\bm{i}}}^2
            +
            {\varepsilon_{ij}^{\bm{j}}}^2 
            + 2 \, p^{\bm{ij}} \, \varepsilon_{ij}^{\bm{i}} \,\varepsilon_{ij}^{\bm{j}}
            \right)
            \; ,
        \end{split}
        \label{eq:deltaFs_variance}
        \end{equation}
        where we see appearing the directionsl $\delta F$'s estimators defined in Eq.~(\ref{eq:deltaFs_deltaFij_directional_estim}), their standard deviations $\hat{\varepsilon}^{\bm{i}}_{ij} :=\sqrt{ \mathbf{r}_{ij}^{\rm{T}} \cdot \hat{\mathbf{var}}[\hat{\mathbf{g}}_i] \cdot \mathbf{r}_{ij}^{\trans} }$ and the Pearson correlation coefficients $p^{\bm{ij}}$ between the estimators $\hat{\delta F}_{ij}^{\bm{i}}$ and $\hat{\delta F}_{ij}^{\bm{j}}$ seen as random variables, as defined in Eq.~(\ref{eq:deltaFij_pearson_def}).
        By separating the modulus and sign contribution in the $p$'s, such that $\chi^{\bm{ij}}:=|p^{\bm{ij}}|$ and $s^{\bm{ij}} := \mathrm{sgn}(p^{\bm{ij}})$, we can expand Eq.~(\ref{eq:deltaFij_pearson_def}) as
        \begin{equation}
            p^{\bm{ij}} \; 
            := \;  
            \frac{
            \mathbf{r}_{ij}^{\rm{T}} \cdot \mathbf{cov}[\hat{\mathbf{g}}_i,\hat{\mathbf{g}}_j] \cdot \mathbf{r}_{ij}
            }{
            \varepsilon_{ij}^{\bm{i}} \,\varepsilon_{ij}^{\bm{j}}
            } 
            \; = \;
            \frac{
            \mathrm{cov}[\hat{\delta F}_{ij}^{\bm{i}},\hat{\delta F}_{ij}^{\bm{j}}]
            }{
            \varepsilon_{ij}^{\bm{i}} \,\varepsilon_{ij}^{\bm{j}}
            } 
            \; = \;
            \mathrm{sgn}\left(\mathrm{cov}[\hat{\delta F}_{ij}^{\bm{i}},\hat{\delta F}_{ij}^{\bm{j}}] \right)
            \left| \frac{
            \mathrm{cov}[\hat{\delta F}_{ij}^{\bm{i}},\hat{\delta F}_{ij}^{\bm{j}}]
            }{
            \varepsilon_{ij}^{\bm{i}} \,\varepsilon_{ij}^{\bm{j}}
            } \right|
            \; =: \;
            s^{\bm{ij}}\, \chi^{\bm{ij}}
            \; .
        \label{eq:deltaFs_pearson_def_expanded}
        \end{equation}
        The Pearson correlation coefficient between two RVs takes values between $-1$ and $1$.
        It is $0$ when the two RVs are completely independent.
        Its modulus is $1$ when the RVs are perfectly linearly dependent, i.e. if they are identical up to a scalar and a constant offset.
        Therefore, $s^{\bm{ij}} \in \{-1,1\}$, while $\chi^{\bm{ij}} \in [0,1]$.
          
        \subsubsection{The cross-covariance among the $\hat{\delta F}$'s}
        \label{sssec:deltaFs_errors_cross-covariance}
        Let us elaborate the definition of a generic element of the $\delta F$'s covariance matrix in Eq.s (\ref{eq:deltaFij_estim}) and (\ref{eq:deltaFij_C_ijlm_def})
        \begin{equation}
            \begin{split}
                C_{ij,lm} :&= \textrm{cov}[\hat{\delta F_{ij}}\, , \,\hat{\delta F_{lm}}]\\
                    &=  \mathbf{r}^{\rm{T}}_{ij}
                        \cdot
                        \mathbf{cov}
                        \left[
                        \frac{\hat{\mathbf{g}}_i + \hat{\mathbf{g}}_j}{2},
                        \frac{\hat{\mathbf{g}}_l + \hat{\mathbf{g}}_m}{2}
                        \right]
                        \cdot
                        \mathbf{r}_{lm}\\
                    &=  \frac{1}{4} \,
                        \mathbf{r}^{\rm{T}}_{ij}
                        \cdot
                        \left(
                        \textbf{cov} [ \hat{\mathbf{g}}_i , \hat{\mathbf{g}}_l]
                        + \textbf{cov} [ \hat{\mathbf{g}}_i , \hat{\mathbf{g}}_m]
                        + \textbf{cov} [ \hat{\mathbf{g}}_j , \hat{\mathbf{g}}_l] 
                        + \textbf{cov} [ \hat{\mathbf{g}}_j , \hat{\mathbf{g}}_m]
                        \right)
                        \cdot
                        \mathbf{r}_{lm} \\
                    &=  \frac{1}{4} \,
                        \left(
                        \mathrm{cov}\left[\hat{\delta F}_{ij}^{\bm{i}},\hat{\delta F}_{lm}^{\bm{l}} \right]
                        +\mathrm{cov}\left[\hat{\delta F}_{ij}^{\bm{i}},\hat{\delta F}_{lm}^{\bm{m}} \right]
                        +\mathrm{cov}\left[\hat{\delta F}_{ij}^{\bm{j}},\hat{\delta F}_{lm}^{\bm{l}} \right]
                        +\mathrm{cov}\left[\hat{\delta F}_{ij}^{\bm{j}},\hat{\delta F}_{lm}^{\bm{m}} \right]
                        \right)
                    \; .
            \end{split}
            \label{eq:deltaFs_C_ijlm_manipulation}
        \end{equation}
        Analogously to what done in Eq.~(\ref{eq:deltaFs_pearson_def_expanded}), we define the Pearson correlation coefficients between terms of the kind appearing in covariances in the last line of Eq.~(\ref{eq:deltaFs_C_ijlm_manipulation})
        \begin{equation}
            p^{\bm{il}}_{i,lm}
            :=  
            \frac{
            \mathbf{r}_{ij}^{\rm{T}} \cdot \mathbf{cov}[\hat{\mathbf{g}}_i,\hat{\mathbf{g}}_l] \cdot \mathbf{r}_{lm}
            }{
            \varepsilon_{ij}^{\bm{i}} \,\varepsilon_{lm}^{\bm{l}}
            } 
            =
            \frac{
            \mathrm{cov}[\hat{\delta F}_{ij}^{\bm{i}},\hat{\delta F}_{lm}^{\bm{l}}]
            }{
            \varepsilon_{ij}^{\bm{i}} \,\varepsilon_{lm}^{\bm{l}}
            } 
            =
            \mathrm{sgn} \! \left(
                \mathrm{cov} \left[ \hat{\delta F}_{ij}^{\bm{i}},\hat{\delta F}_{lm}^{\bm{l}} \right]
            \right)
            \left|
                \frac{
                    \mathrm{cov} \left[ \hat{\delta F}_{ij}^{\bm{i}},\hat{\delta F}_{lm}^{\bm{l}} \right]
                }{
                    \varepsilon_{ij}^{\bm{i}} \,\varepsilon_{lm}^{\bm{l}}
                }
            \right|
            =:
            s^{\bm{il}}_{ij,lm}\, \chi^{\bm{il}}_{ij,lm}
            \; ,
        \label{eq:deltaFs_pearson_ijlm}
        \end{equation}
        of which Eq.~(\ref{eq:deltaFs_pearson_def_expanded}) is a specific case upon identification of $p^{\bm{il}} \equiv p^{\bm{ij}}_{ij,ij}$, $s^{\bm{il}} \equiv s^{\bm{ij}}_{ij,ij}$ and $\chi^{\bm{ij}} \equiv \chi^{\bm{ij}}_{ij,ij}$.
        Also here is $\chi^{\bm{ij}}_{ij,ij} \in [0,1]$.
        We can rewrite Eq.~(\ref{eq:deltaFs_C_ijlm_manipulation}) using Eq.~(\ref{eq:deltaFs_pearson_ijlm}) as
        \begin{equation}
            C_{ij,lm} =
                    \frac{1}{4}
                    \,(\,
                    p^{\bm{il}}_{ij,lm}
                    \,\varepsilon_{ij}^{\bm{i}} \,\varepsilon_{lm}^{\bm{l}}
                    +
                    p^{\bm{im}}_{ij,lm}
                    \,\varepsilon_{ij}^{\bm{i}} \,\varepsilon_{lm}^{\bm{m}}\\
                    +
                    p^{\bm{jl}}_{ij,lm}
                    \,\varepsilon_{ij}^{\bm{j}} \,\varepsilon_{lm}^{\bm{l}}
                    +
                    p^{\bm{jm}}_{ij,lm}
                    \,\varepsilon_{ij}^{\bm{j}} \,\varepsilon_{lm}^{\bm{m}}
                    \,)
                    \; .
            \label{eq:detaFs_C_ijlm_with_pearsons}
        \end{equation}
         Note that the expression for $C_{ij,lm}$ in  Eq.~(\ref{eq:detaFs_C_ijlm_with_pearsons}) is symmetric upon exchange of the $\delta F$s, i.e. $(i,j) \leftrightarrow (l,m)$ and antisymmetric upon exchange of the first-and-second or third-and-fourth indices, i.e. $i \leftrightarrow j$ and $l \leftrightarrow m$.
        
        \subsubsection{Estimating the Pearson correlation coefficients between the $\hat{\delta F}$'s}
        \label{sssec:deltaFs_errors_pearson}
        By looking at Eq.~(\ref{eq:deltaFij_C_ijlm_def}), Eq.~(\ref{eq:deltaFs_pearson_ijlm}) and related expressions, the typical quantity that we want to estimate is a contraction of the cross-covariance between two gradient estimates with some vector differences:
        \begin{equation}
            \mathbf{r}_{ij}^{\rm{T}} \cdot \mathbf{cov}[\hat{\mathbf{g}}_i,\hat{\mathbf{g}}_l] \cdot \mathbf{r}_{lm}
            \; =\; 
            \mathrm{cov}\left[\hat{\delta F}_{ij}^{\bm{i}},\hat{\delta F}_{lm}^{\bm{l}} \right]
            \; =\; 
            p^{\bm{il}}_{ij,lm}
            \,\varepsilon_{ij}^{\bm{i}} \,\varepsilon_{lm}^{\bm{l}}
            \; =\; 
            s^{\bm{il}}_{ij,lm}\, \chi^{\bm{il}}_{ij,lm}
            \,\varepsilon_{ij}^{\bm{i}} \,\varepsilon_{lm}^{\bm{l}}
            ~ .
        \label{eq:deltaFs_most_generic_C_term}
        \end{equation}
        The standard deviations of the directional $\delta F$ can be estimated by using the sample gradient autocovariance estimator in Eq.~(\ref{eq:grad_F_grad_F_sample_autocovariance})
        \begin{equation}
            \hat{\varepsilon}^{\bm{i}}_{ij} :=\sqrt{ \mathbf{r}_{ij}^{\rm{T}} \cdot \hat{\mathbf{var}}[\hat{\mathbf{g}}_i] \cdot \mathbf{r}_{ij}^{\trans} }
            \, \approx\,
            \sqrt{ \mathbf{r}_{ij}^{\rm{T}} \cdot \mathbf{var}[\hat{\mathbf{g}}_i] \cdot \mathbf{r}_{ij}^{\trans} }
            =
            \mathrm{var} \left[\hat{\delta F}_{ij}^{\bm{i}} \right]
            \; .
        \label{eq:deltaFs_directional_deltaF_stdev_estim}
        \end{equation}
        Instead, estimating the Pearson correlation coefficient $p$ is less straightforward.
        In Eq.~(\ref{eq:deltaFs_most_generic_C_term}) we decompose it into the contributions of its sign $s$ and its modulus $\chi$, which can be estimated separately.
            \paragraph{Modulus: the Jaccard index}
            Since the two gradient estimators $\hat{\mathbf{g}}_i$ and $\hat{\mathbf{g}}_l$ are the only RVs on the left-most expression in Eq.~(\ref{eq:deltaFs_most_generic_C_term}), the modulus of this such expression is maximum when $i=l$ and $0$ when the two gradients estimators are uncorrelated.
            In particular, since for all points $i$ the estimator $\hat{\mathbf{g}}_i$ is a sum of $k_i$ i.i.d. RVs depending on the positions of the $k_i$ points contained in the neighbourhood $\Omega_i$, (\ref{eq:deltaFs_most_generic_C_term}) is proportional to the number of points that $\Omega_i$ and $\Omega_l$ have in common, namely $k_{i,l}$.
            The only element in the right-most expression of Eq.~(\ref{eq:deltaFs_most_generic_C_term}) which can incorporate this discrete behaviour is $\chi^{\bm{il}}_{ij,lm}$.
            Thus, an estimator of $\chi^{\bm{il}}_{ij,lm}$ will have $k_{i,l}$ at the numerator, but should be normalised in order to return $0$ when $\Omega_i \cap \Omega_l = \varnothing$, i.e. $k_{i,l}=0$, and $1$, when $i=l$.
            In order to correctly normalise $\chi^{\bm{il}}_{ij,lm}$ we should choose a quantity that goes to $k_{i,l}$ when $i = l$ and goes to a finite value (not $0$) when $k_{i,l}=0$.
            Therefore, a very natural and convenient choice is to make the estimator for $\chi^{\bm{il}}_{ij,lm}$ independent of the pedices $ij,lm$, so that it only depends on the neighbourhoods over which the two gradient estimators $\hat{\mathbf{g}}_i$ and $\hat{\mathbf{g}}_l$ (indicated in the apices) are defined.
            The form is that of the \textit{Jaccard index} \cite{jaccard1912distribution}:
            \begin{equation}
                \hat{\chi}^{\bm{il}}
                \; := \;
                \frac{k_{i,l}}{k_{i}+k_{l}-k_{i,l}}
                \; := \;
                \frac{|\Omega_i \cap \Omega_l|}{|\Omega_i \cup \Omega_l|}
                ~ .
            \label{eq:deltaFs_chi_estimator_Jaccard}
            \end{equation}
            The expression (\ref{eq:deltaFs_chi_estimator_Jaccard}) can also be interpreted in a continuum of points as the ratio between the volume of the intersection of the hyperspherical neighbourhoods of the two points $i$ and $l$: $V_{\Omega_i \cap \Omega_j}/V_{\Omega_i \cup \Omega_j}$.
            Since $k_i$ is proportional to the $d$-volume $V_i = \omega_d r^d_i$ of $\Omega_i$ via the relation $k_i = V_i \; \hat{\rho}_i^{\hat{k}\mathrm{NN}}$, the connection between the discrete and the continuum is straightforward and the picture is coherent.
            Other possible choices for $\hat{\chi}^{\bm{ij}}$ which have the desired properties are discussed in Ref. \cite{carli2022Tesi}.
            \paragraph{Sign: one-shot estimator}
            The sign of $p$, both in Eq.~(\ref{eq:deltaFs_pearson_def_expanded}) and (\ref{eq:deltaFs_pearson_ijlm}), corresponds to the sign of the covariance at the numerator of the definition, since the denominator is made of quantities which are positive by construction.
            Taking Eq.~(\ref{eq:deltaFs_pearson_ijlm}), if we wanted to estimate the covariance $\mathrm{cov}[\hat{\delta F}_{ij}^{\bm{i}},\hat{\delta F}_{lm}^{\bm{l}} ] $ we would need a statistic on the product of the directional $\hat{\delta F}$'s.
            Instead, despite every $\hat{\delta F}_{ij}^{\bm{i}}$ being a statistic itself, we only estimate one realisation for this RV given a dataset.
            We choose to estimate the sign $s^{\bm{il}}_{ij,lm}$ as the one-shot quantity rather than a statistic, by simply evaluating the sign of the product of the two directional $\hat{\delta F}$ RVs estimated from the dataset
            \begin{equation}
                \hat{s}^{\bm{il}}_{ij,lm}
                \; := \;
                \mathrm{sgn}\left(\hat{\delta F}_{ij}^{\bm{i}} \; \hat{\delta F}_{lm}^{\bm{l}} \right)
                \; \approx \;
                \mathrm{sgn}\left(\mathrm{cov}[\hat{\delta F}_{ij}^{\bm{i}},\hat{\delta F}_{lm}^{\bm{l}}] \right)
                \; = \;
                s^{\bm{il}}_{ij,lm}
                ~ .
            \label{eq:deltaFs_sign_estimator}
            \end{equation}

        \subsubsection{Putting all together: an estimator for the $\hat{\delta F}$'s covariance matrix}
        \label{sssec:deltaFs_errors_cross-covariance-estimator}
        Let us consider a generic term of the the $\hat{\delta F}$'s cross-covariance $\mathbf{C}$, i.e. Eq.~(\ref{eq:deltaFij_C_ijlm_def}) in the main text and (\ref{eq:deltaFs_C_ijlm_manipulation}) in the SM.
        Using Eq.s (\ref{eq:deltaFs_sign_estimator}) and (\ref{eq:deltaFs_chi_estimator_Jaccard}) to estimate the sign and modulus of the Pearson correlation coefficient in Eq.~(\ref{eq:deltaFs_pearson_ijlm}) we obtain exactly the estimator in Eq.~(\ref{eq:deltaFij_pil_ijlm_est}), that we report here for convenience
        \begin{equation*}
            \hat{p}^{\bm{il}}_{ij,lm} 
            \,=\, 
            \mathrm{sgn}\left(\hat{\delta F}_{ij}^{\bm{i}} \; \hat{\delta F}_{lm}^{\bm{l}} \right)
            \,
            \hat{\chi}^{\bm{ij}}
            \, ,
        \end{equation*}
        which in the case $(ij)=(lm)$ simplifies to Eq.~(\ref{eq:deltaFij_pearson_est_def}).
        By using Eq.s (\ref{eq:deltaFs_directional_deltaF_stdev_estim}) and (\ref{eq:deltaFij_pil_ijlm_est}) we now have all the elements to estimate $C_{ij,lm}$ and recover Eq.~(\ref{eq:detaFij_C_ijlm_estim}), which we transcribe here
        \begin{equation*}
             \hat{C}_{ij,lm} =
                    \frac{1}{4}
                    \,(\,
                    \hat{p}^{\bm{il}}_{ij,lm}
                    \,\hat{\varepsilon}_{ij}^{\bm{i}} \,\hat{\varepsilon}_{lm}^{\bm{l}}
                    +
                    \hat{p}^{\bm{im}}_{ij,lm}
                    \,\hat{\varepsilon}_{ij}^{\bm{i}} \,\hat{\varepsilon}_{lm}^{\bm{m}}
                    +
                    \hat{p}^{\bm{jl}}_{ij,lm}
                    \,\hat{\varepsilon}_{ij}^{\bm{j}} \,\hat{\varepsilon}_{lm}^{\bm{l}}
                    +
                    \hat{p}^{\bm{jm}}_{ij,lm}
                    \,\hat{\varepsilon}_{ij}^{\bm{j}} \,\hat{\varepsilon}_{lm}^{\bm{m}}
                    \,)
                    \; .
        \end{equation*}        
        This is the $\mathbf{C}$ estimator we chose and implemented in all the results discussed in the present work.
        Its accuracy, at least for the diagonal terms $C_{ij,ij}={\varepsilon_{ij}}^2$, is commented in Sec.~\ref{ssec:numerical_performance} and Sec.~\ref{ssec:deltaFs_performance} of the SM by looking at the pull distributions in Fig.~\ref{fig:deltaF_perform}.

    \subsection{Comment on the performance of the $\hat{\delta F}$ estimator}
    \label{ssec:deltaFs_performance}
    In order to test our estimator $\hat{\delta F}_{ij}$ we resort to correlation plots of $\hat{\delta F}_{ij}$ against $\delta F_{ij}$ and distributions of the pull variables.
    The tests are shown in Fig.~\ref{fig:deltaF_perform}, in the main text.
    All correlation plots and pull distributions are in excellent agreement with the predictions for unbiased estimators. 
    Comparing the $\hat{g}$ and the $\hat{\delta F}$ performances for the bivariate Gaussians, in Fig.s \ref{fig:mean_shift_grads_performance_gaussians_12panels}, \ref{fig:mean_shift_grads_performance_8panels} to Fig.~\ref{fig:deltaF_perform}, we see that the noise present in the gradient components estimates is strongly damped and we observe better overall pull distributions for the $\hat{\delta F}$'s than for the gradient components in Fig.~\ref{fig:mean_shift_grads_performance_8panels}.
    These tests demonstrate that the estimator $\hat{\delta F}_{ij}$ is more robust than the estimator $\hat{\mathbf{g}}_i$; we explain this fact by considering that by taking the semisum of two gradient estimates as in equation (\ref{eq:deltaFij_estim}), errors compensate at second order, bringing the leading-order corrections in the estimator $\hat{\delta F}_{ij}$ to third order.
    Fig.~\ref{fig:deltaF_perform} display an excellent overall performance in a wide range of datasets, embedding dimensionalities and IDs for both the estimators of the neighbours free energy difference $\hat{\delta F}_{ij}$ and of its error $\hat{\varepsilon}_{ij}$, which includes our empirical correction $\hat{p}_{ij}$ discussed in Sec.~\ref{sssec:deltaFs_errors_pearson} of the SM.
    These $\hat{\delta F}$ estimators and their error estimators guarantee that Eq.~(\ref{eq:BMTI_deltaF_normally_distributed}) is satisfied.

\newpage

\section{Solution of the BMTI likelihood}
\label{appendix:BMTI_support}
    
    \subsection{The estimation of error for BMTI}
    \label{ssec:BMTI_support_error}
    
    Given a log-likelihood like the one in Eq.~(\ref{eq:BMTI_loglikelihood_full}), the covariance matrix of the maximum-likelihood estimators that can be derived, once again, by taking the equal sign in the Cram{\'e}r–Rao Bound inequality:
    
    \begin{equation}
        \mathrm{\mathbf{cov}}[\hat{\mathbf{F}}]_{ij} \,:=\, \left\langle -\frac{\partial^2}{\partial F_i \partial F_j}\, \mathcal{L} ( \mathbf{F} \mid \hat{\bm{\delta F}}\,,\, \mathbf{C} ) \right\rangle^{-1}\ .
        \label{eq:beyond_pak_gCorr_CRB}
    \end{equation}
    
    The diagonal elements of this covariance matrix represent our uncertainty estimates on the MLEs $\{\hat{F}_i\}_i$.
    In particular, the inverse of $\rm{\mathbf{cov}}[\hat{\mathbf{F}}]$ corresponds exactly to the matrix $\mathbf{A}$ appearing in Sec.~\ref{ssec:BMTI_derivation}, therefore
    \begin{equation}
        \varepsilon_i^2 = \mathrm{var}[\hat{F}_i] = (\mathbf{A}^{-1})_{ii}.
        \label{eq:BMTI_support_BMTI_F_errors_inverse_A}
    \end{equation}
    and Eq.~(\ref{eq:BMTI_F_error}) is recovered, so estimating the error on the estimates $\hat{\mathbf{F}}$ amounts to inverting matrix $\mathbf{A}$. 

    \subsection{Approximate inversion of the covariance matrix $\mathbf{C}$}
    \label{ssec:BMTI_support_approx_inverse_C}
    As discussed in Sec.~\ref{ssec:BMTI_practical_sol}, we choose to approximate $\mathbf{C}^{-1}$ by a diagonal matrix.
    Ideally, we would like to pseudo-invert $\mathbf{C}$ fully and then only consider its diagonal $\tilde{D}_{ij} \; := \; C^{-1}_{ij,ij}$ or, even better, to use some numeric methods that, with a lower computational cost, are able to only compute the diagonal of $\mathbf{C}^{-1}$.
    In practice, we also use an approximate version $\mathbf{D}$ of matrix $\tilde{\mathbf{D}}$, with which the BMTI full likelihood in Eq.~(\ref{eq:BMTI_loglikelihood_full}) becomes the approximated version in Eq.~(\ref{eq:BMTI_loglikelihood_approx}).
    In the following we discuss two possible ways to specify $\mathbf{D}$, approximating $\tilde{\mathbf{D}}$ with increasing levels of crudity.
    Currently, part of our efforts are in finding ways to compute $\tilde{\mathbf{D}}$ or invert the matrix $\mathbf{C}$ more rigorously, in order to solve the unsatisfactory accuracy of the error estimates provided by the approximate BMTI likelihoods discussed in Sec.~\ref{sssec:BMTI_support_approx_inverse_C_least_square} and \ref{sssec:BMTI_support_approx_inverse_C_gCorr} of the SM.
    
        \subsubsection{Least-squares optimal diagonal inverse}
        \label{sssec:BMTI_support_approx_inverse_C_least_square}
        Since by construction 
        $\mathbf{C}$ is sparse and semipositive-definite, we expect its inverse to be more concentrated on the diagonal and depleted off-diagonal \cite{horn2012matrix}.
        A possible strategy is not to invert the matrix $\{C_{a,b}\}_{a,b}$ directly, but to find the diagonal matrix $\mathbf{D} = \mathrm{diag}(\textbf{d}^*)$ which best approximates the inverse of $\mathbf{C}$. For example:
        \begin{equation}
            \textbf{d}^*(\mathbf{C}) = \underset{\textbf{d}}{\operatorname{arg\,min}} ~\frac{1}{2} \| \mathbf{C} \cdot \mathrm{diag}(\textbf{d}) - \mathbb{1} \|_F^2 
            \label{eq:BMTI_support_optimal_diag_inv_C_def}
        \end{equation}
        optimises the diagonal of the approximate inverse by minimising the $L_2$ Frobenius norm of the difference between $\mathbf{C} \cdot \mathbf{D}$ and the identity. The solution is:
        \begin{equation}
            \textbf{d}^*_a = \frac {C_{aa} } { || \mathbf{C}_a ||^2} = \frac {C_{aa} } { \sum_b {C_{ab}}^2 }
            \label{eq:BMTI_support_optimal_diag_inv_C_sol}
        \end{equation}
        where $\mathbf{C}_a$ indicates the $a$-th row of matrix $\mathbf{C}$.
        
        This approximation provides accurate predictions for the BMTI estimates $\{ \hat{F}_i \}_i$ for all tested cases.
        As for the uncertainties, estimated using Eq.~(\ref{eq:BMTI_support_BMTI_F_errors_inverse_A}), they display mixed results: they are quite accurate in very sparse settings, but tend to be overestimated in low dimensional spaces, where the neighbourhood graph has typically many connections.

        \subsubsection{Diagonal of the inverse as inverse of the diagonal}
        \label{sssec:BMTI_support_approx_inverse_C_gCorr}

        One could take $\mathbf{D}$ as the inverse of a diagonal matrix having the same diagonal as $\mathbf{C}$. In mathematical notation, which makes this concept straightforward:
        \begin{equation}
            D_{ij} \; = \; (C_{ij,ij})^{-1} \; = \; \frac{1}{\varepsilon_{ij}^2}  \; .
        \end{equation}
        With this definition, the approximate log-likelihood in Eq.~(\ref{eq:BMTI_loglikelihood_approx}) reads like Eq.~(\ref{eq:BMTI_loglikelihood_approx_gCorr}), which we report here for the reader's convenience:
        \begin{equation}
            \mathcal{L} ( \mathbf{F} \mid \hat{\bm{\delta F}}\,,\, \mathbf{D} ) \;:=\; - \sum_{i=1}^N \sum_{j\in \Omega_i} \frac{(F_j - F_i - \hat{\delta F}_{ij})^2}{2 \varepsilon_{ij}^2} \; .
        \end{equation}
        The predicted NLDs obtained through this approximation are very accurate and are also extremely similar to those obtained using the approximation in Sec.~\ref{sssec:BMTI_support_approx_inverse_C_least_square} of the SM.
        As for the uncertainties, however, this approach leads to systematically underestimating the errors on the NLDs by applying Eq.~(\ref{eq:BMTI_support_BMTI_F_errors_inverse_A}).
        Nonetheless, computationally this latter approximation is less demanding, since it does not involve the computation and the inversion of the matrix $\mathbf{C}$ - not the full inversion, nor the approximate one.
        Therefore, this is the approximation we adopt every time we are not interested in computing the uncertainties on $\hat{\mathbf{\hat{F}}}$.
        All the numerical experiments presented in Sec.~\ref{sec:numerical} are conducted using this setting, a sign that the approximation does not tragically compromise the performance of BMTI.

    \subsection{Maximisation of the BMTI likelihood with approximate inversion of $\mathbf{C}$}
    \label{ssec:BMTI_support_BMTI_diag_solution}

    In this section, we derive explicitly the elements in Eq.s (\ref{eq:BMTI_likelihood_maximisation}) and (\ref{eq:BMTI_F_estimator}) by maximising the approximate BMTI log-likelihood in Eq.~(\ref{eq:BMTI_loglikelihood_approx}).
    In particular, for a lighter notation, we carry out all the calculations using the approximation in Eq.~(\ref{eq:BMTI_loglikelihood_approx_gCorr}).
    The general solution of Eq.~(\ref{eq:BMTI_loglikelihood_approx}) is obtained with the simple substitution $\varepsilon_{ij}^2 \rightarrow {D_{ij}}^{-1}$.
    
    Let us maximise analytically the log-likelihood in Eq.~(\ref{eq:BMTI_loglikelihood_approx_gCorr}) with respect to the vector $\mathbf{F}$ by setting its gradient to zero for all its $N$ components $i$
    \begin{equation}
    \begin{split}
        0 = \frac{\partial ~}{\partial F_i} \mathcal{L} ( \mathbf{F} \mid \hat{\bm{\delta F}}\,,\, \mathbf{D} )
            &= - \sum_{k} \sum_{j\in \Omega_k} \frac{1}{\varepsilon_{kj}^2}  (F_j - F_k - \hat{\delta F}_{kj})(\delta_{ji} - \delta_{ki})\\
            &= - \sum_{j \mid i\in \Omega_j}  \frac{1}{\varepsilon_{ji}^2} (F_i - F_j - \hat{\delta F}_{ji}) +  \sum_{j\in \Omega_i} \frac{1}{\varepsilon_{ij}^2} (F_j - F_i - \hat{\delta F}_{ij})\ ,
    \end{split}
    \end{equation}
    where the notation $j \mid i \in \Omega_j$ indicates the set of points $j$ which include the point $i$ in their neighbourhood $\Omega_j$.
    Working out the above calculation further, we can bring all the maximisation parameters $\{F_i\}_i$ on the left-hand side of the equal sign. By also considering that $\hat{\delta F}_{ji} = - \hat{\delta F}_{ij}$ and $\varepsilon_{ji}^{2} = \varepsilon_{ij}^{2}$, we can rewrite
    \begin{equation}
        \sum_{j\mid i\in\Omega j}\frac{1}{\varepsilon_{ij}^{2}} \hat{\delta F}_{ij}
        + \sum_{j\in\Omega_{i}}\frac{1}{\varepsilon_{ij}^{2}}  \hat{\delta F}_{ij} 
        \; = \;
        \sum_{j\mid i\in\Omega j}\frac{1}{\varepsilon_{ji}^{2}}(F_{j}-F_{i}) 
        + \sum_{j\in\Omega_{i}}\frac{1}{\varepsilon_{ij}^{2}}(F_{j}-F_{i})
        \; .
    \end{equation} 
    Distinguishing on the right-hand side of the expression the sums in which $F$'s are summed over from those who can be factored out and defining
    \begin{equation}
        b_{i} 
        \; := \; 
        \left(
        \sum_{j\mid i\in\Omega j} +\sum_{j\in\Omega_{i}}
        \right)
        \left(
        \frac{1}{\varepsilon_{ij}^{2}}  \hat{\delta F}_{ij}
        \right)
        \; ,
    \label{eq:BMTI_support_b_i_definition}
    \end{equation}
    we obtain:
    \begin{equation}
        b_{i}
        \; = \;       
        \left(
        \sum_{j\mid i\in\Omega j} +\sum_{j\in\Omega_{i}}
        \right)
        \left(
        \frac{1}{\varepsilon_{ij}^{2}} F_j
        \right)
        -
        \left(
        \sum_{j\mid i\in\Omega j} +\sum_{j\in\Omega_{i}}
        \right)
        \left(
        \frac{1}{\varepsilon_{ij}^{2}}
        \right)
        F_i
        \; .
    \label{eq:BMTI_support_b_i_on_left_hand_side}
    \end{equation}
    In order to write this linear system in vector form, we introduce some definitions for notational convenience:
    \begin{equation}
        S_{i \leftarrow} \; := \; \sum_{j\mid i\in\Omega j} \frac{1}{\varepsilon_{j i}^{2}}
        ~~ , ~~
        S_{i \rightarrow} \; := \;\sum_{ j\in\Omega i} \frac{1}{\varepsilon_{i j}^{2}} 
        \qquad \mathrm{and} \qquad
        S_{i \leftrightarrow} \; := \; S_{i \leftarrow} + S_{i \rightarrow}
    \label{eq:BMTI_support_S_definitions}
    \end{equation}
    with the two arrow symbols respectively indicating the points for which $i$ is a neighbour ($i \leftarrow $) and the points in the neighbourhood of point $i$ ($i\rightarrow$). Then we rewrite Eq.~(\ref{eq:BMTI_support_b_i_on_left_hand_side}) inserting the definitions in Eq.~(\ref{eq:BMTI_support_S_definitions}) and using indicator functions for the sets $\Omega_i$ and $\Omega_j$:
    \begin{equation}
    \begin{split}
        b_{i} &= 
            \left(
            \sum_{j\mid i\in\Omega j} +\sum_{j\in\Omega_{i}}
            \right)
            \left(
            \frac{1}{\varepsilon_{ij}^{2}} F_j
            \right)
            -
            S_{i \leftrightarrow} F_{i}\\
        &=
            S_{i \leftrightarrow} (F_j)
            -
            S_{i \leftrightarrow} (1) \, F_{i}\\
        &=
        \sum_j \left[
            \frac{1}{\varepsilon_{ji}^2} I_{\{ i\in \Omega_j \}}
            +
            \frac{1}{\varepsilon_{ij}^2} I_{\{ j \in \Omega_i \}}
            -
            S_{i \leftrightarrow} \, \delta_{ji} 
        \right] F_{j}\\
        &=:
            \sum_j A_{ij} F_{j} \; .
    \end{split}
    \label{eq:BMTI_support_linear_system_gCorr}
    \end{equation}
    Note that the square brackets in the third line define all the elements of matrix $\mathbf{A}$.
    This equation contains a linear system of the exact form as in Eq.~(\ref{eq:BMTI_likelihood_maximisation}), with the definition of the matrix $\mathbf{A}$ and the vector of coefficients $\mathbf{b}$ for the approximate BMTI likelihood in Sec.~\ref{sssec:BMTI_support_approx_inverse_C_gCorr} of the SM given in Eq.s (\ref{eq:BMTI_support_linear_system_gCorr}) and (\ref{eq:BMTI_support_b_i_definition}) respectively.
    
    \subsection{Regularisation of the BMTI likelihood through a \textit{k}NN-based likelihood}
    \label{ssec:BMTI_support_BMTI_regularisation}
    As mentioned in Sec.~\ref{ssec:BMTI_likelihood_regularisation} of the main text, the BMTI log-likelohood can be regularised by combining it with the log-likelihood of a strictly-local normalised method via a mixing hyperparameter $\alpha$.
    We choose as regulariser some \textit{k}NN-based estimator \cite{Rodriguez2018,carli2022Tesi}.
    The total regularised likelihood in these cases reads
    \begin{equation}
        \mathcal{L}^{\mathrm{tot}} (\mathbf{F}\,|\,  \{ \{ v_{ij} \} _{j}\}_i\, , \,\hat{ \boldsymbol{\delta} \mathbf{ F}}\,,\, \mathbf{C})
        \;=\;
        \alpha \, \mathcal{L}^{\mathrm{BMTI}} (\mathbf{F}\,|\,  \hat{ \boldsymbol{\delta} \mathbf{ F}}\,,\, \mathbf{C})     
        \;+\;
        (1-\alpha)\,
        \mathcal{L}^{k\mathrm{NN}} ( \mathbf{F}\,|\, \left\{ \{ v_{ij} \}_{j} \right\}_i ) \; ,
    \label{eq:BMTI_support_BMTI+kNN_likelihood}
    \end{equation}  
    where the \textit{k}NN-based likelihood depends in principle on the hyperspherical volume shells $\left\{ \{ v_{ij} \}_{j} \right\}_i$ between the $j$-th and the $(j+1)$-th neighbours of the $i$-th point (c.f.r. Ref. \cite{Rodriguez2018}).
    Note that generally the \textit{k}NN-based likelihood will be factorisable, due to the assumption of independent estimates $F_i$at every point $i$:
    \[
    \mathcal{L}^{k\mathrm{NN}} ( \mathbf{F}\,|\, \left\{ \{ v_{ij} \}_{j} \right\}_i ) \;=\;\prod_i \mathcal{L}_{i,k_i}^{k\mathrm{NN}} ( F_i\,|\, \{ v_{ij} \}_{j} ) \; ,
    \]
    which implies that the regularisation will only affect the diagonal of the matrix $\textbf{A}$.
    By maximising over the parameters $\mathbf{F}$, this model yields an estimator which, as evident from panel B of Fig.~\ref{fig:times_pak_bmti}B, retains the advantages of both approaches.
    Note that $\mathcal{L}^{k\mathrm{NN}}$ ($\mathcal{L}_{i,k_i}^{k\mathrm{NN}}$) will typically not be a quadratic form in the $\mathbf{F}$ ($F_i$), unlike $\mathcal{L}^{\mathrm{BMTI}}$. 
    If one wants to write the solution of Eq.~(\ref{eq:BMTI_support_BMTI+kNN_likelihood}) again as a linear system, like in Eq.~(\ref{eq:BMTI_likelihood_maximisation}), one should approximate the $\mathcal{L}^{k\mathrm{NN}}$ by its quadratic order expansion around its maximum $\mathbf{F}^0$:
    \begin{equation}
        \mathcal{L}^{k\mathrm{NN}} ( \mathbf{F} ) \approx
        \mathcal{L}^{k\mathrm{NN} \; (2)}( \mathbf{F} ) := \frac{1}{2} \sum_{jk} (\mathbf{F}_j-\mathbf{F}_j^0)^{\trans} \, 
        \frac{\partial}{\partial F_k} \frac{\partial}{\partial F_j} \mathcal{L}^{k\mathrm{NN}}( \mathbf{F} ) \Big\vert_{\mathbf{F}^0}\,
        (\mathbf{F}_k-\mathbf{F}_k^0) \; .
    \label{eq:BMTI_support_likelihood_2nd_order_expansion}
    \end{equation}
    This is the actual likelihood that enters the regularised BMTI presented in Fig.~\ref{fig:times_pak_bmti}B.    

    \subsection{Sparse solvers for the linear system, complexity, and implementation details}
    \label{ssec:BMTI_support_complexity_linear_solvers}
        In this subsection we discuss the memory and time scaling of solving the linear system in Eq.~(\ref{eq:BMTI_support_linear_system_gCorr}), which results by maximizing the likelihood in Eq.~(\ref{eq:BMTI_loglikelihood_approx_gCorr}) with respect to $\mathbf{F}$. 

        \subsubsection{Sparsity of $\mathbf{A}$}
        \label{sssec:BMTI_support_complexity_linear_solvers_A_sparsity}
        
        In the current \textit{DADApy} \cite{Glielmo2022DADApyDA} implementation of BMTI, the maximum number of neighbours that a point can have is limited to a value $k_{\mathrm{max}}$, which is specified when initialising the dataset and computing the nearest-neighbours distances and neighbour list. For all data- and time-efficiency tests, shown in Fig.s~\ref{fig:killer_graph}B and \ref{fig:times_pak_bmti}A, we set $k_{\mathrm{max}} = 100$. Even with one million points, the adaptive neighbourhood selection algorithm only selects $k_i = k_{\mathrm{max}}$ for a small fraction of the datapoints, and there is no tangible accuracy gain in increasing $k_{\mathrm{max}}$.

        By looking at Eq.~\ref{eq:BMTI_support_linear_system_gCorr}, one can see that the upper bound for the number $N_{\mathrm{nz}}$ of non-zero elements of matrix $\mathbf{A}$ is $2 (N_e + N)$. 
        In turn, $N_e$ is smaller than or equal to $N\,(k_{\mathrm{max}}-1)$, which means that $N_{\mathrm{nz}} \leq 2\,N\,k_{\mathrm{max}}$. Since generally $k_{\mathrm{max}} \ll N$, matrix $\mathbf{A}$ is sparse. This allows to store it in sparse form, i.e. with an upper bound to memory complexity $\mathcal{O}(N)$. If $k_i$ were not bounded by $k_{\mathrm{max}}$, one could expect the average number of neighbours selected to scale with the inverse of the ID: $\langle k_i \rangle_i \propto N^{\frac{1}{d}}$, which means $N_{\mathrm{nz}} \propto N^{\frac{d+1}{d}}$.

        \subsubsection{Conjugate gradient sparse solver implementation and scaling}
        \label{sssec:BMTI_support_complexity_linear_solvers_CG}
    
         Besides for storing $\mathbf{A}$, this sparsity can be exploited in solving the linear system in Eq.~(\ref{eq:BMTI_support_linear_system_gCorr}). With a dense solver, the memory and time complexity would be $\mathcal{O}(N^2)$ and $\mathcal{O}(N^3)$ respectively. Using sparse solvers, linear scaling, i.e. $\mathcal{O}(N)$, can be achieved for both. We consider the iterative conjugate gradient (CG) sparse solver implemented in the \textit{scipy} Python library \cite{2020SciPy-NMeth} as \texttt{scipy.sparse.linalg.cg}. This is an approximate solver, i.e. it seeks convergence within a threshold. In our experience, the results obtained with CG do not show any sensible loss of accuracy with respect to the dense solver.    

        \paragraph{Complexity}
        The memory complexity of CG is dominated by storing the sparse matrix itself, so by $\mathcal{O}(N_{\mathrm{nz}})$. The time complexity scales as $\mathcal{O}(\nu \cdot N_{\mathrm{nz}})$, where $\nu$ is the number of iterations. In most cases, when $\mathbf{A}$ is well-conditioned, $\nu \ll N$, so the time complexity goes like $\mathcal{O}(N_{\mathrm{nz}})$. However, the upper bound for $\nu$ is $\mathcal{O}(N)$, so that for time complexity it becomes $\mathcal{O}(N \cdot N_{\mathrm{nz}})$; in these cases, preconditioning $\mathbf{A}$ might be worth the trade-off.

        Fig.~\ref{fig:times_pak_bmti}A presents the time scaling of BMTI as a function of the sample size $N$ for the $6d$ potential. As mentioned in Sec.~\ref{sssec:BMTI_support_complexity_linear_solvers_A_sparsity}, we set $k_{\mathrm{max}} = 100$. However, since in this setting, even with one million datapoints, most $k_i$'s do not saturate to $100$, we observe a slightly superlinear time scaling $\mathcal{O}(N_{\mathrm{nz}}) \sim \mathcal{O}(N^{\frac{d+1}{d}})$. We expect that saturation would be observed for larger dataset sizes.

        \paragraph{Implementation details} CG requires the matrix to be symmetric and semi-positive definite. Nonsingularity is necessary to guarantee convergence. Therefore, a good practice is to take a regularised version of $\mathbf{A}$ (although empirically we have often observed convergence even for non-regularised matrices). Our best option is using the regularised likelihood in Eq.~(\ref{eq:BMTI_support_BMTI+kNN_likelihood}). Even with a very small addition, like $\alpha = 0.99$, the terms added on the diagonal of $\mathbf{A}$ make it full-rank and ensure the correct normalisation of the NLDs $\mathbf{F}$. Alternatively, one could renormalise or rescale $\mathbf{A}$ and apply Tikhonov regularisaion $\mathbf{A} + \varepsilon \, \boldsymbol{\mathbb{1}} $, with $\boldsymbol{\mathbb{1}}$ identity matrix and $\varepsilon$ a small scalar.

        One source of potentially great speedup for the convergence of CG is providing it with an initial guess for the solution. Even a set of NLDs estimated with an extremely cheap $k$NN estimator could drastically reduce the number of CG iterations.

        \paragraph{Alternative: exact sparse solver} If an exact solution is needed, sparse direct solvers are preferred to CG, such as the one implemented in \texttt{scipy.sparse.linalg.spsolve}, based on \textit{SuperLU} decomposition of $\mathbf{A}$. This direct sparse solver achieves $\mathcal{O}(N_{\mathrm{nz}})$ and $\mathcal{O}(N_{\mathrm{nz}}^{\frac{3}{2}})$ for memory and time complexity respectively; these scalings are bounded by the dense solver's limits, $\mathcal{O}(N^2)$ and $\mathcal{O}(N^3)$, for ill-conditioned or not-so-sparse matrices.
\newpage 

\section{Test datasets}
\label{append:test_sys}

    \subsection{Synthetic distributions}
    \label{ssec:test_sys_synthetic}
        \begin{figure}[!h]
        \centering
            \includegraphics[width=\textwidth]{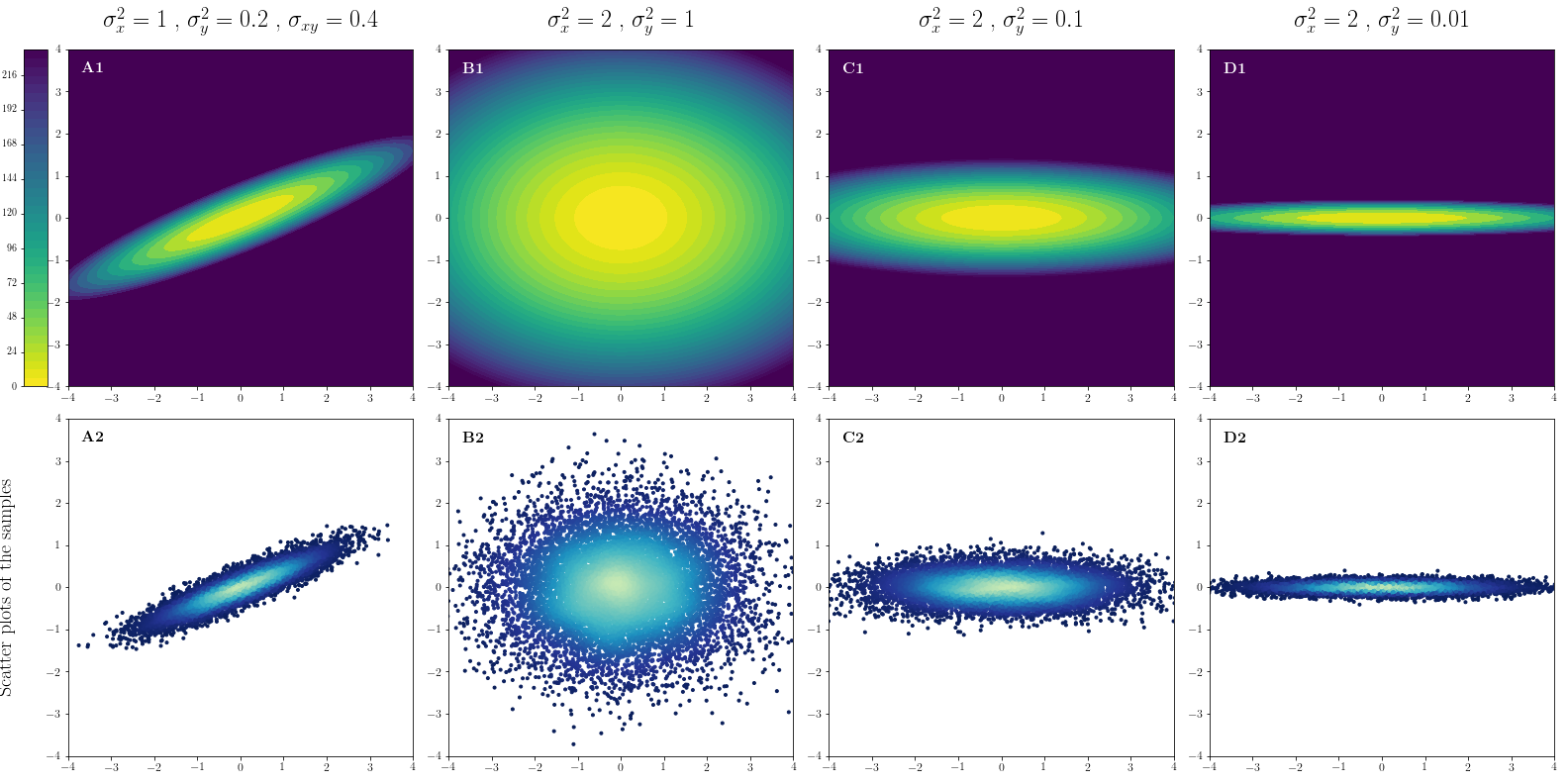}
        \caption[Bivariate Gaussian potentials used as test datasets.]{
        \label{fig:test_sys_2d_Gaussians}
        \textbf{Bivariate potentials from four Gaussian distributions used as test datasets}. 
        Each column represents a different dataset. All Gaussians are centred at the origin. The parameters of each dataset's covariance matrix are indicated in the header of each column.
        \textbf{Top}: contour plots of the potential surfaces. \textbf{Bottom}: four samples of $10,000$ points from the above potentials.
        }
        \end{figure}

        \subsubsection{2-dimensional Gaussian distibutions}
        \label{sssec:test_sys_2d_Gauss}
        
        These four systems have bivariate normal distributions $\mathcal{N}(\mathbf{0}, \bm{\Sigma})$. We name the two elements on the diagonal of this matrix $\sigma^2_x$ and $\sigma^2_y$, while the two identical off-diagonal terms are $\sigma_{xy}$:
        \begin{equation}
            \bm{\Sigma} = \begin{pmatrix}
                            \sigma^2_x & \sigma_{xy} \\
                            \sigma_{xy} & \sigma^2_y
                        \end{pmatrix}
        \end{equation}
        The first Gaussian, whose corresponding potential is represented in panel A1 of Fig.~\ref{fig:test_sys_2d_Gaussians}, has $\sigma^2_x=1$, $\sigma^2_y=0.2$ and $\sigma_{xy}=0.4$. 
        The second dataset, in panel B1, has $\sigma^2_x=2$, $\sigma^2_y=1$ and $\sigma_{xy}=0$. 
        The third dataset, in panel C1, has $\sigma^2_x=2$, $\sigma^2_y=0.1$ and $\sigma_{xy}=0$. 
        The fourth dataset, in panel D1, has $\sigma^2_x=2$, $\sigma^2_y=0.01$ and $\sigma_{xy}=0$. In the bottom row of the figure scatter plots of samples of $10,000$ points from the above potentials are shown.
        
        In the main text, when referring to a 2-d Gaussian, we mean a dataset constituted of $2,000$ points sampled from the first system, except for the performance of $\hat{\delta F}$ in Fig.~\ref{fig:deltaF_perform}, which is computed with $10,000$ points.

        \subsubsection{6-dimensional potential: 2-dimensional double well potential plus 4-dimensional harmonic directions}
        \label{sssec:test_sys_6d}

        \begin{figure}[!h]
        \centering
            \includegraphics[width=\textwidth]{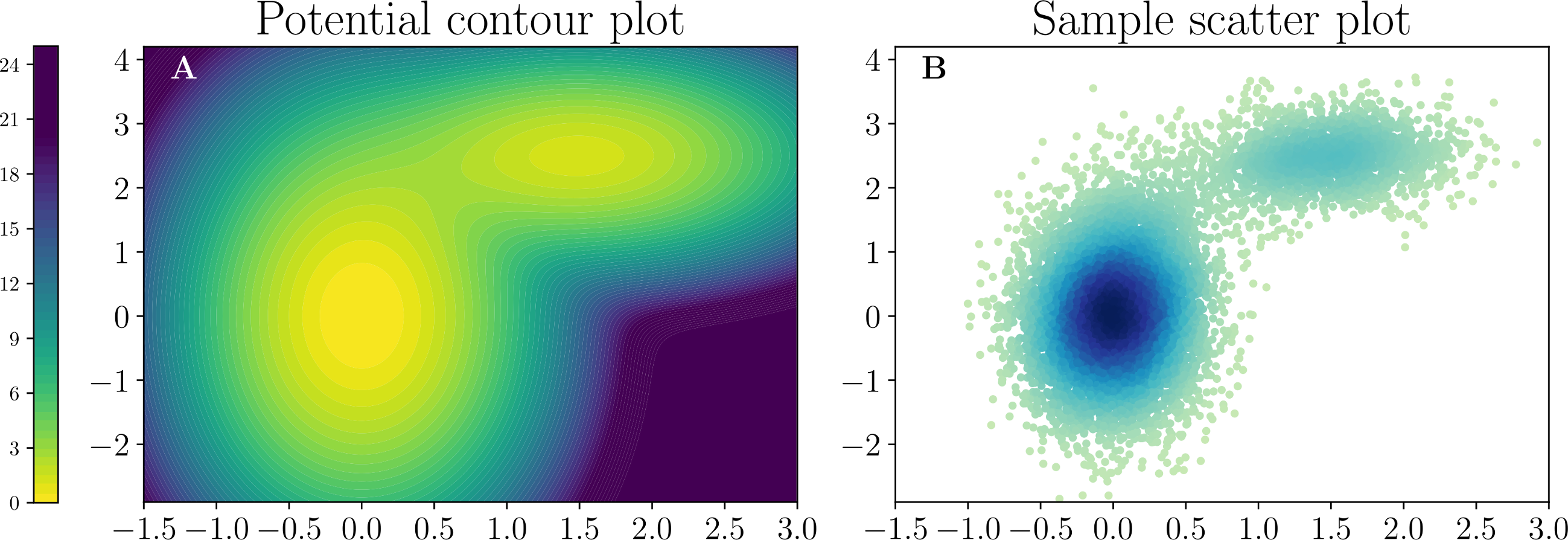}
        \caption[6d potential]{
        \label{fig:test_sys_6d}
        \textbf{Bivariate potential $\mathbf{U_{2d}}$ used to define the first two directions in the 6-dimensional potential}. 
        \textbf{A}: Contour plots of the potential surface. \textbf{B}: $10,000$ points sampled from the above potentials.
        }
        \end{figure}
        
        The negative logarithm of the PDF from which this set of $10,000$ datapoints is sampled is a potential which in the first two dimensions has the form:
        \begin{equation}
            U_{2d} \;:=\; \left(2\, e^{-(x\,-\,1.5)^2 \,-\, \,(y\,-\,2.5)^2 } 
            + 3\, e^{-2\,x^2 \,-\, 0.25\,y^2 }\right)^3 \, .
        \end{equation}
        while the additional 4 dimensions feel a (convex) harmonic potential centred at the origin and with unitary curvature.
        The $U_{2d}$ potential is represented in panel A of Fig.~\ref{fig:test_sys_6d}. 

    \subsection{Synthetic distributions with realistic features}
    \label{ssec:test_sys_synthetic_but_realistic}
    
        \subsubsection{2-dimensional Mueller-Brown potential}
        \label{sssec:test_sys_2d_MB}
        
        The dataset is a sample of $5,000$ points sampled from a PDF whose negative logarithm is proportional to the classical bivariate Mueller-Brown potential\cite{Mueller1979LocationOS}, whose expression is:
        
        \begin{equation}
            \begin{split}
            U_{\mathrm{MB}} ~~:= ~~15 \,&e^{0.7\,(x\,+\,1)^2 \,+\, 0.6\,(x\,+\,1)(y-1) \,+\,0.7\, (y-1)^2}
                                \,-\,200\, e^{- (x - 1)^2 -10 \,y^2} \\
                                -\,100\, &e^{- x^2 -10\, (y-0.5)^2}
                              \,-\,170\, e^{-6.5\,(x\,+\,0.5)^2 \,+\, 11\,(x\,+\,0.5)(y-1.5) -6.5\,(y-1.5)^2}
            \label{eq:test_sys_MuellerBrown_formula}
            \end{split}
        \end{equation}
        
        and whose contour plot can be seen in panel A of Fig.~\ref{eq:test_sys_MuellerBrown_formula}. From this potential, we generate a sample in a temperature range around which all three basins of the potential are visited even extracting only $5000$ points, which is quite an undersampling regime. We found that rescaling the potential by an inverse thermodynamic temperature $\beta = 0.035$, the saddle points are fairly, although slightly, populated, as can be seen in Fig.~\ref{eq:test_sys_MuellerBrown_formula}. The system displays a NLD barrier from the global minimum to the neighbouring basin which is around $\sim 3.7 k_B T$, as visible in panel C of Fig.~\ref{eq:test_sys_MuellerBrown_formula}. We use this setting in order to test our NLD estimators in conditions of moderate connectivity of the neighbourhood graph.
        
        In order to compute the minimum energy path (MEP) connecting the two main minima, we use the \textit{Nudged Elastic Band} algorithm\cite{Henkelman2000} in its improved tangent formulation \cite{Henkelman2000a} with $32$ images. For the exact location of the two minima we use the values in reference \cite{BREDEN2019MuellerBrownMinima}. We call the MEP for this system the polygonal chain that linearly interpolates between the $32$ images. Next, we want to find a path as close as possible to the MEP but which only connects points in the sample. We sample our MEP homogeneously 20 times for each image, so that we extract a set of $621$ points along the MEP. For each of these MEP points, we look for its nearest neighbour in the data sample we are considering. If the distance between the MEP point and the NN in the dataset is below a given threshold we keep the point, otherwise we reject it. The collection of all these sample points forms what we call the NN-interpolated MEP. For the data sample of $5000$ points extracted from the Mueller-Brown potential at $\beta = 0.035$, we consider a NN interpolation threshold of $2\times10^{-2}$. The NN-interpolated MEP contains $460$ points. The MEP is clearly visible as the dashed curve in panel A of Fig.~\ref{eq:test_sys_MuellerBrown_formula}. The NN-interpolated MEP is represented as a red solid line in panel B1. The ground truth NLD along the path is visible in red in panel A of Fig.~\ref{fig:killer_graph}.
        
        \begin{figure}[!ht]
        \centering
            \includegraphics[width=\textwidth]{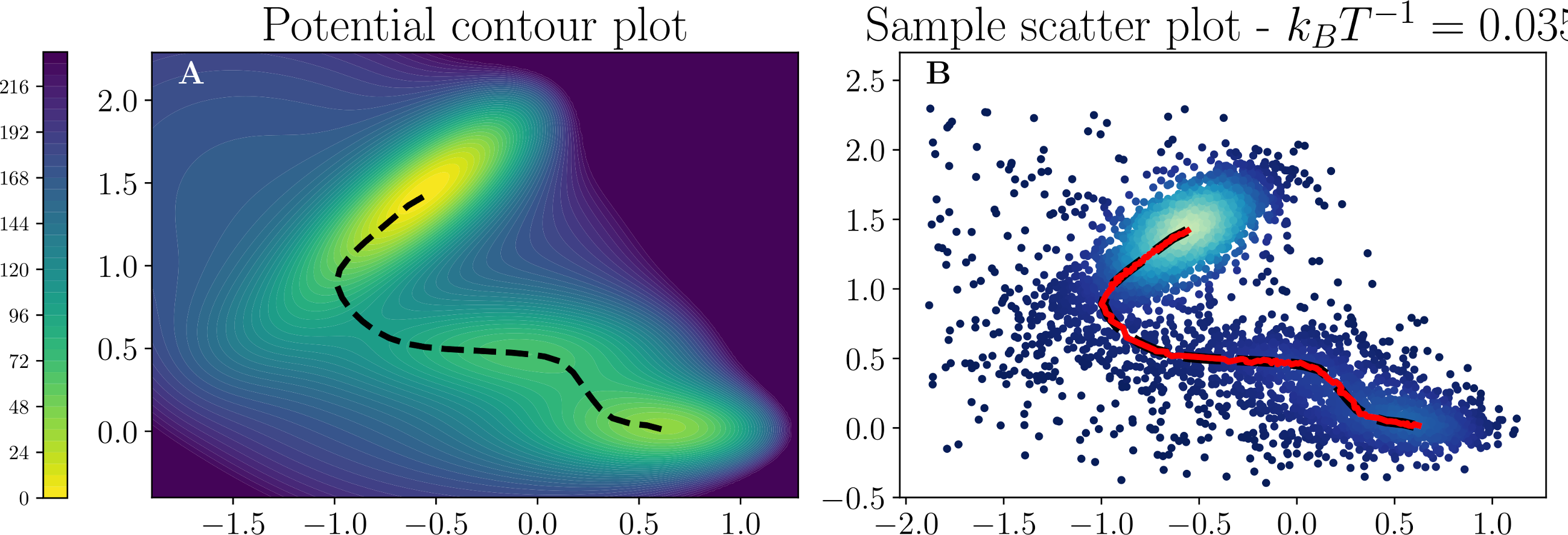}
        \caption[Scaled Mueller-Brown potential used as test system.]{
        \label{fig:test_sys_2d_MB}
        \textbf{Illustration of the Mueller-Brown potential used as test system}. 
        \textbf{A}: Contour plot of the Mueller-Brown potential in Eq.~(\ref{eq:test_sys_MuellerBrown_formula}). For the reader's convenience, the minimum of the potential has been shifted to $0$. Also, for better readability, the colour map has been cut to $230$, otherwise it would be saturated by the diverging behaviour in the top right corner. The black dashed curve represents the MEP connecting the two minima computed via the NEB algorithm.
        \textbf{B}: Scatter plots of $5,000$ points sampled from the Mueller-Brown potential. The scale factor is the inverse thermodynamic temperature $\beta = 0.035$. In red, the minimum energy path connecting the two main minima.
        }
        \end{figure}

        \subsubsection{2-dimensional multimodal potential on a glassy background}
        \label{sssec:test_sys_2d_MB_glassy}
    
        This is a synthetic potential which was designed in order to challenge density estimators despite being defined in a low-dimensional space ($D=2$). The dataset contains $10,000$ points sampled from the corresponding PDF. This dataset is not used in the main text, but only for additional tests presented in the SM.
        
        The PDF is obtained by superimposing on a box $[-4,4]\times[-2,2]$ with periodic boundary conditions the following distributions:
        
        \begin{itemize}
            \item a bivariate multi-peak PDF of form which integrates to $1$:
            \begin{equation}
                \begin{split}
                f_{\mathrm{mp}} ~~:= ~~    0.11\,
                \Large[
                3.4 \, &e^{-6.5\,(x\,+\,1)^2 \,+\, 11\,(x\,+\,1)(y-0.5) \,-\, 6.5\,(y-0.5)^2 }\\
                + 2\, &e^{-(x\,+\,0.5)^2 \,-\, 10\,(y\,+\,0.5)^2 }
                + 4\,e^{-(x\,-\,0.5)^2 \,-\, 10\,(y\,+\,1)^2 } 
                \Large] \; ,
                \end{split}
            \end{equation}
            rescaled by $0.6$;
            \item $90$ rescaled bivariate Gaussians $0.005 \times\mathcal{N}\left(\mathbf{\mu}, 0.04 \cdot \bm{\mathbb{1}}_2 \right)$ with the centres $\mathbf{\mu}$ randomly sampled in the rectangle  $[-3.6,3.6]\times[-1.8,1.8]$; their total integral is $0.18$;
            \item a uniform background which integrates to $0.22$.
        \end{itemize}
        
        Adding these three contributions, we obtain a PDF that displays metastability between the two main basins. Moreover, due to the presence of the Gaussians and of the constant background, we obtain a behaviour typical of glassy systems: there are very many local minima weakly populated by almost-isolated points. These features are designed to stress-test density estimators.

        \begin{figure}[!ht]
        \centering
            \includegraphics[width=\textwidth]{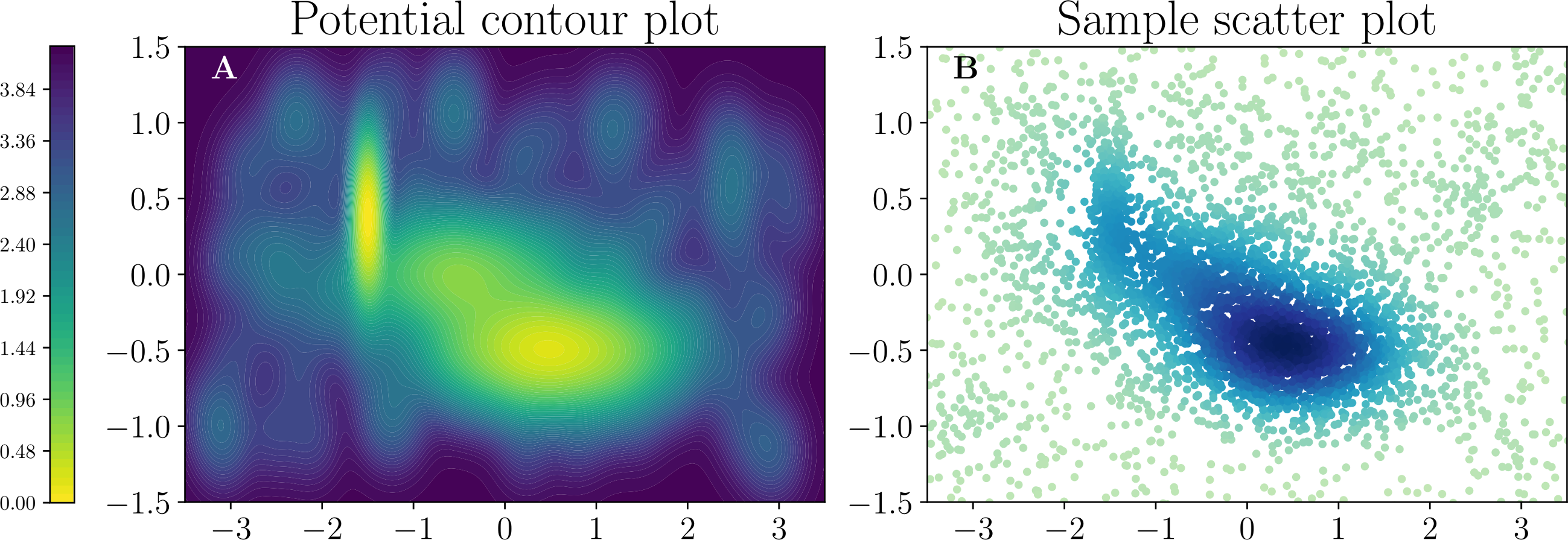}
        \caption[2-dimensional multimodal potential on a glassy background.]{
        \label{fig:test_sys_2d_MB_glassy}
        \textbf{2-dimensional multimodal potential on a glassy background}. 
        \textbf{A}: Contour plot of the negative logarithm of the PDF defined in Sec.~\ref{sssec:test_sys_2d_MB_glassy} of the SM. \textbf{B}: Scatter plots of $5,000$ points sampled from the potential.
        }
        \end{figure}
        
        \subsubsection{9-dimensional smoothed landscape of CLN025 decapeptide}
        \label{sssec:test_sys_CLN025}
            
        As a realistic system we consider a $\beta$-hairpin called CLN025\cite{honda2008crystal}. This molecule is a small protein of 10 residues and 166 atoms and is one of the smallest peptides that display a stable secondary structure, in this case a $\beta$-sheet. 
        Thanks to the relatively small size of the molecule we are able to sample all the relevant parts of its configuration space and then compute the ground truth NLD.
            
        \paragraph{Simulation of the system} We simulate the protein in Gromacs\cite{abraham2015gromacs} in explicit solvent. Since we are not interested in the precise physical chemistry of the system, we use quite a small box, resulting in a total of 2959 atoms, 166 of CLN025, the rest from the 931 water molecules. To enhance the sampling of configuration space, we run a Replica Exchange molecular dynamics \cite{sugita1999replica} simulation with 16 replicas using equally spaced temperatures from 340K to 470K as done previously in reference \cite{Rodriguez2011}.
        
        We choose as feature space the $9$-dimensional $\psi$-backbone-dihedra space. This choice implies of course a drastic dimensional reduction on the over-$400$-dimensional original atomic configuration space. Still, even after this huge projection, the dataset will show complex features and a reasonably high dimensionality, so that we are entitled to consider it a realistic case. Thus $D=9$ is the embedding space dimension. The distance between two configurations $\mathbf{X}^a$ and $\mathbf{X}^b$ in this space is:
        \begin{equation}
            \theta(\mathbf{X}^a, \mathbf{X}^b) = \sqrt{ \sum_n (( \psi^a_n - \psi^b_n ))^2}
        \end{equation}
        where $(( \bullet ))$ stands for $2 \pi$-periodicity within the brackets.
    
        \paragraph{Generation of the synthetic dataset via point-adaptive Gaussian KDE smoothing}
        We analyse a sample of $38000$ points in the space of the $\psi$-backbone-dihedrals. The estimated ID \cite{Facco2017} is $d=7$. With such ID we generate a $9$-dimensional smooth potential using our point-adaptive Gaussian KDE introduced in \cite{carli2022Tesi}. Therefore, we know the analytic value of the ground truth NLD everywhere. The ID of the smoothed dataset is $d=9$.  We sample $80,000$ points from this smoothed distribution. Throughout the performance assessment of BMTI in Sec \ref{ssec:numerical_performance} only a subsample of $20,000$ points is used, except for the performance of $\hat{\delta F}$ in Fig.~\ref{fig:deltaF_perform}, for which all $80,000$ points are retained.

    \subsection{Realistic datasets of analytically-unknown ground truth}
    \label{ssec:test_sys_realistic}
    
        \subsubsection{4-dimensional projection of GB3 protein}
        \label{sssec:test_sys_4d_GB3}
        The original dataset before embedding is the projection on four collective variables of the folding of the third IgG-binding domain of protein G from streptococcal bacteria (GB3)\cite{20dB}. The sample dataset has $10,000$ points.

        \subsubsection{20-dimensional embeddings of real NLD landscapes}
        \label{sssec:test_sys_20d}
        We consider two datasets used for the validation of PA$k$ estimator in reference \cite{Rodriguez2018} and briefly described in Sec.s \ref{par:test_sys_20dA},\ref{par:test_sys_20dC}. They are trajectories of respectively $2$ and $7$ CVs of which the ground truth NLD is known. All the datasets are treated in the same way to embed them in $20$ dimensions. Initially, the FES is resampled in the space of the collective variables with a probability proportional to the exponential of the negative of the NLD value. Then, the data points are twisted on a Swiss-roll by splitting the first of its coordinates in two by means of the transformation $x_1 = x \cos{x}$ and $x_2 = x \sin{x}$. Finally, a rotation around a random vector in $D = 20$ is performed. In this manner each point sampled from the original distribution is embedded in a $20$-dimensional space. 
        
            \paragraph{A: 2-dimensional dataset before embedding}
            \label{par:test_sys_20dA}
            The original dataset before embedding is the projection on two collective variables of the nucleation of the C-terminal of amyloid-$\beta$\cite{20dA}. We use a sample dataset of $2,000$ points -- a good sampling in $2$ dimensions, but a severe undersampling regime in $20$ dimensions -- to emphasise the effect of the curse of dimensionality.
            
            \paragraph{B: 7-dimensional dataset before embedding}
            \label{par:test_sys_20dC}
            The original dataset before embedding is the projection on seven collective variables of the conformational space of the intrinsically disordered protein human islet amyloid polypeptide (hIAPP)\cite{20dC}. We use $30,000$ datapoints from this dataset.

\newpage

\section{The adaptive neighbourhood selection}
\label{append:k_adaptive}
In this section we provide insight on the adaptive neighbourhood selection method introduced in Ref. \cite{Rodriguez2018} (Sec. \ref{append:k_adaptive-kstar_dc}) and we show how the selected neighbourhood graph (NG) affects the performance of the BMTI and the $k$NN density estimators (Sec. \ref{append:k_adaptive-effect_on_BMTI}).

    \subsection{The selection method at work}
    \label{append:k_adaptive-kstar_dc}
    \begin{figure}[!h]
        \centering
        \includegraphics[width=\textwidth]{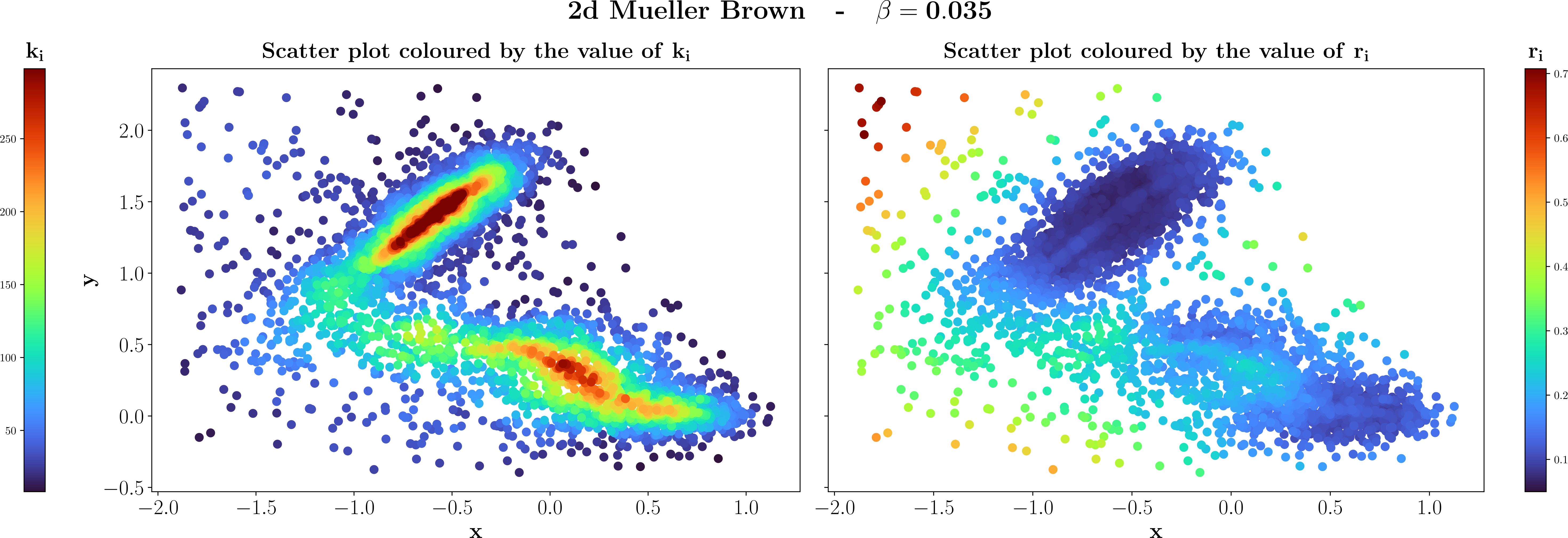}
        \caption{\label{fig:appendix-adaptive-neigh-selection}
        \textbf{Selected neighbourhoods}. Colour scatter plots illustrating the $k_i$ (left) and the corresponding $r_i := r_{k_i}$ selected for each point $i$ by the adaptive neighbourhood selection method on a 2d dataset of $5,000$ points sampled from the Mueller-Brown potential from Sec.~\ref{sssec:test_sys_2d_MB}.
        }    
    \end{figure}
    
    We illustrate the functioning of the adaptive neighbourhood selection procedure presented in Ref.~\cite{Rodriguez2018}.
    In short, for each point $i$, the method incrementally increases a test size $\tilde{k}_i$ and computes the $\tilde{k}_i$NN density of point $i$ and of its $\tilde{k}_i$-th nearest neighbour.
    A standard likelihood ratio test is used to assess whether the two densities are statistically compatible within a predefined threshold.
    If they are not significantly different, $\tilde{k}_i$ is increased by $1$, and the test is repeated. 
    This iterative procedure continues until the two densities at point $i$ and at its $\tilde{k}_i$-th nearest neighbour are found to be statistically different, at which point the value $k_i = \tilde{k}_i$ is selected.    
    
    In Fig.~\ref{fig:appendix-adaptive-neigh-selection}, we show the selected adaptive neighbourhood sizes, in terms of number of neighbours $k_i$ (left panel) and neighbourhood radius $r_i$ (right panel), on a 2d Mueller-Brown dataset.
    The algorithm handles the bias-variance tradeoff locally for via point-adaptive bandwidth selection.
    In low-density (high-energy) regions, the selected $k_i$ is, in fact, typically much smaller than in high-density (low-energy) ones, corresponding to the three basins of the potential (c.f.r. Fig.~\ref{fig:test_sys_2d_MB}). 
    At fixed density, the radius $r_i$ spanned by these $k_i$ neighbours is larger where the energy landscape is flatter, and smaller in the presence of sharp density variations.
    Still, in high-density regions, even a large number of neighbours will not span a volume of very large radius. 

    \subsection{Effects of the selected neighbourhood graph on BMTI and $k$NN performance}
    \label{append:k_adaptive-effect_on_BMTI}
    \begin{figure}[!h]
        \centering
        \caption{
        \label{fig:append-MAE_BMTIs}
        Comparison of BMTI (solid lines) against $k$NN (dashed lines) performance for various choices of NG, measured by the MAE as a function of dataset size $N$ on the 6d potential dataset (c.f.r. Sec. \ref{sssec:test_sys_6d}).
        Green corresponds to Abramson's rule of thumb \cite{Abramson1984}, $k=N^{D/{D+4}}$.
        This choice of fixed $k$ is the one adopted in all the $k$NN tests in the main text.
        }
        \begin{subfigure}{\textwidth}
            \includegraphics[width=.95\linewidth]{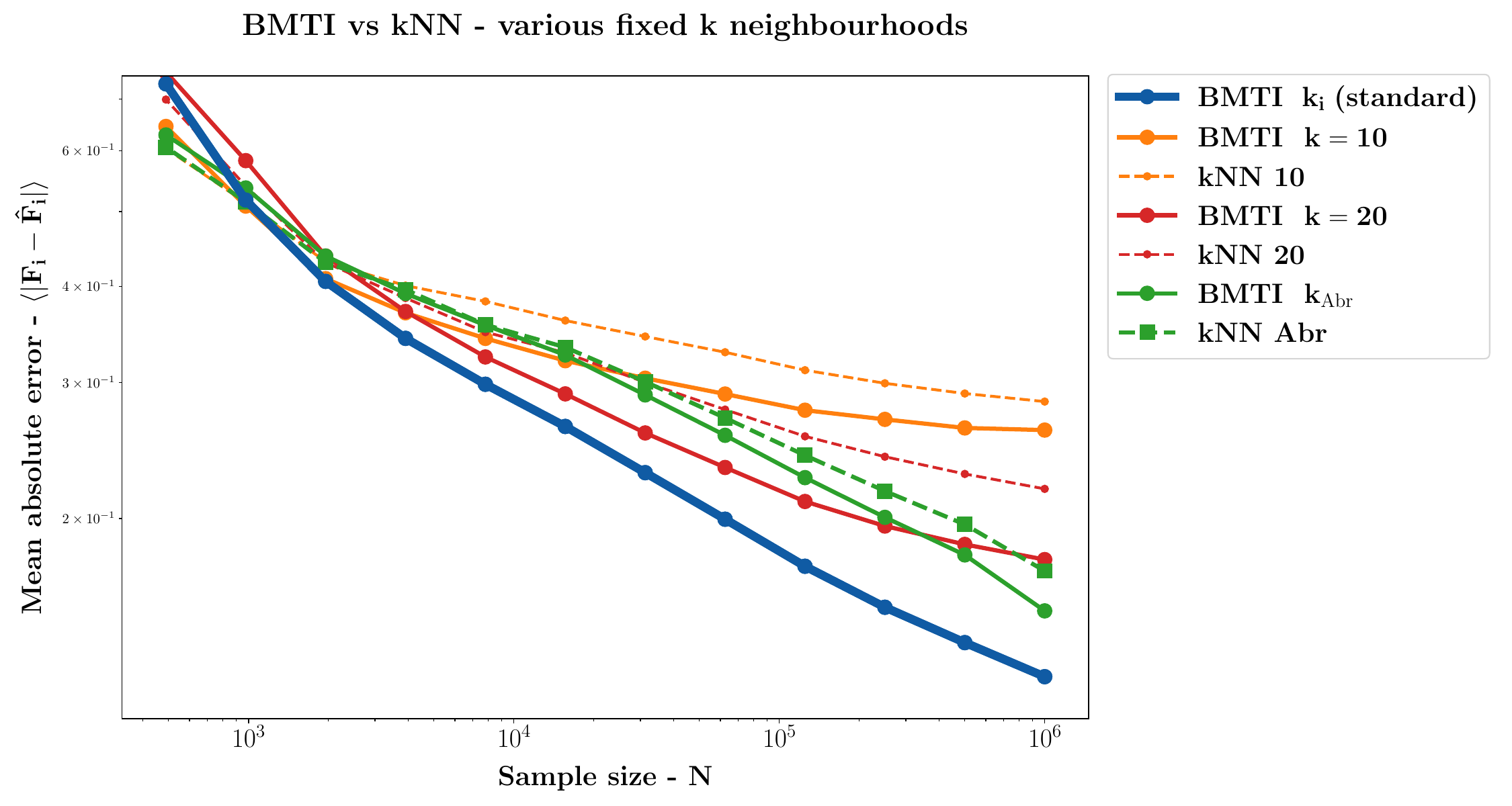}
            \caption{\label{sfig:append-MAE_BMTIs_const_k}
            Different colours correspond to different fixed-$k$ NGs, except for the blue, corresponding to the point-adaptive NG $\{k_i\}_i$ discussed in Sec. \ref{sssec:BMTI_deltaF_manifold+adaptive} of the main text and Sec \ref{append:k_adaptive-kstar_dc} of the SM.
            }
        \end{subfigure}
        \begin{subfigure}{\textwidth}
            \includegraphics[width=.95\linewidth]{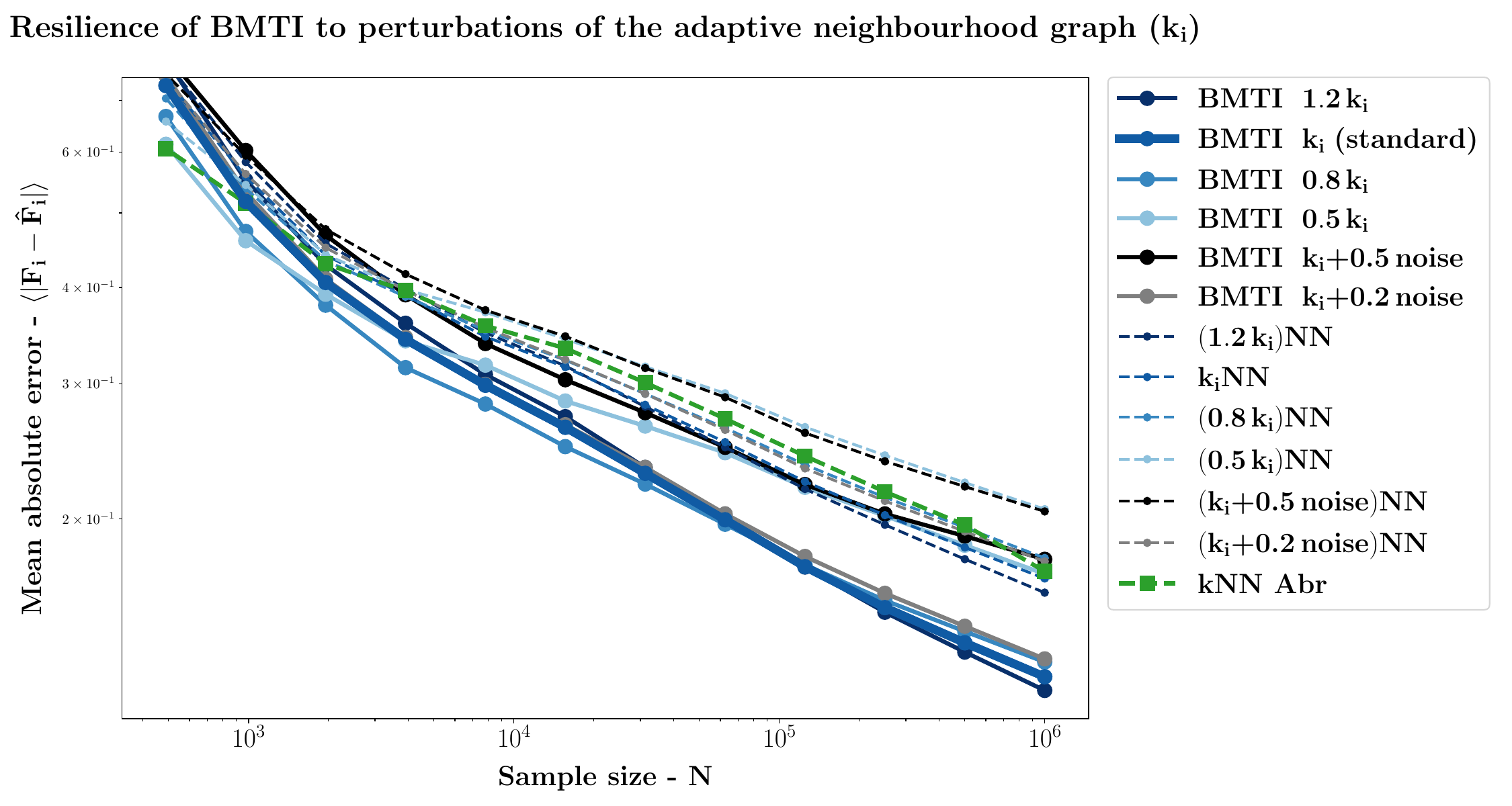}
            \caption{
            \label{sfig:append-MAE_BMTI_resilience}
            Different colours correspond to different point-adaptive NGs.
            $\{k_i\}_i$ indicates the point-adaptive NG used in standard BMTI and discussed in Sec. \ref{sssec:BMTI_deltaF_manifold+adaptive} of the main text and Sec \ref{append:k_adaptive-kstar_dc} of the SM.
            Shades of blue correspond to linearly increasing (darker) or decreasing (lighter) the number of neighbours $\{k_i\}_i$ considered for each point.
            Grey and black correspond to altering each neighbourhood size $k_i$ by adding a Gaussian random noise rounded to the closest unit.
            The standard deviation of the added Gaussian noise is taken as $20\%$ and $50\%$ of the mean value of $k_i$ over the whole dataset respectively.
            We regard these two cases as moderate and large added noise respectively.
            }
        \end{subfigure}
    \end{figure}
    
    In Fig.~\ref{sfig:append-MAE_BMTIs_const_k}, by comparing the dashed lines of one colour ($k$NN density estimators) to the solid lines of the same colour (BMTI with the same choice of fixed neighbourhood number $k$) we see that the second ones always have a smaller MAE w.r.t. the first ones, except below $\mathcal{O}(10^3)$ datapoints. 
    In other words, given a fixed NG, BMTI is always a more efficient density estimation algorithm than $k$NN, except for very small dataset sizes, where $k$NN is more robust.
    BMTI (depicted in blue), which adopts a point-adaptive NG, is the best-performing, as expected.

    The same conclusions regarding the BMTI outperforming $k$NN given the same choice of NG can be reached by looking at Fig.~\ref{sfig:append-MAE_BMTI_resilience}.
    In this test, however, an adaptive number of neighbours $k_i$ is used for each point $i$.
    Except for very low sample sizes, the dashed lines of a given colour stay well above the solid lines of the same colour.
    Interestingly, BMTI seems quite resilient to perturbations of our NG selection $\{k_i\}_i$.
    Most adaptive-NG versions of BMTI outperform the best-performing fixed-$k$
    $k$NN estimator, even when the optimally-selected $\{k_i\}_i$ is perturbed by moderate noise.
    The worst performing BMTI version, unsurprisingly, is the one to which a large random noise is added to the NG.
    The second worst is when only half the neighbours in our optimal NG are used for every point: in this case, due to the lower statistics, the local gradient estimates are noisier, and thus BMTI NLDs.
    Nonetheless, even in the worst cases, BMTI performance is comparable to the best $k$NN. 
    This suggests that BMTI would perform well even using other NG selection algorithms, as long as they are able to select big (small) $k$'s for high (low) density regions.    
    
\begin{figure}[!t]
        \centering
        \includegraphics[width=0.95\linewidth]{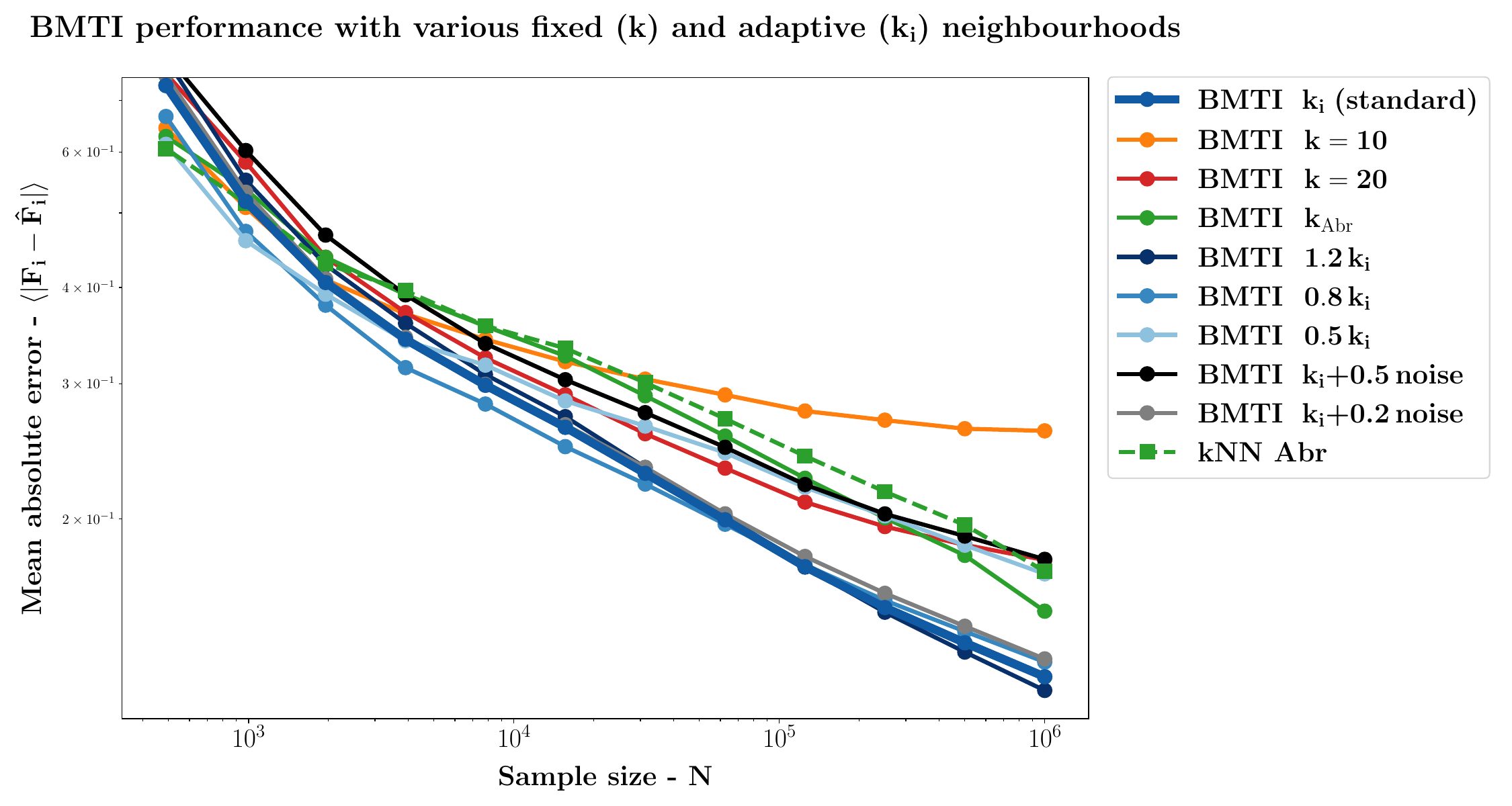}
        \caption{\label{fig:append-MAE_only_BMTIs_kNN_Abr}
        Performance (MAE as a function of dataset size $N$ on the 6d potential dataset (c.f.r. Sec. \ref{sssec:test_sys_6d})) of several BMTI density estimators for different choices of NG, compared to the standard $k$NN estimator adopted in the main text (solid lines vs dashed line).
        }    
    \end{figure}

    Finally, Fig.~\ref{fig:append-MAE_only_BMTIs_kNN_Abr} summarises all versions of BMTI presented in Fig.~\ref{fig:append-MAE_BMTIs} and compares the performance to the best performing fixed-$k$ $k$NN estimator (the dashed green line).
    As seen in the previous figure, in most cases BMTI clearly outperforms or at least marches the best-performing $k$NN. The only exception is when $k=10$ (yellow), in which case the statistical noise has a quite large lower bound. This plot highlights the effectiveness of BMTI as an algorithm, quite independently of the NG selection. Nonetheless, it also pictures the advantage of selecting the NG point-adaptively rather than using a fixed $k$.

\newpage
    
\section{Additional performance test: Kullback-Leibler Divergence}
\label{append:kld}

The Kullback–Leibler divergence (KLD), $D_{\mathrm{KL}}(p \,\|\, q)$, is often used to assess the quality of a density estimate $q$ relative to the true distribution $p$. It is not a metric in the strict mathematical sense, since
it is not symmetric and does not satisfy the triangle inequality. It can however be useful in quantifying the discrepancy between distributions and, due to its information-theoretical interpretation as relative entropy, it is often used in training ML models. It is defined as
    \begin{equation}
        D_{\mathrm{KL}}(p \,\|\, q) \;=\; \int p(\mathbf{x})\,\ln \!\left(\frac{p(\mathbf{x})}{q(\mathbf{x})}\right)\,\mathrm{d}\mathbf{x}.
        \label{eq:DKL_def}
    \end{equation}

If $p$ and $p$ are PDFs, i.e. they are normalised to $1$, the KLD is bound to be greater than zero, and equal to zero only if $p \equiv q$. In the case of a sample of $N$ i.i.d. points $\{\mathbf{x}_i\}_i$ drawn from $p$, it can be estimated empirically as
    \begin{equation}
        \hat{D}_{\mathrm{KL}}(p \,\|\, q) \;\approx\; \frac{1}{N} \sum_i \ln \!\left(\frac{p(\mathbf{x}_i)}{q(\mathbf{x}_i)}\right)\;,
        \label{eq:DKL_empirical}
    \end{equation}
which converges to Eq.~(\ref{eq:DKL_def}) in the limit $N \to \infty$. 

In our specific case, we have an estimator $\hat{F}$ of the NLD $F = - \ln \rho$, where $\rho$ is the ground truth PDF. Therefore, by using the equalities $\rho = e^{- F}$ and correspondingly $\hat{\rho} = e^{- \hat{F}}$, we can write the empiric KLD in Eq.~(\ref{eq:DKL_empirical}) as 
    
    \begin{equation}
        \hat{D}_{\mathrm{KL}}(\rho \,\|\, \hat{\rho}) \;\approx\; \frac{1}{N} \sum_i \ln \!\left(\frac{\rho(\mathbf{x}_i)}{\hat{\rho}(\mathbf{x}_i)}\right)
        \;=\;
        \frac{1}{N} \sum_i \hat{F}_i - F_i
        \;=\;
        \langle \hat{F}_i - F_i \rangle \;.
        \label{eq:DKL_empirical_F}
    \end{equation}

In principle, to evaluate $\hat{D}_{\mathrm{KL}}(\rho \,\|\, \hat{\rho})$ one would need to draw from the ground truth PDF a sample of points large enough to guarantee convergence of the empirical KLD to the analytical value. In practice, this would require sampling $\rho$ many times, and evaluating $\hat{F}$ even on points outside its training set. 

We choose, therefore, to compute the empirical KLD on the same points we use to estimate the $\hat{F}$s. This choice, however, might introduce a bias with respect to computing it on a separate sample. Intuitively, we could expect the nonparametrically-estimated densities to be systematically slightly over-estimated on the training sample with respect to the ground truth (overfitting). This can result in underestimating the KLD and can even lead to negative empirical KLD estimates for finite samples. To get around this problem, we propose a common estimation scheme for all the tested estimators. We assume the estimated PDFs are actually a not-properly normalised version $\tilde{\rho}$ and we compute the normalization constant empirically via importance sampling

\begin{equation}
    \mathcal{Z}_{\tilde{\rho}}
    \;=\;
    \int \tilde{\rho}(\mathbf{x}) \,\mathrm{d}(\mathbf{x})
    \;=\;
    \int \frac{\tilde{\rho}(\mathbf{x})}{\rho(\mathbf{x})} \, \rho(\mathbf{x}) \,\mathrm{d}(\mathbf{x})
    \;\approx\;
    \frac{1}{N} \sum_i \frac{\tilde{\rho}(\mathbf{x}_i)}{\rho(\mathbf{x}_i)}
    \;=\;
    \frac{1}{N} \sum_i e^{F_i} - e^{\tilde{F}_i}
    \;=\;
    \langle e^{F_i} - e^{\tilde{F}_i} \rangle
    \; ,   
\end{equation}

such that the normalised estimated PDF reads $\hat{\rho} = \tilde{\rho} / \mathcal{Z}_{\tilde{\rho}}$. This renormalised estimated density is the one we will actually compare to the GT $\rho$ over the training sample, computing $\hat{D}_{\mathrm{KL}}(\rho \,\|\, \hat{\rho})$:

\begin{equation}
    \hat{D}_{\mathrm{KL}}(\rho \,\|\, \hat{\rho}) 
    \;=\;
    \hat{D}_{\mathrm{KL}}(\rho \,\|\, \tilde{\rho}) + \ln \mathcal{Z}_{\tilde{\rho}}
    \;=\;
    \langle \tilde{F}_i - F_i \rangle  \,+\, \ln \, \langle e^{F_i} - e^{\tilde{F}_i} \rangle \;.
\end{equation}

Note that this renormalisation does not affect the shape of the estimated NLDs, it just amounts to the choice of an additive offset. 

\subsection{Results and discussion}
    \begin{figure}[!t]
        \centering
        \includegraphics[width=0.6\linewidth]{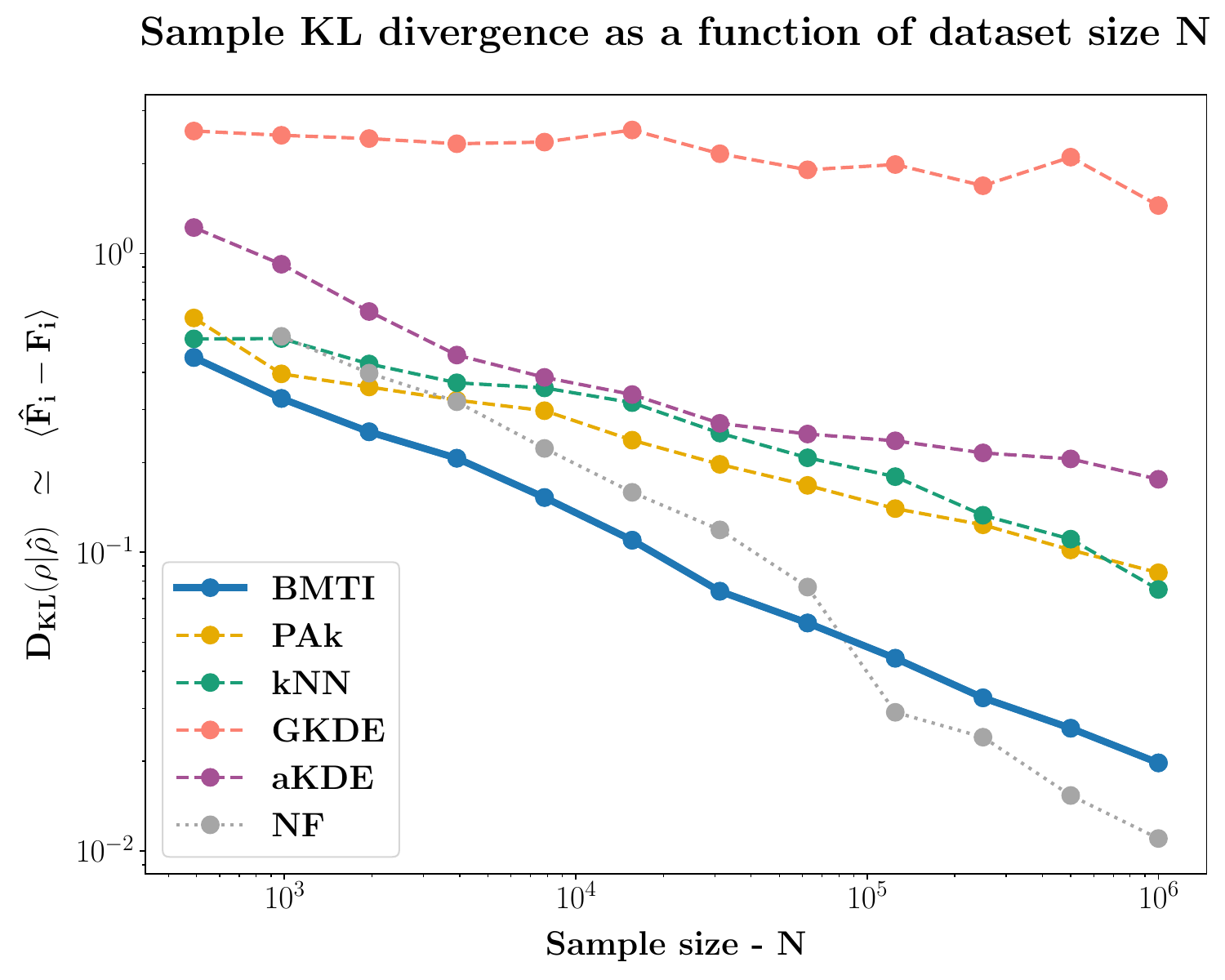}
        \caption{\label{fig:kld_1M}
        Kullback-Leibler divergence from the ground truth density to the renormalised estimated PDF as a function of sample size. The same nonparametric methods as in Fig.~\ref{fig:killer_graph}B are shown. Points in the plot are computed as mean MAE over 5 different runs. The standard deviations are very small even with a few hundred points, so they are not plotted.
        }    
    \end{figure}

The KLD from the GT to the renormalised estimated densities for various methods is shown in Fig.~\ref{fig:kld_1M}, which should be compared directly to Fig.~\ref{fig:killer_graph}B. The results are qualitatively the same, with BMTI outperforming all competing kernel-based nonparametric estimators across the board, and even normalising flows up to $\mathcal{O}(10^2)$ datapoints.

\vfill

\end{document}